\documentclass{article}

\usepackage{PRIMEarxiv}
\usepackage[utf8]{inputenc}
\usepackage[T1]{fontenc}
\usepackage{hyperref}
\usepackage{url}
\usepackage[numbers,sort&compress]{natbib}

\newif\ifpreprint
\preprinttrue

\usepackage{amsmath,amssymb}
\usepackage{graphicx}
\graphicspath{{figures/}{images/}}
\usepackage{adjustbox}
\usepackage{booktabs}
\usepackage{siunitx}
\usepackage{multirow}
\usepackage{tabularx}
\usepackage{float}
\usepackage[version=4]{mhchem}
\usepackage{microtype}
\usepackage[capitalise]{cleveref}

\usepackage{xspace}
\newcommand*{\ModelName}{GRACE\xspace}
\newcommand*{\Dataset}{CCSBench\xspace}

\renewcommand{\headeright}{Preprint}
\renewcommand{\undertitle}{Preprint}
\renewcommand{\shorttitle}{\ModelName}

\title{Predicting Collision Cross Sections with \ModelName: Geometric Residual Adduct
  Conditioning via Early-fusion}

\author{
  Parthasarathy Suryanarayanan$^{1,\ddag}$,\; Susanta Das$^{2,\ddag}$,\; Shreyans Sethi$^{1}$,\\
  \textbf{Kenneth M. Merz, Jr.}$^{2,3}$\thanks{email: kmerz1@gmail.com},\;
  \textbf{Joseph A. Morrone}$^{1}$\thanks{email: jamorron@us.ibm.com} \\[4pt]
  $^1$IBM Research, Yorktown Heights, NY 10598, United States \\
  $^2$Center for Computational Life Sciences, Cleveland Clinic Research, \\
  \quad The Cleveland Clinic, Cleveland, OH 44195, United States \\
  $^3$Department of Chemistry, Michigan State University,\\
  \quad East Lansing, Michigan 48824, United States
}

\date{}

\begin{document}

\maketitle

\begin{abstract}
Collision cross section (CCS), derived from ion mobility mass spectrometry, is a common descriptor for molecular annotation.  Prediction is challenging for machine learning models because it reflects the size, shape, and ionization state of a gas-phase molecular ion. Most predictors either ignore explicit 3D structure or treat adduct identity as a late categorical feature, which limits their ability to capture adduct-dependent geometric effects.  We present \ModelName{} (Geometric Residual Adduct Conditioning via Early-fusion), a 3D
CCS predictor that adapts a pretrained molecular geometry encoder using geometric residual
adduct conditioning via early fusion.
\ModelName{} combines two inductive biases: a residual objective relative to an
adduct-aware physical descriptor baseline, and adduct conditioning within the encoder via
a learned adduct token and low-rank attention adapters.
We evaluate the model on a curated set of over 9{,}000 experimental
molecule--adduct CCS records with random, scaffold, and adduct-sensitive splits designed to separate
interpolation, scaffold generalization, and adduct-driven generalization.
\ModelName{} achieves the best mean percentage difference among the evaluated learned models
on all three splits: 1.67\% on the random split, 2.11\% on the scaffold split, and 2.36\%
on the adduct-sensitive split.
Diagnostic analyses suggest that residual learning stabilizes training by removing the
dominant mass--CCS trend, while early fusion improves adduct-sensitive prediction relative
to late fusion. Across four independent external test sets, \ModelName{} shows consistently
lower error than the other evaluated models. On a held-out set, \ModelName{} attains the lowest
mean percent difference when compared with four previously reported physics-based workflows. These results support residual learning and encoder-level adduct conditioning as practical inductive biases for fast, accurate CCS prediction.
\end{abstract}


\section{Introduction}
\label{sec:intro}

Metabolomics has transformed our understanding of biological systems through its application
to disease markers~\cite{johnson2016metabolomics, klupczynska2024metabolomics,
vinayavekhin2010exploring, zhang2024recent} and its growing role in healthcare,
therapeutics, and precision medicine~\cite{puchades2017metabolomics, wishart2016emerging,
beger2020current}.
In practice, metabolomics experiments typically combine chromatography, high-resolution mass
spectrometry, and increasingly ion mobility separations to profile hundreds to thousands of
metabolites across cohorts with diverse clinical
phenotypes~\cite{gertsman2014validation, rochat2018lc, whiley2019ultrahigh,
fraisier2020ms, dwivedi2010metabolic, zhang2018application, paglia2022ion,
asef2022ion, rakusanova2024tips}.
In metabolomics, understanding complex biological systems depends on computational
methodologies~\cite{krumsiek2016computational, wishart2009computational} spanning feature
extraction and alignment, pathway and network inference, and structure- or property-based
models for annotating unknown signals.
A central bottleneck in metabolomics is structural annotation: a substantial fraction of
the measured metabolome consists of unknown or only partially characterized molecules, and
robust strategies to identify or prioritize candidate structures are urgently
needed~\cite{zhou2022metabolite, schrimpe2016untargeted, uppal2016computational,
nafie2025comprehensive, zhou2020ion, bingol2017knowns, da2015illuminating,
sindelar2020chemical, viant2017close}.

Ion mobility spectrometry (IMS) is now routinely used for metabolite identification and
annotation~\cite{zhang2018application, paglia2022ion, may2022integrating, zhou2020ion}.
IMS separates gas-phase ions according to their mobility in a buffer gas under an applied
electric field, from which the collision cross section (CCS) can be derived.
CCS is a physicochemical observable that reflects the effective size and shape of a charged
species in the gas phase and is highly sensitive to molecular structure and adduct
type~\cite{krasnova2025determination, wisanpitayakorn2024accurate, christofi2023ion,
song2023application}.
Different IMS implementations, such as drift tube, traveling wave, trapped ion, and
structures for lossless ion manipulations, provide complementary CCS measurements that can
be interconverted through appropriate calibration
schemes~\cite{gabelica2019recommendations, may2021resolving, hinnenkamp2018comparison,
ridgeway2018trapped}.
Over the past decade, large CCS datasets have been collected and made available in public
databases~\cite{zhou2020ion, zheng2017structural, payne2018metabolomics,
nichols2018untargeted, elapavalore2025pubchemlite, song2022collision, baker2024metlin,
fenn2009characterizing} for a variety of analytes and adducts, making CCS an established
orthogonal descriptor alongside accurate mass and MS/MS spectra for metabolite identification
in complex samples~\cite{paglia2014ion, zhou2018advancing,
nichols2018untargeted, witting2023re}.
An accurate predicted CCS value filters candidate structures before any structural
elucidation step by eliminating those whose predicted CCS falls outside the experimental
tolerance window, reducing the annotation search space.
Experimental CCS measurements typically exhibit relative uncertainties on the order of a
few percent, and values within approximately $3\%$ are often considered indistinguishable in
practice~\cite{zhou2016large, wisanpitayakorn2024accurate, rainey2022ccs, luo2023mass,
metz2024introducing}.
This experimental uncertainty sets a natural resolution limit for CCS-based annotation and
provides a physically meaningful tolerance for benchmarking and comparing computational
prediction schemes.

To complement experimental data and extend CCS coverage to unmeasured compounds, several
physics-based computational workflows have been developed for CCS
prediction~\cite{das2022silico, das2023rapid, zanotto2018high, kartowikromo2023collision,
ross2024computational, colby2019isicle, haack2023mobcal, gorges2025qcxms2,
gorges2025evaluation, armakovic2025advanced}.
These workflows generally involve three key components: (i) construction of a conformational
ensemble in three-dimensional space, (ii) quantum mechanical calculation of the charge
distribution and other electronic properties, and (iii) evaluation of CCS using methods
such as the trajectory method or related approximations.
High-throughput, low-level models such as those provided in RDKit
(e.g.\ ETKDG)~\cite{bento2020open} to AI-based force fields such as
Auto3D~\cite{liu2022auto3d} to semi-empirical energy coupled to enhanced sampling molecular
dynamics (CREST)\cite{pracht2024crest} span the accuracy-cost tradeoff.
The quality and diversity of the underlying conformer ensemble have a direct impact on the
predicted CCS, particularly for flexible metabolites with multiple low-lying minima.
Another key step is computation of the molecular charge distribution using electronic
structure (quantum mechanics) calculations.
End-to-end physics-based CCS workflows have been successfully applied in metabolomics
contexts; however, these approaches face practical limitations for large and chemically
diverse metabolite libraries, particularly when many adduct types must be considered,
molecules are large and flexible, or high throughput is required, since the computational
cost of generating and refining conformer ensembles and running CCS calculations scales
poorly with system size~\cite{das2022silico, borges2021quantum, keng2024eliminating,
st2020quantum, axelrod2022geom}.

Machine learning offers a practical alternative for CCS
prediction~\cite{plante2019predicting, zhou2020ion, ross2020breaking, li2023collision, liao2025sigmaccs2,
impact2025collision, xie2024large}.
These models run in milliseconds per molecule and scale to library sizes that are
impractical for physics-based workflows.
Early work used small curated datasets and hand-crafted descriptors; more recent models
use deep neural networks trained on much larger corpora.
While some use molecular SMILES or chemically bonded graphs as input (e.g.,
DeepCCS~\cite{plante2019predicting}, GraphCCS~\cite{xie2024large}), others use 3D
representations of the molecule as input and accordingly depend on a conformational engine
as in physics-based models (e.g., SigmaCCS~\cite{guo2023highly}).  Molecular foundation models~\cite{zhou2023uni, suryanarayanan2024multi, choi2025perspective} that have been pretrained on large corpora of molecular graphs, SMILES, or 3D geometries may be fine-tuned or incorporated in models for tasks such as CCS prediction.  Published CCS models are typically evaluated on random held-out splits of existing datasets; evaluation splits designed to probe generalization across adduct types or chemical scaffolds have not been reported, making it difficult to assess which models perform reliably in practice.
Adduct identity alters gas-phase conformation and charge distribution, yet its treatment in
deep learning models has been largely ad hoc: typically a one-hot encoding appended at the
prediction head, with no principled approach to conditioning a 3D encoder on ionization
state throughout the representation-learning layers.
Machine learning models for CCS prediction have also been developed as unconstrained
regressors, with little effort to incorporate physio-chemical descriptors or chemical inductive
biases into the architecture or training objective.

Here we present \ModelName, a CCS predictor that combines 3D molecular geometry with
chemically motivated inductive biases.
We evaluate it on \Dataset{}, a new benchmark with splits designed to expose adduct-driven
generalization failures.

Our main contributions are:
\begin{itemize}
  \item \Dataset{}, a curated benchmark of 9{,}209 experimental CCS values spanning
    [\textit{M}+H]\textsuperscript{+}, [\textit{M}$-$H]\textsuperscript{$-$}, and
    [\textit{M}+Na]\textsuperscript{+} adducts, with three complementary evaluation
    splits: a random split for standard benchmarking, a scaffold split for out-of-distribution
    generalization, and a novel adduct-sensitive split that places the most adduct-responsive molecules
    exclusively in the test set, targeting models that treat adduct identity as a categorical
    tag rather than a geometric signal and thereby exposing whether a model has learned the
    conformational consequences of different ionization states.
  \item A 3D CCS prediction model with two chemically motivated design choices: adduct
    conditioning within the encoder attention layers via parameter-efficient adapters,
    and residual learning against a physio-chemical descriptor baseline that substantially reduces the
    generalization gap across all three splits; the improvement holds on four
    independent external test sets, and on a prospective set of 20 amino acids and
    metabolites the model reaches accuracy near the experimental CCS measurement limit,
    with less than half the error of the best quantum mechanical workflow and a fraction
    of its computational cost.
\end{itemize}

\ifpreprint
In Section~\ref{sec:dataset} we describe dataset curation and the three evaluation splits.
In Section~\ref{sec:methods} we describe the model architecture and training procedure.
Section~\ref{sec:results} presents quantitative results on internal and external test sets,
including the prospective experimental comparison.
Section~\ref{sec:discussion} examines the geometry-dependent and counterfactual findings, and
Section~\ref{sec:conclusion} summarizes the main findings.
\fi
\section{Related Work}
\label{sec:related}

The main distinction among machine-learning CCS predictors is how they treat molecular
structure, whether geometry is absent, approximated through precomputed descriptors, or
encoded from explicit 3D conformers.

\noindent\textbf{Descriptor- and sequence-based models.}
Early CCS predictors used hand-crafted molecular descriptors with shallow regressors.
MetCCS~\cite{zhou2016large} used support vector regression over 14 molecular descriptors
to build a large predicted CCS database.
AllCCS~\cite{zhou2020ion,10.1021/acs.analchem.3c02267} broadened this into an ion-mobility atlas that combines curated
experimental records with large-scale predicted values for metabolite annotation.  CCSBase~\cite{ross2020breaking} developed machine learning models for CCS prediction with high accuracy on diverse chemical structures that were trained on a comprehensive dataset of experimental measurements.  A recent update to CCSBase~\cite{Bantwal_CCSBase2} expanded the  experimental database by combining multiple public sources and compared their ML model with state of the art 3D- and graph-based models.
CCS-Predictor~2.0~\cite{rainey2022ccs} uses descriptors computed from SMILES strings in
an open-source workflow focused on false-positive filtering in untargeted metabolomics.

DeepCCS~\cite{plante2019predicting} moved from descriptor-only regression to a
convolutional network using SMILES strings together with adduct information.
These methods are efficient for annotation workflows, but none explicitly models molecular
conformers or the geometric response to ionization; adduct identity enters as a categorical
label rather than shaping conformer generation or geometry encoding.

\noindent\textbf{Adduct-aware graph models.}
GraphCCS~\cite{xie2024large} constructs adduct graphs from molecular structures and adduct
types, and trains a very deep graph convolutional network across multiple adduct species.
This encodes adduct identity in the graph topology rather than as a post hoc categorical
variable.
The adduct graph is still a heuristic structural representation; it captures a putative
ionized topology without generating or relaxing an adduct-specific 3D conformer.
GraphCCS is best understood as an adduct-aware graph model, not a conformer-based 3D
model.

Other graph-based CCS models augment molecular connectivity with auxiliary topological,
physicochemical, or geometric descriptors.
They improve over pure descriptor regression by learning from graph structure, but whether
geometry enters as derived descriptors, explicit conformer coordinates, or invariant
quantities varies across methods.

\noindent\textbf{Explicit geometric and conformer-based models.}
SigmaCCS~\cite{guo2023highly} generates a single 3D conformer, encodes it as a molecular
graph with Cartesian coordinates alongside chemical identity node features, and processes
it with edge-conditioned convolutional layers.
A one-hot adduct vector is concatenated to the pooled graph representation before the
final prediction layers, so adduct identity enters after geometric encoding rather than
shaping it. SigmaCCS2~\cite{liao2025sigmaccs2} uses a dual molecular and line graph representation.
A single ETKDG+MMFF94 conformer is generated per molecule; the resulting bond lengths and
dihedral angles are encoded as node and edge features of the line graph, while atom-type and
topological features form the molecular graph. Adduct identity is appended as a one-hot
vector after pooling. The approach encodes 3D-derived geometric quantities as fixed
descriptors rather than passing Cartesian positions directly to the encoder. IMPACT-4CCS~\cite{impact2025collision} combines \textit{ab initio} calculations with
machine-learned potentials.
It is more physically grounded than direct learned predictors but substantially more
expensive.

We use GraphCCS as the principal learned baseline because it is the most directly
relevant adduct-aware graph model: it explicitly encodes ionization through graph
construction and has been evaluated across multiple adduct species.
It therefore tests whether an adduct-aware topological representation is sufficient
without explicit conformer geometry.
We additionally include SigmaCCS as a complementary explicit-conformer baseline.
SigmaCCS tests the opposite design choice: whether 3D conformer encoding is sufficient
when adduct identity is fused only after graph-level representation learning.
Together, GraphCCS and SigmaCCS isolate the two methodological alternatives most
relevant to \ModelName{}: adduct-aware graph construction without adduct-conditioned
conformer geometry, and explicit conformer encoding without early adduct-conditioned
representation learning.
We re-evaluate both on our curated dataset using the same splits and random seeds as
\ModelName{}.

\section{Dataset and Evaluation Splits}
\label{sec:dataset}

\subsection{Dataset Curation}

Several ion mobility spectrometry (IMS) data sources are publicly available.
We based our dataset on those curated by previous CCS modeling efforts, namely
GraphCCS~\cite{xie2024large} and SigmaCCS~\cite{guo2023highly}, which are derived from CCSBase~\cite{ross2020breaking,pubchem_ccsbase_2026} and AllCCS~\cite{zhou2020ion,10.1021/acs.analchem.3c02267}, as well as datasets
provided by experimental laboratories, the Unified CCS Compendium~\cite{picache2019collision} and MetlinCCS~\cite{baker2023metlin}.
Because there is overlap in molecules across these sources, the combined dataset requires
careful cleaning and de-duplication prior to machine learning.
We focus on three adduct types: protonated ([\textit{M}+H]\textsuperscript{+}),
deprotonated ([\textit{M}$-$H]\textsuperscript{$-$}), and sodiated
([\textit{M}+Na]\textsuperscript{+}).
These are the most common adduct states in the compiled experimental datasets and are also
most relevant for comparison with physics-based models.

The data obtained from select sources do not contain SMILES strings.
We therefore obtained SMILES representations from compound identifiers.
For each sample, we first verified the presence of a valid CAS number.
We then verified that each sample had either a valid molecular formula or an InChI Key.
Next, ions were stripped from the SMILES representation of each molecule.
To further validate the SMILES, we regenerated the molecular formula from the stripped SMILES
and verified that it matched the provided molecular formula.
If a mismatch was found, we compared atom-wise counts between the SMILES-derived molecule and
the given molecular formula and discarded any samples where discrepancies remained after this
verification step.
We note that the molecular formula is not unique to a given compound; however, this procedure
is a sanity check that in particular ensures correctness of the hydrogen count.

\begin{figure}[H]
  \centering
  \includegraphics[width=0.7\textwidth]{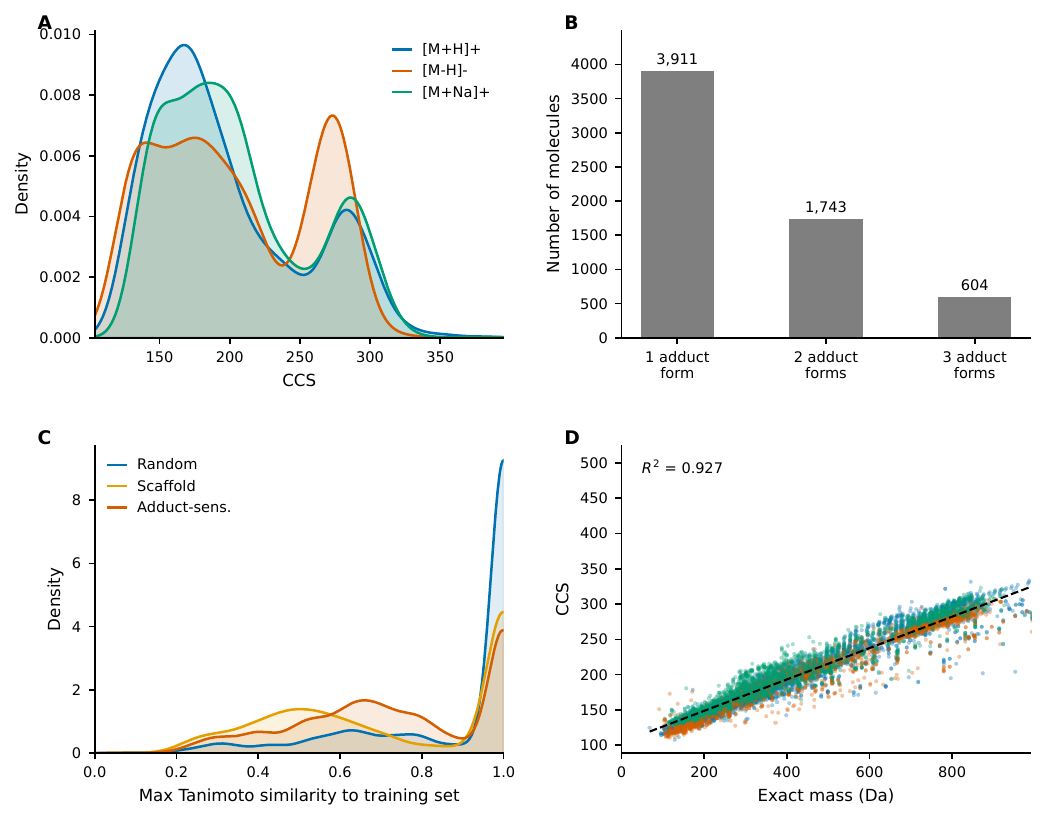}
  \caption{\Dataset{} statistics ($n = 9{,}209$ CCS measurements).
    \textbf{(A)}~Distribution of CCS values by adduct type: \mbox{[M+H]$^+$},
    \mbox{[M$-$H]$^-$}, and \mbox{[M+Na]$^+$}. Adducts differ systematically in
    CCS range and distribution shape.
    \textbf{(B)}~Number of molecules measured in one, two, or three adduct forms
    (3{,}911 / 1{,}743 / 604), showing the adduct coverage of the dataset.
    \textbf{(C)}~Maximum Tanimoto similarity (ECFP4) to the training set for
    molecules in the random, scaffold, and adduct-sensitive test sets. The scaffold
    and adduct-sensitive splits have lower similarity to training, confirming
    structural diversity between partitions.
    \textbf{(D)}~CCS as a function of exact molecular mass; molecular mass alone
    explains 92.7\% of variance ($R^2 = 0.927$), motivating the linear Ridge
    baseline used in residual learning.}
  \label{fig:dataset-stats}
\end{figure}

The curated experimental data were combined with the GraphCCS and SigmaCCS datasets.
Samples were filtered to retain only the three relevant adduct types.
Adduct notation was standardized across sources, and ions were stripped consistently prior to
canonicalization.
All molecules were canonicalized using their SMILES representation so that each entry was
uniquely identified by its canonical SMILES string and adduct type.
To resolve duplicate entries corresponding to the same (canonical SMILES, adduct) pair with
multiple CCS measurements, we applied the following procedure:
\begin{enumerate}
\item If duplicate CCS values differed by less than 3\%, one representative sample was retained.
\item If two duplicates differed by more than 3\%, both were discarded.
\item If more than two duplicates exhibited differences greater than 3\%, we applied Hartigan's dip test to assess whether the distribution of CCS values was bimodal. If bimodality was detected all duplicates were discarded, otherwise we removed the most significant outlier by Z-score and re-evaluated the remaining values, retaining one sample if they agreed within 3\% and discarding all otherwise.
\end{enumerate}
After duplicate resolution, one sample per (canonical SMILES, adduct) pair was retained.  We note that SMILES with different chiral markings were considered as distinct 
\ifpreprint
molecules.  See Section~\ref{sec:steroisomerleakage} for further discussion.
\else
molecules (see Discussion).
\fi
We refer to the curated benchmark as \Dataset{}.
Figure~\ref{fig:dataset-stats} A,B summarizes \Dataset{}.
In our experiments we use 9{,}209 measurements from \Dataset{}; approximately 8\% of
molecules could not be successfully processed into a set of three-dimensional conformers and were
excluded.
These molecules tend to be larger and fall outside the mass range typical of metabolites in \Dataset{} (Figure~\ref{fig:supp-excluded-mols}).

% This can be a separate subsection but I think works as a simple paragraph
In addition to \Dataset{}, we assess our model on four additional sets.
The first two are derived from external tests used to validate GraphCCS in Xie, et al.~\cite{xie2024large}.
Another is the experimental dataset provided in the work of Colby, et al.~\cite{colby2019isicle}.
Adducts outside those present in the training set and duplicate SMILES-adduct pairs (ignoring chirality) with respect to \Dataset{} are removed from the source datasets.
We refer to these sets as Testset~1 (TS1), Testset~2 (TS2), and Testset~3 (TS3), respectively.
Finally, we validate \ModelName{} using the 20 molecules in the gas-phase conformation library (GPCL), also referred to as TS4~\cite{das2024molecular}.
For evaluations involving GPCL, duplicate SMILES-adduct pairs (ignoring chirality) are removed from \Dataset{} and the model is retrained.
The external test sets are summarized in Table~\ref{tab:supp-ext-datasets}.

\subsection{Evaluation Splits}

We construct three complementary evaluation splits of increasing difficulty.

\noindent\textbf{Random split.}
Molecules are assigned to train, validation, and test at an 80/10/10 ratio by random
sampling (7{,}374 / 913 / 922 samples).
This split provides a standard i.i.d.\ benchmark but conflates interpolation with
generalization: test molecules often share scaffolds and adduct distributions with training
molecules, so strong performance here does not imply OOD robustness.

\noindent\textbf{Scaffold split.}
When applicable, molecules are grouped by Bemis--Murcko scaffold~\cite{bemis1996properties} and scaffolds
are assigned entirely to train or test, ensuring that no scaffold appears in both partitions
(7{,}369 / 920 / 920 samples).
This split is intended to probe generalization to chemically novel scaffolds and is a
standard out-of-distribution benchmark in molecular property prediction in computational
drug discovery.
Approximately 3{,}300 molecules in \Dataset{} do not contain a Murcko scaffold; these are
distributed randomly between train and test.

\noindent\textbf{Adduct-sensitive split.}
For molecules measured in more than one adduct, we compute the observed CCS range
(max minus min across available adducts).
Molecules whose CCS range exceeds 7.8~\AA$^2$ (the top 50\% by adduct
sensitivity, with a mean range of 14.2~\AA$^2$ in this subset) are
restricted to the validation and test sets (6{,}446 / 1{,}381 / 1{,}382 samples).
Training uses the remaining molecules, which are either measured in a single adduct or
show small CCS variation across adducts.
This split is a targeted stress test rather than an i.i.d. deployment estimate; it isolates
cases where adduct-dependent CCS variation is large and probes whether models treat adduct
identity as more than a categorical offset.
A protonated and a sodiated form of the same molecule occupy different regions of
conformational space, since the adduct atom modifies local electrostatics and can shift the
preferred ring geometry, intramolecular hydrogen-bonding network, or degree of folding.
For molecules where this conformational response is large, a model blind to adduct geometry
will systematically underfit.
Standard random and scaffold splits spread adduct-sensitive molecules uniformly across
train and test, diluting this signal; the adduct-sensitive split makes it the primary
evaluation criterion.

\section{Methods}
\label{sec:methods}

\subsection{Problem Formulation}

Collision cross section (CCS) is a physicochemical property measured in ion mobility
spectrometry (IMS) that reflects the effective rotationally averaged size and shape of an
ion as it drifts through a buffer gas under an applied electric field.
For a given molecular ion, CCS depends on its three-dimensional geometry, intramolecular
compactness, and the spatial distribution of partial charges induced by ionization and adduct
formation.
We pose CCS prediction as a supervised regression problem: given a neutral molecule
$\mathcal{M}$ and an experimental adduct label $a$, learn a mapping
\[
  f: (\mathcal{M}, a) \rightarrow \mathbb{R},
\]
where the scalar output is the CCS of the measured adducted ion.

Unlike many molecular properties that correlate primarily with local functional groups or 2D
connectivity, CCS is inherently a global, geometry-dependent quantity.
Molecules with identical molecular formulas or similar 2D graphs can exhibit substantially
different CCS values due to differences in three-dimensional conformation, flexibility, and
long-range intramolecular interactions.
Representations that rely solely on SMILES or 2D graph features may struggle to capture
the spatial extent and shape anisotropy that govern ion mobility, especially for flexible
or adduct-sensitive molecules.
Effective CCS prediction requires features that encode both local chemical identity and
global 3D structure, including pairwise interatomic distances and overall molecular
compactness.

Direct supervised learning of CCS from 3D inputs presents several challenges.
First, CCS datasets are modest in size and heterogeneous across sources and instrument
conditions, and the label noise floor is non-negligible.
Second, CCS measurements implicitly marginalize over ensembles of thermally accessible
conformations, while typical training data provides only a finite set of sampled conformers.
Third, CCS depends on subtle geometric and electrostatic effects that are difficult to infer
from sparse supervision alone.
We therefore frame CCS prediction as a transfer learning problem: a general-purpose 3D
molecular encoder pretrained on unlabeled structures is adapted to the downstream CCS
regression task, incorporating three chemically motivated design choices described below.

\subsection{Conformer Generation}

\begin{figure}[tbp]
  \centering
  \includegraphics[width=\textwidth]{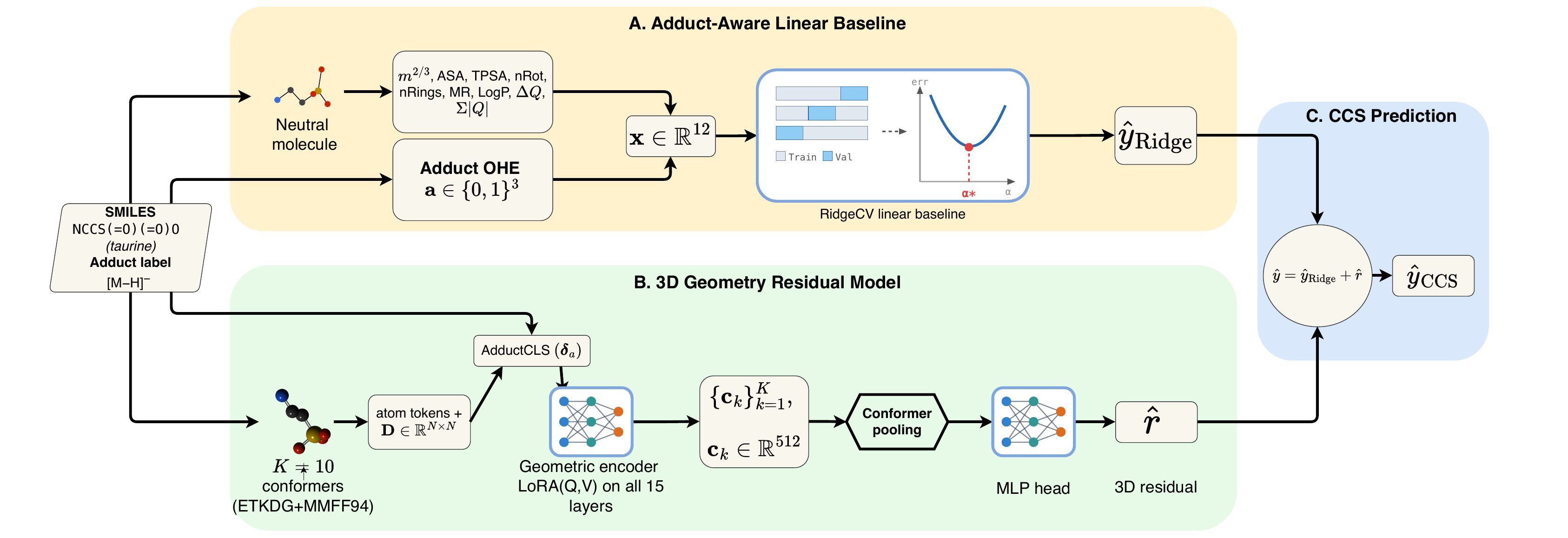}
  \caption{\textbf{\ModelName{} architecture.}
  CCS predictions are formed by adding an adduct-aware linear
  baseline~(A) and a 3D geometry residual~(B).
  \textbf{(A)}~A RidgeCV model is fit on a 12-dimensional feature vector
  comprising nine molecular descriptors
  ($m^{2/3}$, Labute ASA, TPSA, rotatable-bond count, ring count, MR,
  $\log P$, Gasteiger $\Delta Q$, $\Sigma|Q|$)
  concatenated with a three-dimensional adduct one-hot encoding
  $\mathbf{a}\!\in\!\{0,1\}^3$, yielding a linear CCS estimate
  $\hat{y}_\text{Ridge}$.
  \textbf{(B)}~$K{=}10$ conformers per molecule (RDKit ETKDG\,+\,MMFF94)
  are encoded by a pretrained geometric backbone (15 transformer layers) augmented
  with two adduct-conditioning mechanisms:
  (\textit{i})~AdductCLS, a zero-initialised per-adduct offset
  $\boldsymbol{\delta}_a$ injected at the \texttt{[CLS]} token; and
  (\textit{ii})~low-rank adaptation of the query and value projections
  (LoRA, rank\,16) in every layer.
  The $K$ per-conformer \texttt{[CLS]} embeddings
  $\mathbf{c}_k\!\in\!\mathbb{R}^{512}$ are pooled and passed through an
  MLP head to produce a geometry-dependent residual~$\hat{r}$.
  \textbf{(C)}~The final prediction
  $\hat{y}_\text{CCS} = \hat{y}_\text{Ridge} + \hat{r}$
  combines the linear and 3D contributions.
  The input example is taurine
  (\texttt{NCCS(=O)(=O)O}, $[\text{M}-\text{H}]^{-}$,
  CCS\,=\,117.9\,\AA$^2$).}
  \label{fig:overview}
\end{figure}

Figure~\ref{fig:overview} provides an overview of the \ModelName{} architecture and
prediction pipeline.
For each (SMILES, adduct) pair, we generate $K = 10$ three-dimensional conformers using
RDKit ETKDG~\cite{riniker2015better} followed by MMFF94 force-field
optimization~\cite{halgren1996merck}.
Conformers are generated on the \emph{neutral} molecule for two reasons.
First, generating conformers on the neutral molecule avoids the ambiguity of where charged
groups should reside in the chemical structure.
Second, the pretrained encoder (see below) was also trained on neutral
ETKDG+MMFF conformers; generating conformers with the same protocol eliminates distribution
shift between pretraining and fine-tuning geometries.

For Boltzmann-weighted pooling, per-conformer weights are computed as
\[
  w_i = \frac{e^{-(E_i - E_{\min})/k_BT}}
             {\sum_j e^{-(E_j - E_{\min})/k_BT}},
\]
where $E_i$ is the MMFF94 strain energy, $E_{\min}$ is the lowest energy in the ensemble,
and $k_BT = 0.5924$~kcal/mol at 298~K.

\subsection{Encoder}

Several architecture families have been applied to 3D molecular property prediction.
SE(3)-equivariant networks~\cite{thomas2018tensor,schutt2018schnet,gasteiger2020directional}
enforce rotational and translational symmetry at the representation level, but they are
computationally more expensive than distance-based invariant transformers.
Because CCS is a scalar, rotationally averaged observable, an invariant representation
based on pairwise distances is sufficient for the present model.
Message-passing GNNs operating on 3D graphs capture local geometry well but struggle with
long-range interactions, which matter for CCS because the effective projected area of a
molecule depends on its global fold, not just local bonding environments.
Transformers with distance-conditioned attention biases attend globally across all atom
pairs at every layer, making them better suited to CCS as a shape-sensitive quantity.
We use UniMol~\cite{zhou2023uni} as the geometry encoder backbone; it was pretrained on
19~million molecules with explicit 3D conformers using the same ETKDG+MMFF protocol we
adopt for conformer generation, eliminating distribution shift between pretraining and
fine-tuning geometries.
The geometry encoder backbone is a 15-layer transformer (47M parameters) pretrained with
masked atom prediction and 3D coordinate denoising as objectives.
Pairwise interatomic distances are encoded as attention biases via a Gaussian radial basis
expansion rather than as graph edges, enabling global geometric reasoning without assuming
local connectivity.
The encoder produces a 512-dimensional CLS token as the global molecular representation and
per-atom embeddings ($N \times 512$) for atom-level operations.

\subsection{Conformer Pooling}

Given $K$ conformers per molecule, we pool their CLS embeddings before passing to the
prediction head.
We evaluate four pooling strategies:

\noindent\textbf{Single conformer.}
One conformer is selected at random during training and by lowest MMFF94 energy at inference.
No pooling is performed.

\noindent\textbf{Uniform pooling.}
The arithmetic mean of all $K$ CLS embeddings.

\noindent\textbf{Boltzmann-weighted pooling.}
A weighted mean using the Boltzmann weights $w_i$ defined above.

\noindent\textbf{Learned attention pooling.}
A single-layer attention module learns scalar weights over the $K$ CLS embeddings,
with the weights normalized by softmax.

\subsection{Adduct Conditioning}

CCS is measured on an ionized species, and the adduct ([\textit{M}+H]\textsuperscript{+},
[\textit{M}$-$H]\textsuperscript{$-$}, [\textit{M}+Na]\textsuperscript{+}) directly
influences gas-phase geometry and effective molecular compactness.
The pretrained encoder is adduct-blind by construction: given the same neutral-molecule
conformer it produces an identical CLS embedding regardless of adduct.
A model that injects adduct identity only at the prediction head cannot allow that
information to modulate the geometric reasoning inside the encoder.
We therefore adopt \emph{early fusion}, introducing two zero-initialized components that
together break adduct symmetry while preserving the full pretrained representation at
initialization.

\noindent\textbf{AdductCLS embedding.}
A learnable embedding table $\Delta_a \in \mathbb{R}^{3 \times 512}$ stores one
512-dimensional delta per adduct.
The delta for the active adduct is added to the CLS token immediately after input embedding
normalization, before the transformer stack; atom positions are untouched.
Because the CLS token attends to all atom positions at every subsequent layer, this
injection propagates adduct information into every attention operation throughout the
encoder.

\noindent\textbf{QV LoRA adapters.}
We insert rank-16 LoRA~\cite{hu2022lora} corrections ($\alpha = 32$, effective
scale $\alpha/r = 2$) into the query and value projections of all 15 attention layers,
leaving key projections unchanged.
This design preserves pretrained atom-key representations while allowing adduct-dependent
query routing and value aggregation.
Both components are zero-initialized, making them exact no-ops at $t = 0$.

The two components are co-trained with the MLP head: encoder learning rate
$\eta = 10^{-5}$ with a 10-epoch linear warmup, head learning rate $\eta = 10^{-4}$.
The base encoder (47M parameters) stays frozen throughout.
Parameter counts: AdductCLS delta 1,536; QV LoRA adapters 491,520; MLP head 198,401;
total trainable 735K (1.53\% of the 48M full model).

\ifpreprint
For ablation purposes we also evaluate a \emph{late fusion} baseline; architectural details
and results are in Section~\ref{sec:late_fusion}.
The development trajectory from late fusion through successive early fusion variants is
documented in Section \ref{sec:supp_design_progression} (Figure~\ref{fig:design-progression}).
\else
For ablation purposes we also evaluate a \emph{late fusion} baseline; architectural details
and results are given in Discussion.
The development trajectory from late fusion through successive early fusion variants is documented in Supporting Information (Figure~\ref{fig:design-progression}).
\fi

\ModelName{} does not generate adduct-specific conformers; it conditions the representation
of neutral conformers on the adduct label.
The model can therefore learn adduct-dependent corrections to neutral geometry but cannot
explicitly simulate ion-induced conformational rearrangement.

\subsection{Residual Learning}

CCS is dominated by molecular size.
An ion drifting through a buffer gas is slowed by collisions whose frequency scales with
its rotationally averaged cross-sectional area, so larger, heavier molecules have larger
CCS values almost by construction.
To quantify how much of the CCS signal is captured by simple 2D descriptors, we fit a
series of RidgeCV regressions of increasing complexity on the training set and evaluated
them on the test set (Table~\ref{tab:descriptor-baseline}).
A single-feature model using molecular mass already explains 92.7\% of variance.
Replacing raw mass with mass\textsuperscript{2/3} (which scales as the surface area of a
sphere of equivalent volume) reduces RMSE further, consistent with CCS being a
projected-area quantity rather than a volume quantity.
Adding the solvent-accessible surface area and a small set of 2D electronic and topological
descriptors (Table~\ref{tab:residual-descriptors}) fills in most of the remaining variance:
the full descriptor-plus-adduct baseline reaches RMSE $7.82$~\AA$^2$ ($R^2 = 0.9791$) with no neural
component.
The reference ML model, GraphCCS, sits at RMSE $4.82$~\AA$^2$; the gap between it and the Ridge
baseline is the portion attributable to learned molecular features, including stereochemistry,
conformational flexibility, and adduct-induced geometry changes that simple descriptors
cannot capture.

\begin{table}[H]
  \caption{Physicochemical descriptors used in the Ridge baseline and residual target.
    All features are computed from the 2D molecular graph; no 3D conformer is required.
    The adduct one-hot (3 binary features, features 10--12) is appended, giving a 12-dimensional input to RidgeCV.}
  \label{tab:residual-descriptors}
  \centering
  \small
  \begin{tabular}{clll}
    \toprule
    \# & Feature & RDKit call & What it measures \\
    \midrule
    1 & mass\textsuperscript{2/3}  & \texttt{ExactMolWt(mol)**\,(2/3)}          & Exact MW scaled to CCS volume-like scaling \\
    2 & CalcLabuteASA              & \texttt{CalcLabuteASA(mol)}                & Labute surface area from VDW and bond adjacency \\
    3 & CalcTPSA                   & \texttt{CalcTPSA(mol)}                     & Topological polar surface area (2D fragments) \\
    4 & nrotb                      & \texttt{CalcNumRotatableBonds(mol)}        & Rotatable bond count from 2D graph \\
    5 & nrings                     & \texttt{CalcNumRings(mol)}                 & Ring count from 2D SSSR \\
    6 & MolMR                      & \texttt{MolMR(mol)}                        & Crippen molar refractivity \\
    7 & MolLogP                    & \texttt{MolLogP(mol)}                      & Crippen logP \\
    8 & Gasteiger range            & $\max(q)-\min(q)$ over atoms               & Charge spread; polarity of adduct-binding sites \\
    9 & Gasteiger abs sum          & $\sum|q|$ over atoms                       & Total charge magnitude; H-bonding capacity \\
    \bottomrule
  \end{tabular}
\end{table}

\begin{table}[H]
  \caption{Ridge regression baseline ablation on the random split.
    Features are added cumulatively to a RidgeCV model.
    CCS is 93.2\% explained by molecular mass alone ($R^2=0.9316$);
    the full descriptor baseline reaches RMSE~7.82
    before any neural component.}
  \label{tab:descriptor-baseline}
  \centering
  \begin{tabular}{lccc}
    \toprule
    Feature set & RMSE & MPD & $R^2$ \\
    \midrule
    mass                                                      & 14.14 & 4.94 & 0.9316 \\
    mass + adduct                                             & 13.59 & 4.71 & 0.9369 \\
    mass\textsuperscript{2/3} + adduct                        & 12.83 & 4.40 & 0.9437 \\
    mass\textsuperscript{2/3} + adduct + ASA                  &  8.80 & 3.36 & 0.9736 \\
    mass\textsuperscript{2/3} + adduct + ASA + TPSA + nrotb + nrings &  8.02 & 3.17 & 0.9780 \\
    full descriptor-plus-adduct baseline                       &  \textbf{7.82} & \textbf{3.04} & \textbf{0.9791} \\
    \midrule
    GraphCCS (reference)                                      &  4.82 & 1.74 & 0.9961 \\
    \bottomrule
  \end{tabular}
\end{table}

Given this structure, training a neural network to reproduce the full CCS signal from
scratch wastes gradient budget on re-learning the mass correlation.
We instead train the network to predict only the residual
$r_i = y_i - \hat{y}_i^{\text{Ridge}}$, where $\hat{y}_i^{\text{Ridge}}$ is the Ridge
prediction from the nine molecular descriptors and three-dimensional adduct one-hot encoding.
The final linear layer of the MLP head is zero-initialized, so the model starts by
predicting exactly the Ridge baseline and learns the adduct- and geometry-dependent
remainder from the first epoch.

\subsection{Training Details}

All experiments use AdamW optimization with a cosine learning rate
schedule.
Models are trained with early stopping on validation MSE (patience = 20 epochs).
Single-conformer models use batch size 32; multi-conformer models use batch size 8 to
accommodate the $K = 10$ conformers per molecule.
Each configuration is run with five independent random seeds (0--4); we report mean and
standard deviation across seeds.
Final evaluation uses the best-validation-loss checkpoint per seed.
The primary metric is mean percentage difference (MPD); we also report RMSE in
\AA$^2$, Pearson $r$, Spearman $\rho$, and the generalization gap
(test RMSE minus train RMSE at the best-validation checkpoint) as a measure of overfitting.

Atom coordinates are centered to remove translational degrees of freedom; no rotational
alignment is applied because the attention biases operate on pairwise distances, which are inherently rotation-invariant.  During training, the full backbone is updated using two optimizer groups: backbone  parameters at a learning rate of $10^{-5}$ and head parameters at $10^{-4}$.
\section{Results}
\label{sec:results}

Table~\ref{tab:result_internal_best} summarizes the best \ModelName{} configuration per
split against GraphCCS and SigmaCCS on the \Dataset{} test set.
Using MPD as the primary metric, \ModelName{} performs best across all three splits;
RMSE is similar to GraphCCS on the scaffold split.
The generalization gap for \ModelName{} is consistently lower than for GraphCCS, reflecting
the regularizing effect of the residual learning objective.
The following subsections report the internal and external results in detail and discuss
conformer pooling and fusion strategy trade-offs.

\begin{table}[H]
  \centering
  % Old: Internal test set: best model configuration per split. Bold: lowest Test MPD.
  \caption{\ModelName{} achieves the lowest MPD on all three \Dataset{} splits and competitive RMSE relative to GraphCCS and SigmaCCS. For \ModelName{}, the conformer pooling mode minimising Test MPD per split is selected. Mean percentage difference (MPD) is the primary metric; RMSE and the generalization gap (Gap $=$ test $-$ train RMSE at the best-validation checkpoint) are also reported. Mean $\pm$ std across 5 random seeds.}
  \label{tab:result_internal_best}
  \begin{adjustbox}{max width=\linewidth, max totalheight=0.85\textheight, keepaspectratio}
\begin{tabular}{ll r rr rr r}
    \toprule
    & & & \multicolumn{2}{c}{Train} & \multicolumn{2}{c}{Test} & \\
    \cmidrule(lr){4-5}\cmidrule(lr){6-7}
    Split & Model & $n$ & RMSE & MPD & RMSE & MPD & Gap \\
    \midrule
  \multirow{3}{*}{Random} & GraphCCS & \multirow{3}{*}{922} & 3.287 $\pm$ 0.460 & 1.150 $\pm$ 0.171 & 4.858 $\pm$ 0.112 & 1.766 $\pm$ 0.050 & 1.571 $\pm$ 0.494 \\
   & SigmaCCS &  & 4.664 $\pm$ 0.390 & 1.728 $\pm$ 0.102 & 5.137 $\pm$ 0.166 & 1.950 $\pm$ 0.051 & 0.473 $\pm$ 0.251 \\
   & \ModelName{} (Single) &  & 4.556 $\pm$ 0.098 & 0.996 $\pm$ 0.034 & 4.637 $\pm$ 0.047 & \textbf{1.670} $\pm$ 0.023 & 0.081 $\pm$ 0.085 \\
    \midrule
  \multirow{3}{*}{Scaffold} & GraphCCS & \multirow{3}{*}{920} & 3.844 $\pm$ 0.921 & 1.357 $\pm$ 0.299 & 6.396 $\pm$ 0.282 & 2.278 $\pm$ 0.182 & 2.552 $\pm$ 0.718 \\
   & SigmaCCS &  & 6.771 $\pm$ 1.183 & 2.534 $\pm$ 0.450 & 6.867 $\pm$ 0.693 & 2.795 $\pm$ 0.416 & 0.096 $\pm$ 0.696 \\
   & \ModelName{} (Uniform) &  & 5.270 $\pm$ 0.567 & 1.317 $\pm$ 0.244 & 6.540 $\pm$ 0.095 & \textbf{2.113} $\pm$ 0.042 & 1.270 $\pm$ 0.528 \\
    \midrule
  \multirow{3}{*}{Adduct-sens.} & GraphCCS & \multirow{3}{*}{1382} & 4.117 $\pm$ 0.400 & 1.373 $\pm$ 0.086 & 6.671 $\pm$ 0.076 & 2.512 $\pm$ 0.038 & 2.554 $\pm$ 0.385 \\
   & SigmaCCS &  & 5.263 $\pm$ 0.916 & 2.099 $\pm$ 0.379 & 6.918 $\pm$ 0.280 & 2.756 $\pm$ 0.193 & 1.655 $\pm$ 0.783 \\
   & \ModelName{} (Single) &  & 4.780 $\pm$ 0.161 & 1.023 $\pm$ 0.054 & 6.341 $\pm$ 0.076 & \textbf{2.363} $\pm$ 0.032 & 1.562 $\pm$ 0.113 \\
    \bottomrule
  \end{tabular}
\end{adjustbox}
\end{table}

\subsection{Conformer Pooling and Fusion Strategy}

\begin{figure}[H]
  \centering
  \includegraphics[width=\textwidth]{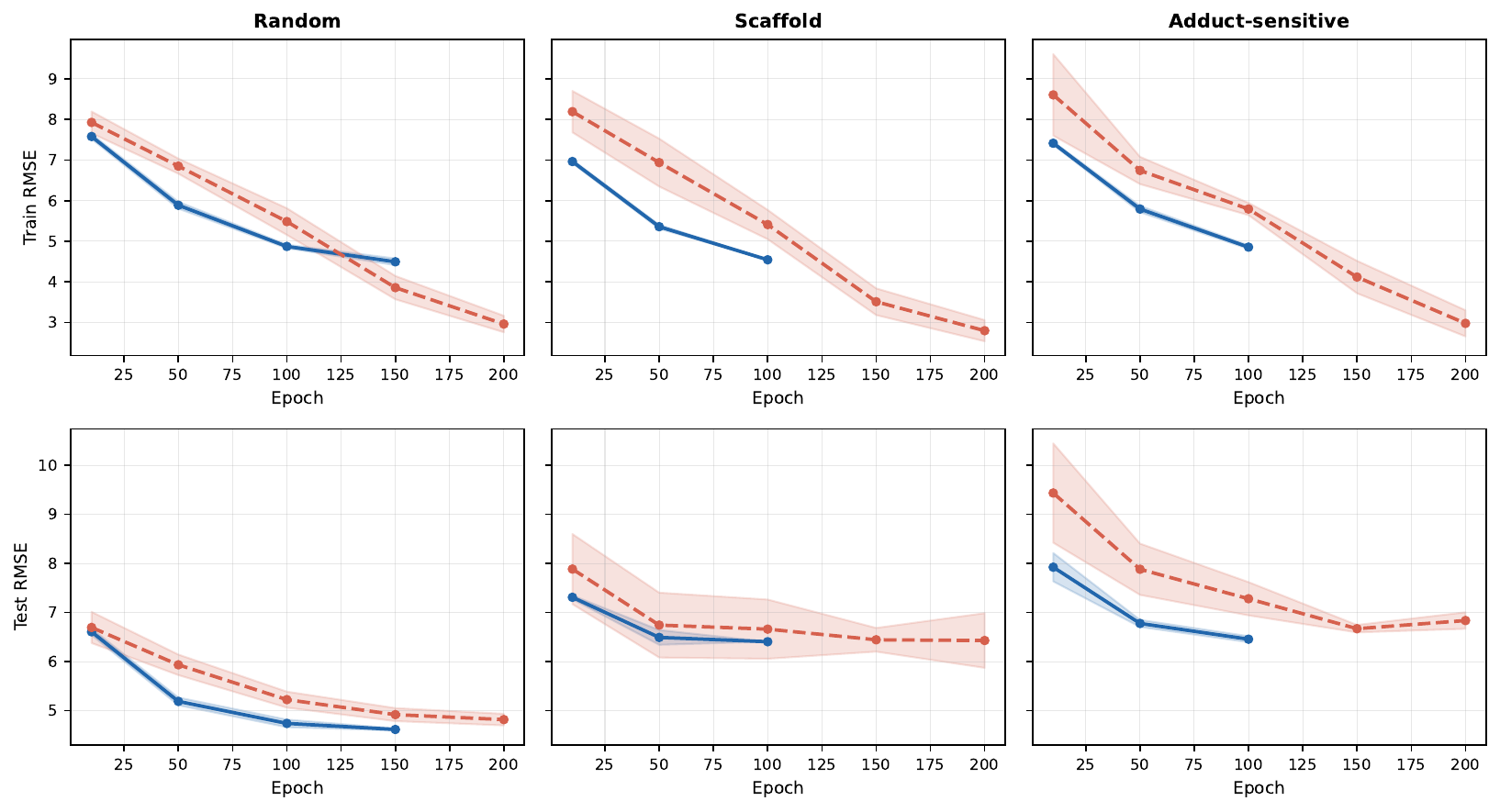}
  \caption{Train and test RMSE as a function of training epoch for \ModelName{}
    (best conformer pooling per split; solid blue) and GraphCCS (dashed red).
    Each panel corresponds to one training split strategy: Random ($n_\text{train}=7{,}374$),
    Scaffold ($n_\text{train}=7{,}369$), and Adduct-sensitive ($n_\text{train}=6{,}446$).
    Shaded bands show $\pm$1 standard deviation across 5 random seeds.
    \ModelName{} uses Single-conformer pooling on the Random and Adduct-sensitive splits,
    and Uniform pooling on the Scaffold split (best-MPD selection).
    Both models are evaluated at epochs 10, 50, 100, 150, and 200;
    \ModelName{} additionally records the best-validation checkpoint.
    GraphCCS trains for 200 epochs; \ModelName{} uses early stopping and typically terminates before epoch 200.
}
  \label{fig:learning-curves}
\end{figure}

On the random and scaffold splits all pooling modes achieve similar RMSE; the
differences are within one standard deviation across seeds.
Multi-conformer pooling does not improve results on the adduct-sensitive split, where a single conformer
outperforms Boltzmann and uniform pooling (Table~\ref{tab:ccs3d_graphccs_main}).
Learned attention pooling performs consistently worst across all splits despite having more
parameters, overfitting on the $\sim$9{,}000 \Dataset{} training molecules available.
Early fusion consistently improves over late fusion across all splits; 
\ifpreprint
the diagnostic
evidence is in Section~\ref{sec:discussion}.
\else
the diagnostic
evidence is in Discussion.
\fi
Figure~\ref{fig:learning-curves} shows train and test RMSE across epochs for \ModelName
and GraphCCS; the generalization gap for \ModelName is substantially smaller throughout
training.

\subsection{\Dataset{} Evaluation}

Table~\ref{tab:result_internal_best} compares the best \ModelName configuration per split
against GraphCCS and SigmaCCS.
On the random split, \ModelName (single conformer) achieves
$(4.637 \pm 0.047)$~\AA$^2$ RMSE and MPD $(1.670 \pm 0.023)$\% versus
GraphCCS at $(4.858 \pm 0.112)$~\AA$^2$ and $(1.766 \pm 0.050)$\%, and
SigmaCCS at $(5.137 \pm 0.166)$~\AA$^2$ and $(1.950 \pm 0.051)$\%.
On the scaffold split, \ModelName{} (uniform) reaches MPD $(2.113 \pm 0.042)$\% versus
GraphCCS at $(2.278 \pm 0.182)$\%; RMSE on this split is $(6.540 \pm 0.095)$~\AA$^2$
versus $(6.396 \pm 0.303)$~\AA$^2$ for GraphCCS, with overlapping uncertainty intervals.
On the adduct-sensitive split, \ModelName (single) achieves MPD $(2.363 \pm 0.032)$\%
versus GraphCCS at $(2.512 \pm 0.038)$\%.

The generalization gap (test RMSE minus train RMSE at the best-validation checkpoint)
is consistently lower for \ModelName than for GraphCCS.
On the random split, \ModelName's gap is $(0.081 \pm 0.085)$~\AA$^2$ versus
$(1.571 \pm 0.494)$~\AA$^2$ for GraphCCS, a 19-fold reduction.
This reduction is consistent with the regularizing effect of residual learning and
zero-initialized residual prediction, although differences in model capacity and training
dynamics may also contribute. The backbone does not need to re-learn the dominant mass correlation from scratch,
so it does not overfit to training-set size and adduct patterns before generalizing.  All results for tested \ModelName{} configurations and splits are given in Table~\ref{tab:result_internal_full}.

\subsection{Evaluation on External Datasets}

\begin{figure}[H]
  \centering
  \includegraphics[width=\textwidth]{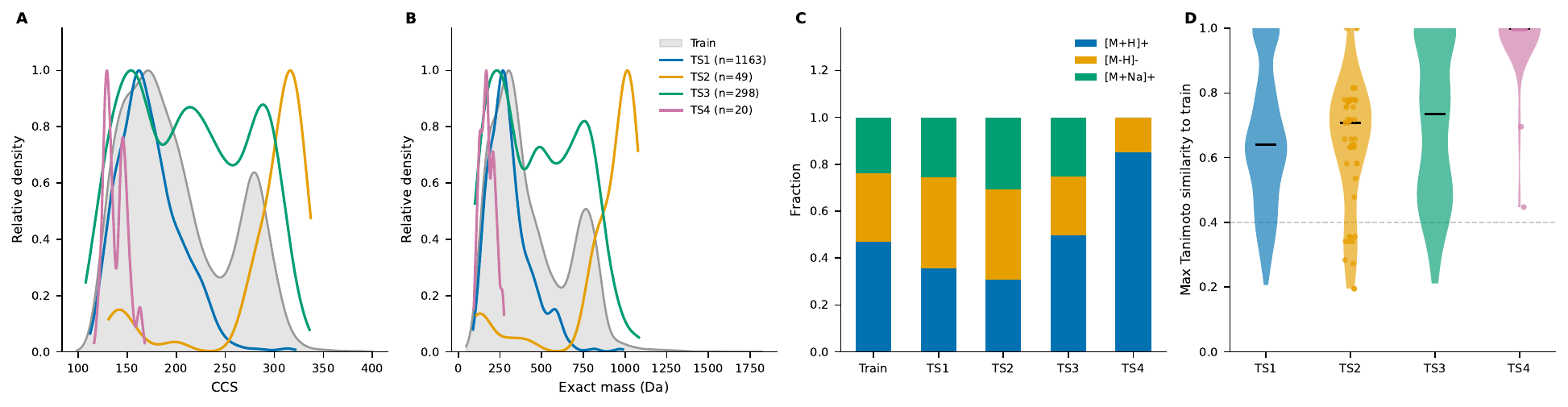}
  \caption{Characterization of the four external test sets relative to the
    training distribution ($n = 9{,}209$).
    \textbf{(A)}~CCS distribution (kernel density estimate): TS1--3 span
    nearly the full training CCS range; TS4 (amino acids and metabolites)
    occupies a narrow low-CCS band (121--163).
    \textbf{(B)}~Exact molecular mass distribution: TS4 molecules are
    consistently lighter than the typical training molecule, while TS1--3
    overlap well with training.
    \textbf{(C)}~Adduct composition: Train, TS1, and TS3 are
    \mbox{[M+H]$^+$}-dominant and carry all three adducts in broadly similar
    proportions.
    TS2 is the only set where \mbox{[M$-$H]$^-$} is the most abundant adduct
    ($\approx$39\%), with \mbox{[M+H]$^+$} and \mbox{[M+Na]$^+$} each at
    $\approx$31\%, a notably different balance from the training distribution.
    TS4 contains no \mbox{[M+Na]$^+$} entries.
    \textbf{(D)}~Chemical novelty quantified as the maximum ECFP4 Tanimoto
    similarity of each testset molecule to any training molecule
    (violin\,+\,individual points for $n \leq 60$; dashed line at 0.4).
    TS1 and TS3 show broad novelty distributions with medians
    below 0.4, confirming genuine out-of-distribution generalization;
    TS2 and TS4 are small but similarly novel.}
  \label{fig:testset-comparison}
\end{figure}

Figure~\ref{fig:testset-comparison} characterizes the four external test sets relative to
the training distribution.
TS1 ($n=1{,}163$) and TS3 ($n=298$) broadly overlap the training chemical space but
contain novel scaffolds (median Tanimoto similarity below 0.4).
TS2 ($n=49$) is the only set dominated by [\textit{M}$-$H]\textsuperscript{$-$} ions
($\approx$39\%), with a different adduct balance from training (Figure~\ref{fig:testset-comparison}C).
TS3 ($n=298$) is relatively enriched in molecules with higher mass and CCS compared with
the training set (Figure~\ref{fig:testset-comparison}A--B).
GPCL (TS4, $n=20$, amino acids and metabolites) occupies a narrow low-mass, low-CCS band
(121--163) not well-represented in training.

Table~\ref{tab:result_external_best} reports the best observed configuration per external
test set, while Table \ref{tab:result_external_full} report all training-split and pooling combinations.
Across the 12 external testset--training-split combinations in Table \ref{tab:main_external}, \ModelName{}
achieves the lowest RMSE and MPD in 11 cases; the only exception is TS2 under the random
split, where all models perform substantially worse and SigmaCCS is marginally best.
Thus, although Table~\ref{tab:result_external_best} should be interpreted as a
best-configuration analysis rather than a blinded fixed-configuration benchmark, the full
supplementary results show that the external gains are not driven by a single
cherry-picked split.

\begin{table}[H]
  \centering
  % Old: External test set: best model configuration per testset. Training split as config param.
  \caption{Summary of best-observed external performance across four independent test sets.
    For each model and test set, the best training split is reported; for \ModelName{},
    conformer pooling mode and training split are jointly selected.
    This table should be interpreted as a best-configuration analysis; full results across
    all splits are in Tables~\ref{tab:result_external_full} and~\ref{tab:main_external}.
    Mean $\pm$ std across 5 random seeds.
    $^*$Testset~4: the molecule \texttt{[H]/N=C(/N)NCC(=O)O} contains an explicit hydrogen
    absent from the SigmaCCS atom vocabulary; SigmaCCS is evaluated on 19 of the 20
    molecules ($n = 19$).}
  \label{tab:result_external_best}
  \begin{adjustbox}{max width=\linewidth, max totalheight=0.85\textheight, keepaspectratio}
\begin{tabular}{ll l r rr}
    \toprule
    & & & & \multicolumn{2}{c}{Test} \\
    \cmidrule(lr){5-6}
    Testset & Model & Best Config & $n$ & RMSE & MPD \\
    \midrule
  \multirow{3}{*}{Testset 1} & GraphCCS & Adduct-sens. & 1163 & 5.504 $\pm$ 0.079 & 2.280 $\pm$ 0.046 \\
   & SigmaCCS & Random & 1163 & 5.630 $\pm$ 0.209 & 2.342 $\pm$ 0.110 \\
   & \ModelName{} (Single) & Single, Random & 1163 & 5.467 $\pm$ 0.108 & \textbf{2.138} $\pm$ 0.035 \\
    \midrule
  \multirow{3}{*}{Testset 2} & GraphCCS & Random & 49 & 11.439 $\pm$ 1.142 & 3.390 $\pm$ 0.278 \\
   & SigmaCCS & Random & 49 & 11.420 $\pm$ 0.725 & 3.386 $\pm$ 0.304 \\
   & \ModelName{} (Single) & Single, Adduct-sens. & 49 & 10.641 $\pm$ 0.785 & \textbf{3.021} $\pm$ 0.141 \\
    \midrule
  \multirow{3}{*}{Testset 3} & GraphCCS & Random & 298 & 6.587 $\pm$ 0.410 & 2.249 $\pm$ 0.045 \\
   & SigmaCCS & Random & 298 & 9.042 $\pm$ 1.443 & 2.819 $\pm$ 0.197 \\
   & \ModelName{} (Learned) & Learned, Scaffold & 298 & 6.127 $\pm$ 0.283 & \textbf{2.150} $\pm$ 0.082 \\
    \midrule
  \multirow{3}{*}{Testset 4\textsuperscript{*}} & GraphCCS & Random & 20 & 3.193 $\pm$ 0.268 & 1.798 $\pm$ 0.184 \\
   & SigmaCCS & Random & 19 & 3.623 $\pm$ 0.279 & 1.836 $\pm$ 0.145 \\
   & \ModelName{} (Single) & Single, Random & 20 & 2.268 $\pm$ 0.095 & \textbf{1.340} $\pm$ 0.066 \\
    \bottomrule
  \end{tabular}
\end{adjustbox}
\end{table}

\subsection{Prospective Experimental Validation}

Computing CCS from first principles involves building a conformer ensemble, assigning
charges via quantum chemistry, and running trajectory-method simulations.
The cost of this pipeline grows steeply with molecular size and flexibility, and negatively
charged adducts are harder to converge than protonated
species~\cite{das2022silico, borges2021quantum, keng2024eliminating, st2020quantum}.
The 20 GPCL molecules~\cite{das2024molecular} used here skew toward small, singly protonated
metabolites with at most 7 rotatable bonds, precisely the regime where physics-based
methods are most competitive, so this is a conservative test for \ModelName{}.

\begin{figure}[H]
  \centering
  \includegraphics[width=0.6\textwidth]{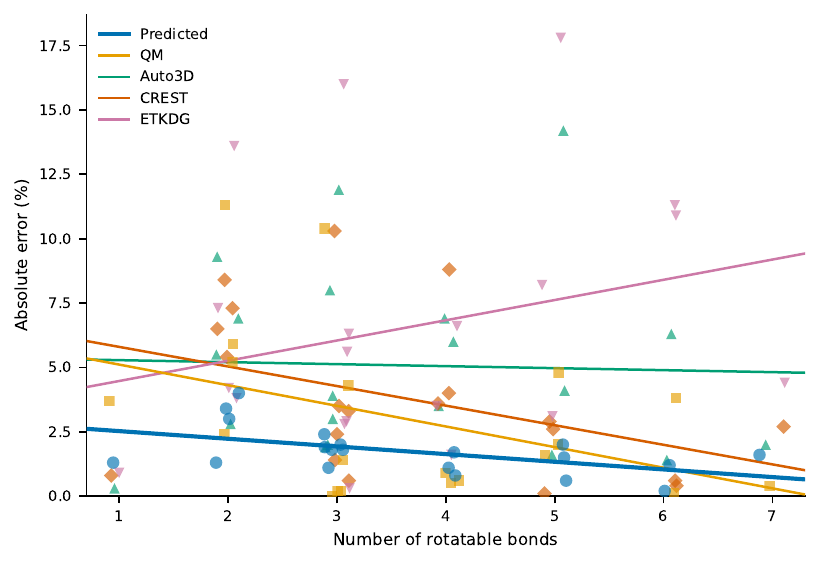}
  \caption{Absolute percentage error as a function of the number of rotatable bonds for \ModelName{} (clean trainset, blue) and four physics-based CCS estimation approaches on Testset~4 ($n=20$). Values for the physics-based approaches (QM, Auto3D, CREST, ETKDG) are taken from \cite{das2022silico}. Lines show per-method linear regression; points are jittered horizontally for visibility. \ModelName{} maintains consistently low error across the full range of molecular flexibility (mean 1.7\%, max 4.0\%, slope $-0.30$\,\%/bond, $p=0.020$). Physics-based CCS estimation methods show substantially higher errors at all flexibility levels (mean 4.6\%, max 17.8\%), with wider molecule-to-molecule variation.}
  \label{fig:rotbond-scaling}
\end{figure}

Table~\ref{tab:gpcl_comparison} compares \ModelName{} against four physics-based workflows
on GPCL (TS4)~\cite{das2022silico}.
\ModelName{} was retrained with all 20 GPCL molecules withheld and
achieves RMSE $2.68$~\AA$^2$ and MPD~1.74\%, with lower RMSE than all four published
estimates on these 20 molecules
(QM: $6.22$, Auto3D: $8.57$, CREST: $6.68$, ETKDG: $11.68$~\AA$^2$).
Figure~\ref{fig:rotbond-scaling} shows that \ModelName{} maintains low error across the
full range of rotatable bonds, with a slight improvement at higher flexibility
(mean 1.7\%, slope $-0.30$\%/bond, $p=0.020$).
On this small set, physics-based methods show higher errors and wider molecule-to-molecule
variation, though the sample size limits generalization of this trend.
Where conformational complexity makes physics-based calculation most expensive,
\ModelName{} prediction cost stays constant.

\begin{table}[H]
  \centering
  \caption{CCS predictions for the 20 GPCL compounds (Testset~4, amino acids and metabolites) from \ModelName{} and four physics-based CCS estimation approaches. \ModelName{} was retrained with all 20 molecules withheld (RMSE\,=\,2.68~\AA$^2$, MPD\,=\,1.74\%). Values for the physics-based approaches (QM, Auto3D, CREST, ETKDG) are taken from \cite{das2022silico}. Each method column shows predicted CCS with absolute percentage error in parentheses.}
  \label{tab:gpcl_comparison}
  \begin{adjustbox}{max width=\linewidth, keepaspectratio}
\begin{tabular}{ll r r r r r r}
  \toprule
  Compound & Adduct & Exp. & \multicolumn{5}{c}{Pred.\ (\%err)} \\
  \cmidrule(lr){4-8}
  & & & \ModelName{} & QM & Auto3D & CREST & ETKDG \\
  \midrule
  L-tryptophan          & [M+H]+ & 144.7 & 147.5\,(2.0) & 159.7\,(10.4) & 162.0\,(11.9) & 159.6\,(10.3) & 167.8\,(16.0) \\
  Nicotinic-Acid        & [M+H]+ & 127.5 & 125.8\,(1.3) & 132.2\,(3.7)  & 127.2\,(0.3)  & 126.5\,(0.8)  & 128.6\,(0.9)  \\
  Quinolinic-Acid       & [M+H]+ & 135.1 & 133.4\,(1.3) & 142.1\,(5.2)  & 138.8\,(2.8)  & 143.9\,(6.5)  & 140.8\,(4.2)  \\
  L-asparagine          & [M+H]+ & 128.7 & 125.6\,(2.4) & 128.9\,(0.2)  & 124.9\,(3.0)  & 124.2\,(3.5)  & 125.1\,(2.8)  \\
  N-methyl-L-glutamate  & [M+H]+ & 131.9 & 134.6\,(2.0) & 129.3\,(2.0)  & 134.0\,(1.6)  & 128.0\,(2.9)  & 135.9\,(3.1)  \\
  Serotonin             & [M+H]+ & 148.1 & 142.2\,(4.0) & 131.4\,(11.3) & 158.3\,(6.9)  & 158.9\,(7.3)  & 168.2\,(13.6) \\
  Glutamine             & [M+H]+ & 130.7 & 129.3\,(1.1) & 129.9\,(0.6)  & 126.1\,(3.5)  & 126.0\,(3.6)  & 132.8\,(1.6)  \\
  L-anserine            & [M+H]+ & 153.9 & 155.7\,(1.2) & 159.7\,(3.8)  & 156.0\,(1.4)  & 152.9\,(0.6)  & 171.2\,(11.3) \\
  NN-Dimethylglycine    & [M+H]+ & 125.6 & 121.8\,(3.0) & 118.2\,(5.9)  & 113.9\,(9.3)  & 115.0\,(8.4)  & 116.4\,(7.3)  \\
  L-citrulline          & [M-H]- & 135.4 & 134.6\,(0.6) & 141.9\,(4.8)  & 154.7\,(14.2) & 135.6\,(0.1)  & 159.5\,(17.8) \\
  Kynurenine            & [M+H]+ & 147.7 & 150.2\,(1.7) & 146.9\,(0.5)  & 156.6\,(6.0)  & 160.7\,(8.8)  & 157.4\,(6.6)  \\
  O-succinyl-L-homoserine & [M+H]+ & 144.9 & 147.2\,(1.6) & 145.4\,(0.4) & 147.9\,(2.0)  & 141.0\,(2.7)  & 138.5\,(4.4)  \\
  L-2-AminoadipicAcid   & [M+H]+ & 131.6 & 133.6\,(1.5) & 129.6\,(1.6)  & 137.0\,(4.1)  & 128.2\,(2.6)  & 142.4\,(8.2)  \\
  L-mimosine            & [M+H]+ & 143.3 & 146.0\,(1.9) & 145.4\,(1.4)  & 146.0\,(1.9)  & 142.5\,(0.6)  & 147.5\,(2.9)  \\
  Citramalic-Acid       & [M-H]- & 121.3 & 119.1\,(1.8) & 121.6\,(0.2)  & 116.5\,(3.9)  & 118.4\,(2.4)  & 120.9\,(0.3)  \\
  Abscisic-Acid         & [M+H]+ & 162.8 & 161.0\,(1.1) & 162.9\,(0.0)  & 166.1\,(2.0)  & 160.6\,(1.4)  & 173.1\,(6.3)  \\
  L-ornithine           & [M+H]+ & 128.6 & 129.6\,(0.8) & 127.4\,(0.9)  & 119.8\,(6.9)  & 123.5\,(4.0)  & 133.1\,(3.5)  \\
  Carnosine             & [M+H]+ & 150.1 & 150.4\,(0.2) & 150.2\,(0.1)  & 159.6\,(6.3)  & 149.5\,(0.4)  & 166.5\,(10.9) \\
  L-tyrosine            & [M-H]- & 142.4 & 139.8\,(1.8) & 148.5\,(4.3)  & 153.8\,(8.0)  & 147.1\,(3.3)  & 150.4\,(5.6)  \\
  Guanidinoacetic-Acid  & [M+H]+ & 127.1 & 122.8\,(3.4) & 130.2\,(2.4)  & 120.2\,(5.5)  & 120.2\,(5.4)  & 122.3\,(3.8)  \\
  \midrule
  \multicolumn{3}{r}{\textit{RMSE}}     & \textit{2.68} & \textit{6.22} & \textit{8.57} & \textit{6.68} & \textit{11.68} \\
  \multicolumn{3}{r}{\textit{MPD (\%)}} & \textit{1.74} & \textit{2.98} & \textit{5.08} & \textit{3.78} & \textit{6.55}  \\
  \bottomrule
\end{tabular}
  \end{adjustbox}
\end{table}

\section{Discussion}
\label{sec:discussion}

\subsection{Generalization Gains from Residual Learning and Early Fusion}

On the random split, \ModelName{} achieves nearly identical training and test RMSE
($4.56$ and $4.64$~\AA$^2$, gap $0.08$~\AA$^2$), while GraphCCS trains to
$3.29$~\AA$^2$ but tests at $4.86$~\AA$^2$ (gap $1.57$~\AA$^2$), a 19-fold
difference in generalization gap.
Two mechanisms likely contribute to this gap.
The residual objective removes the dominant mass-CCS correlation (Ridge $R^2 = 0.961$)
before training, substantially reducing the risk of overfitting to a signal trivially
predictable from molecular weight.
Injecting the adduct token into the geometry encoder backbone (early fusion) compounds the
effect, allowing the model to learn adduct-specific structural representations from the
outset rather than recovering adduct sensitivity from raw CCS targets at the head.
GraphCCS must learn the mass correlation implicitly from its per-atom feature graph while
simultaneously learning the residual structural signal; the larger gap suggests this joint
objective is less stable across the train-to-test distribution shift, though contributions
from differences in model capacity or regularization cannot be ruled out.
\ifpreprint
Section~\ref{sec:supp_lf_ef} quantifies the contribution of each component: the
late-fusion model with residual learning but without early fusion achieves a gap of
$3.00$~\AA$^2$ on the random split, 37-fold larger than \ModelName{}, with early fusion as
the primary driver of the generalization improvement.
\else 
The late-fusion model with residual learning but without early fusion achieves a gap of
$3.00$~\AA$^2$ on the random split, 37-fold larger than \ModelName{}, with early fusion as
the primary driver of the generalization improvement.  See Supporting Information for more details.
\fi

\subsection{Stereoisomer Leakage} \label{sec:steroisomerleakage}

During dataset curation, stereoisomers are treated as distinct molecules.  However, while CCS measurements yield distinct values ($\Delta \text{CCS} \approx 7\%$) for diastereomers~\cite{hofmann2015identification}, they are identical for enantiomers when experiments are performed with commonly used achiral buffer gases, although specialized IMS can separate enantiomers~\cite{dwivedi2006gas}. To assess whether stereoisomer leakage between training and test sets inflates reported
performance, we identified stereoisomeric pairs by stripping chirality annotations from each
SMILES and grouping molecules with identical connectivity.
A test molecule leaks if a stereoisomeric sibling appears in training; we re-evaluated
accuracy on the remaining clean subset after removing such molecules.
Leakage rates are 20.1\%, 14.1\%, and 6.9\% for the random, scaffold, and adduct-sensitive
splits, respectively.
Removing leaked molecules changes \ModelName{} RMSE by at most $+0.19$~\AA$^2$ and
GraphCCS RMSE by at most $+0.24$~\AA$^2$ \ifpreprint
(Section~\ref{sec:supp_leakage}),
confirming that reported results are not materially inflated by leakage.
\else 
(Supporting Information),
confirming that reported results are not materially inflated by leakage.
\fi

\subsection{Effect of Multi-Conformer Pooling on the Adduct-Sensitive Split}

\begin{table}[H]
  \centering
  % Old: \ModelName{} (residual, all conformer pooling modes) vs.\ GraphCCS. Train RMSE, Train MPD, Test RMSE, Test MPD, Pearson~$R$, Spearman~$\rho$, Kendall~$\tau$, and generalization gap (Gap = test$-$train RMSE at the same best-val checkpoint). Mean $\pm$ std across 5 seeds.
  \caption{\ModelName{} matches or outperforms GraphCCS on the \Dataset{} test set across all three training splits. \ModelName{} predicts the residual from a linear mass-adduct baseline using a pretrained 3D geometry encoder; four conformer pooling strategies are compared: Single (one RDKit conformer), Uniform (unweighted ensemble mean), Boltzmann (MMFF94 energy-weighted), and Learned (attention-weighted). Generalization gap: Gap $=$ test $-$ train RMSE at the nearest best-validation checkpoint. Mean $\pm$ std across 5 random seeds.}
  \label{tab:ccs3d_graphccs_main}
  \begin{adjustbox}{max width=\linewidth, max totalheight=0.85\textheight, keepaspectratio}
\begin{tabular}{llllrrrrrrrr}
    \toprule
    Split & Model & Pooling & $n$ & Train RMSE & Train MPD & Test RMSE & Test MPD & Pearson $R$ & Spearman $\rho$ & Kendall $\tau$ & Gap \\
    \midrule
  \multirow{5}{*}{Random} & GraphCCS & — & \multirow{5}{*}{922} & 3.287 $\pm$ 0.460 & 1.150 $\pm$ 0.171 & 4.858 $\pm$ 0.112 & 1.766 $\pm$ 0.050 & 0.9960 $\pm$ 0.0002 & 0.9938 $\pm$ 0.0003 & 0.9399 $\pm$ 0.0016 & 1.571 $\pm$ 0.494 \\
   & \ModelName{} & Single &  & 4.556 $\pm$ 0.098 & 0.996 $\pm$ 0.034 & 4.637 $\pm$ 0.047 & \textbf{1.670} $\pm$ 0.023 & 0.9963 $\pm$ 0.0001 & 0.9946 $\pm$ 0.0002 & 0.9431 $\pm$ 0.0009 & 0.081 $\pm$ 0.085 \\
   & \ModelName{} & Uniform &  & 4.611 $\pm$ 0.242 & 1.073 $\pm$ 0.074 & 4.750 $\pm$ 0.057 & 1.700 $\pm$ 0.009 & 0.9962 $\pm$ 0.0001 & 0.9943 $\pm$ 0.0001 & 0.9421 $\pm$ 0.0006 & 0.139 $\pm$ 0.263 \\
   & \ModelName{} & Boltzmann &  & 4.430 $\pm$ 0.131 & 1.007 $\pm$ 0.050 & 4.711 $\pm$ 0.031 & 1.715 $\pm$ 0.009 & 0.9962 $\pm$ 0.0000 & 0.9945 $\pm$ 0.0001 & 0.9421 $\pm$ 0.0002 & 0.281 $\pm$ 0.155 \\
   & \ModelName{} & Learned &  & 4.597 $\pm$ 0.216 & 1.079 $\pm$ 0.084 & 4.705 $\pm$ 0.077 & 1.694 $\pm$ 0.017 & 0.9962 $\pm$ 0.0001 & 0.9944 $\pm$ 0.0002 & 0.9423 $\pm$ 0.0007 & 0.108 $\pm$ 0.207 \\
    \midrule
  \multirow{5}{*}{Scaffold} & GraphCCS & — & \multirow{5}{*}{920} & 3.844 $\pm$ 0.921 & 1.357 $\pm$ 0.299 & 6.396 $\pm$ 0.282 & 2.278 $\pm$ 0.182 & 0.9925 $\pm$ 0.0004 & 0.9907 $\pm$ 0.0003 & 0.9235 $\pm$ 0.0023 & 2.552 $\pm$ 0.718 \\
   & \ModelName{} & Single &  & 5.190 $\pm$ 0.312 & 1.254 $\pm$ 0.144 & 6.539 $\pm$ 0.080 & 2.116 $\pm$ 0.022 & 0.9920 $\pm$ 0.0002 & 0.9912 $\pm$ 0.0002 & 0.9258 $\pm$ 0.0011 & 1.349 $\pm$ 0.266 \\
   & \ModelName{} & Uniform &  & 5.270 $\pm$ 0.567 & 1.317 $\pm$ 0.244 & 6.540 $\pm$ 0.095 & \textbf{2.113} $\pm$ 0.042 & 0.9920 $\pm$ 0.0002 & 0.9915 $\pm$ 0.0003 & 0.9271 $\pm$ 0.0014 & 1.270 $\pm$ 0.528 \\
   & \ModelName{} & Boltzmann &  & 5.329 $\pm$ 0.723 & 1.348 $\pm$ 0.329 & 6.534 $\pm$ 0.184 & 2.144 $\pm$ 0.083 & 0.9920 $\pm$ 0.0004 & 0.9914 $\pm$ 0.0006 & 0.9263 $\pm$ 0.0028 & 1.205 $\pm$ 0.577 \\
   & \ModelName{} & Learned &  & 5.269 $\pm$ 0.435 & 1.329 $\pm$ 0.179 & 6.552 $\pm$ 0.121 & 2.134 $\pm$ 0.076 & 0.9922 $\pm$ 0.0002 & 0.9916 $\pm$ 0.0002 & 0.9279 $\pm$ 0.0010 & 1.283 $\pm$ 0.381 \\
    \midrule
  \multirow{5}{*}{Adduct-sens.} & GraphCCS & — & \multirow{5}{*}{1382} & 4.117 $\pm$ 0.400 & 1.373 $\pm$ 0.086 & 6.671 $\pm$ 0.076 & 2.512 $\pm$ 0.038 & 0.9911 $\pm$ 0.0004 & 0.9875 $\pm$ 0.0005 & 0.9105 $\pm$ 0.0021 & 2.554 $\pm$ 0.385 \\
   & \ModelName{} & Single &  & 4.780 $\pm$ 0.161 & 1.023 $\pm$ 0.054 & 6.341 $\pm$ 0.076 & \textbf{2.363} $\pm$ 0.032 & 0.9919 $\pm$ 0.0002 & 0.9887 $\pm$ 0.0003 & 0.9149 $\pm$ 0.0012 & 1.562 $\pm$ 0.113 \\
   & \ModelName{} & Uniform &  & 4.989 $\pm$ 0.234 & 1.169 $\pm$ 0.074 & 6.476 $\pm$ 0.127 & 2.425 $\pm$ 0.057 & 0.9914 $\pm$ 0.0002 & 0.9878 $\pm$ 0.0003 & 0.9117 $\pm$ 0.0015 & 1.487 $\pm$ 0.144 \\
   & \ModelName{} & Boltzmann &  & 4.790 $\pm$ 0.210 & 1.056 $\pm$ 0.067 & 6.583 $\pm$ 0.034 & 2.458 $\pm$ 0.015 & 0.9912 $\pm$ 0.0001 & 0.9878 $\pm$ 0.0003 & 0.9114 $\pm$ 0.0008 & 1.793 $\pm$ 0.187 \\
   & \ModelName{} & Learned &  & 5.286 $\pm$ 0.395 & 1.321 $\pm$ 0.198 & 6.446 $\pm$ 0.088 & 2.429 $\pm$ 0.062 & 0.9914 $\pm$ 0.0003 & 0.9876 $\pm$ 0.0004 & 0.9113 $\pm$ 0.0020 & 1.160 $\pm$ 0.337 \\
    \bottomrule
  \end{tabular}
\end{adjustbox}
\end{table}

On the adduct-sensitive split, single-conformer pooling outperforms all multi-conformer
strategies (Table~\ref{tab:ccs3d_graphccs_main}): Single $6.34$ $<$ Learned $6.45$ $<$
Uniform $6.48$ $<$ Boltzmann $6.58$~\AA$^2$, with Boltzmann performing worst.
The ordering within multi-conformer modes is informative: equal-weight averaging over the
same ten neutral conformers is less harmful than Boltzmann-weighting them, implicating the weighting scheme rather
than multi-conformer averaging per se.
Boltzmann weights derived from MMFF94 energies approximate the thermal population of the
neutral molecule in vacuum, whereas the measured CCS is that of the adducted ion.
The ionic energy landscape differs from the neutral one: the adduct charge site stabilises
conformers with favourable intramolecular charge solvation while destabilising others, so
the neutral Boltzmann ensemble mixes in conformers the ion does not occupy.
A single representative ETKDG+MMFF neutral conformer avoids this noise because it does not
actively upweight geometries that are wrong for the ion.

This finding is specific to the adduct-sensitive split: on the scaffold split, where the
test set does not specifically challenge adduct generalisation, uniform averaging over ten
conformers marginally outperforms single-conformer pooling (MPD $2.11$\% vs.\ $2.21$\%),
confirming that multi-conformer diversity is beneficial in general and that the failure mode
is the Boltzmann weighting scheme applied to neutral-molecule energies.
Ion-specific conformer generation, optimising and weighting conformers after adduct
formation using an ionic force field or semiempirical QM, is a natural direction for
future improvement on this split.

\subsection{Adduct Conditioning Using Late Fusion}
\label{sec:late_fusion}

We initially implemented a late-fusion strategy as a design baseline before settling on early
fusion in \ModelName{}.
In the late-fusion baseline, the geometry encoder backbone receives no adduct information.
Two token representations are compared under late fusion.
In the \emph{CLS} variant, the 3-dimensional adduct one-hot is concatenated to the pooled
CLS embedding immediately before the MLP head (515-dimensional input).
In the \emph{Gasteiger} variant, RDKit Gasteiger partial charges serve as attention bias
weights over the per-atom embeddings to produce a 512-dimensional charge-biased atom-pool
embedding; for cationic adducts atoms with more negative charge receive higher weight, with
weights reversed for the anionic adduct.
A learnable scalar $\lambda$ (initialized to zero) gates the charge bias, and the atom-pool
embedding is concatenated to the CLS embedding and adduct one-hot (1027-dimensional input).
This design isolates the contribution of encoder-level adduct conditioning in \ModelName{}.
Across all configurations and splits, late fusion performs worse than early fusion, with the
gap widest on the adduct-sensitive split (best late-fusion RMSE 7.96~\AA$^2$ vs.\
\ModelName{} 6.34~\AA$^2$), where the test set is specifically constructed to challenge
adduct generalisation.
On the random split the best late-fusion configuration (Uniform pooling, Gasteiger token)
reaches 5.30~\AA$^2$ RMSE, compared to 4.64~\AA$^2$ for early fusion.
Results for all late-fusion pooling modes and token representations are given in
\ifpreprint
Section~\ref{sec:supp_late_fusion}
\else
Supporting Information
\fi
(Table~\ref{tab:late_fusion_ablation}).
A focused three-way comparison of late fusion (LF), late fusion with residual learning
(LF+Residual), and early fusion (EF) is given in
\ifpreprint
Section~\ref{sec:supp_lf_ef} 
\else
Supporting Information
\fi
(Table~\ref{tab:lf-ef-ablation}).
The consistent advantage of early fusion suggests that injecting adduct identity into the
encoder, rather than appending it downstream, matters for learning adduct-specific
structural features.

\subsection{Geometry-Stratified Performance Analysis}

\begin{figure}[H]
  \centering
  \includegraphics[width=\textwidth]{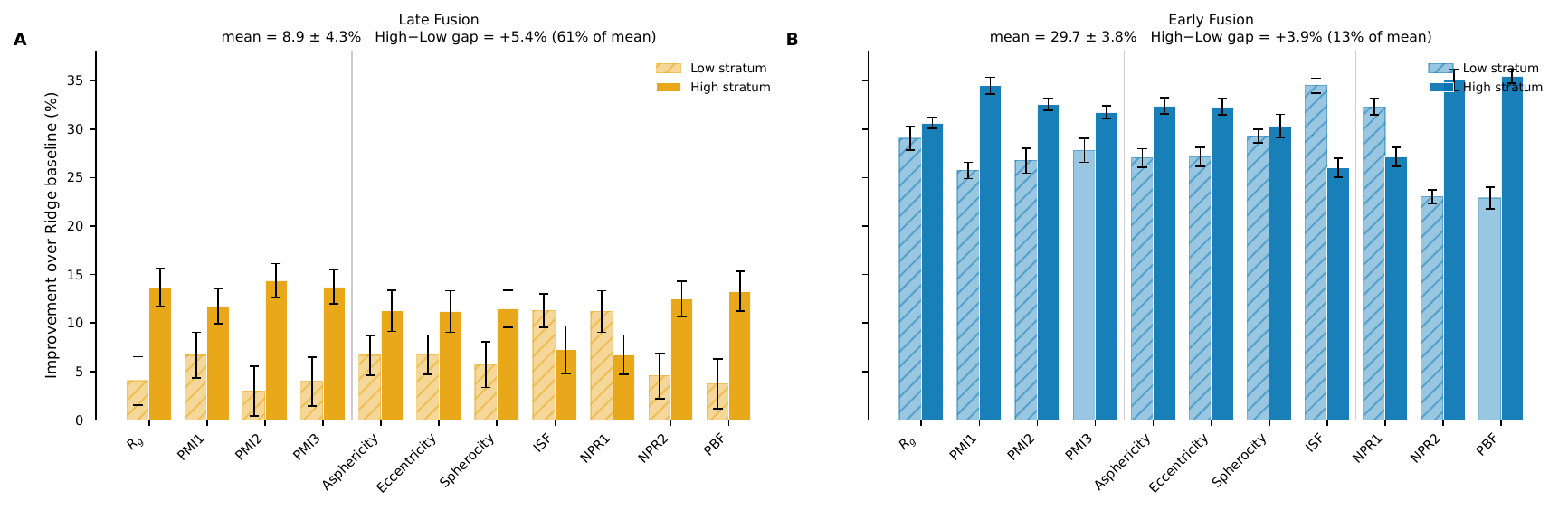}
  \caption{Stratified geometrical performance analysis on the adduct-sensitive test set
    (mean $\pm$ 1\,s.d.\ across five seeds).
    Molecules are split at the median of each RDKit 3D descriptor into a LOW (below
    median, hatched bars) and a HIGH (above median, solid bars) stratum; bars show the
    percentage RMSE improvement over the Ridge 2D baseline within each stratum.
    Descriptors are grouped into size (\textit{$R_g$, PMI1--3}),
    shape (\textit{Asphericity, Eccentricity, Spherocity, ISF, NPR1--2}), and
    planarity (\textit{PBF}), separated by vertical grey lines.
    \textbf{(A)}~Late Fusion (orange): overall improvement of $8.9 \pm 4.3$\%
    (CV\,=\,48\%) with a systematic High\,$>$\,Low bias (mean gap $+5.4$\%,
    equal to 61\% of the mean improvement).
    \textbf{(B)}~\ModelName{} Early Fusion (blue): overall improvement of
    $29.7 \pm 3.8$\% (CV\,=\,13\%), with the High\,$-$\,Low gap of only $+3.9$\%
    representing 13\% of the mean improvement, a four-fold reduction in relative
    geometry-dependence compared to Late Fusion.}
  \label{fig:geom-stratified}
\end{figure}

We performed a stratified analysis on the adduct-sensitive test set to test whether
\ModelName{} has learned geometry-sensitive representations, using eleven RDKit
3D descriptors spanning molecular size ($R_g$, PMI1--3), shape (asphericity, eccentricity,
spherocity index, inertial shape factor, NPR1--2), and planarity (PBF).
For each descriptor, molecules were divided at the sample median into a LOW stratum (below
median) and a HIGH stratum (above median), and the percentage RMSE improvement over the
Ridge 2D baseline was computed within each stratum across five seeds
(Figure~\ref{fig:geom-stratified}).
A model that relies primarily on mean adduct offsets would be expected to improve relatively
uniformly across geometry strata; one that encodes 3D structure would show larger gains
where geometric features are most informative.

Late fusion achieves an overall improvement of only $8.9 \pm 4.3$\% (CV\,=\,48\%) over
the Ridge baseline, with a systematic HIGH\,$>$\,LOW gap of $+5.4$\% on average,
representing 61\% of its mean improvement.
Late-fusion predictions are thus driven primarily by molecular size and shape complexity,
not a uniform 3D understanding.
\ModelName{} (early fusion) achieves an overall improvement of $29.7 \pm 3.8$\%
(CV\,=\,13\%), with a HIGH\,$-$\,LOW gap of only $+3.9$\%, representing 13\% of its mean
improvement, a four-fold reduction in relative geometry-dependence compared to late
fusion.
Across the 11 geometry descriptors and five seeds, early fusion improves over the Ridge
baseline by $29.7 \pm 3.8$\%, compared with $8.9 \pm 4.3$\% for late fusion.
Because many descriptors are correlated and the same test molecules are reused across
strata, we treat this as a diagnostic consistency analysis rather than a formal
independent hypothesis test.
Encoder-level adduct conditioning allows the model to learn 3D structural information
across the full range of molecular geometries.

\subsection{Counterfactual Adduct Swap Analysis}

\begin{figure}[H]
  \centering
  \includegraphics[width=\textwidth]{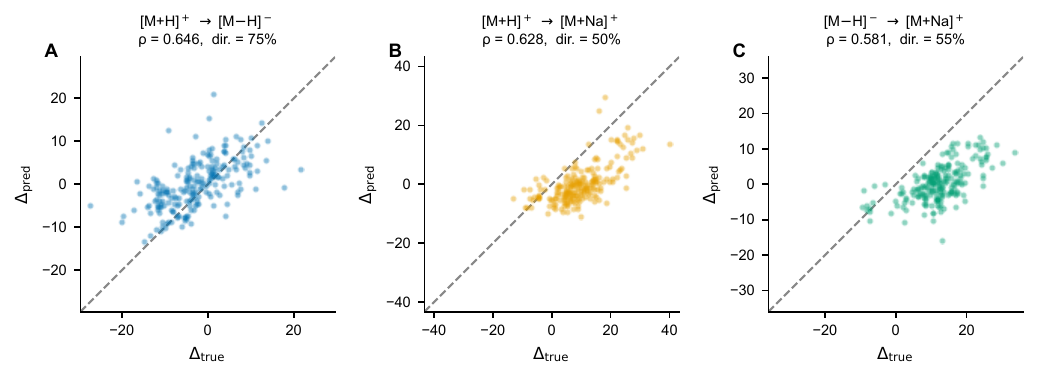}
  \caption{Counterfactual adduct swapping: predicted vs.\ true CCS differences (seed~0).
    The 3-D conformer is held fixed while only the adduct conditioning token changes;
    $\Delta_\mathrm{pred}$ is the resulting change in the head's residual output and
    $\Delta_\mathrm{true}$ is the corresponding measured CCS difference.
    The Ridge-baseline contribution is excluded from $\Delta_\mathrm{pred}$ so that the
    analysis isolates changes in the learned residual output rather than the linear adduct
    offset.
    Panels show the three adduct-pair transitions:
    \mbox{[M+H]$^+$}$\leftrightarrow$\mbox{[M$-$H]$^-$}~(A),
    \mbox{[M+H]$^+$}$\to$\mbox{[M+Na]$^+$}~(B), and
    \mbox{[M$-$H]$^-$}$\to$\mbox{[M+Na]$^+$}~(C);
    points on the grey dashed diagonal satisfy $\Delta_\mathrm{pred}=\Delta_\mathrm{true}$.}
  \label{fig:counterfactual-scatter}
\end{figure}

\begin{figure}[H]
  \centering
  \includegraphics[width=0.7\textwidth]{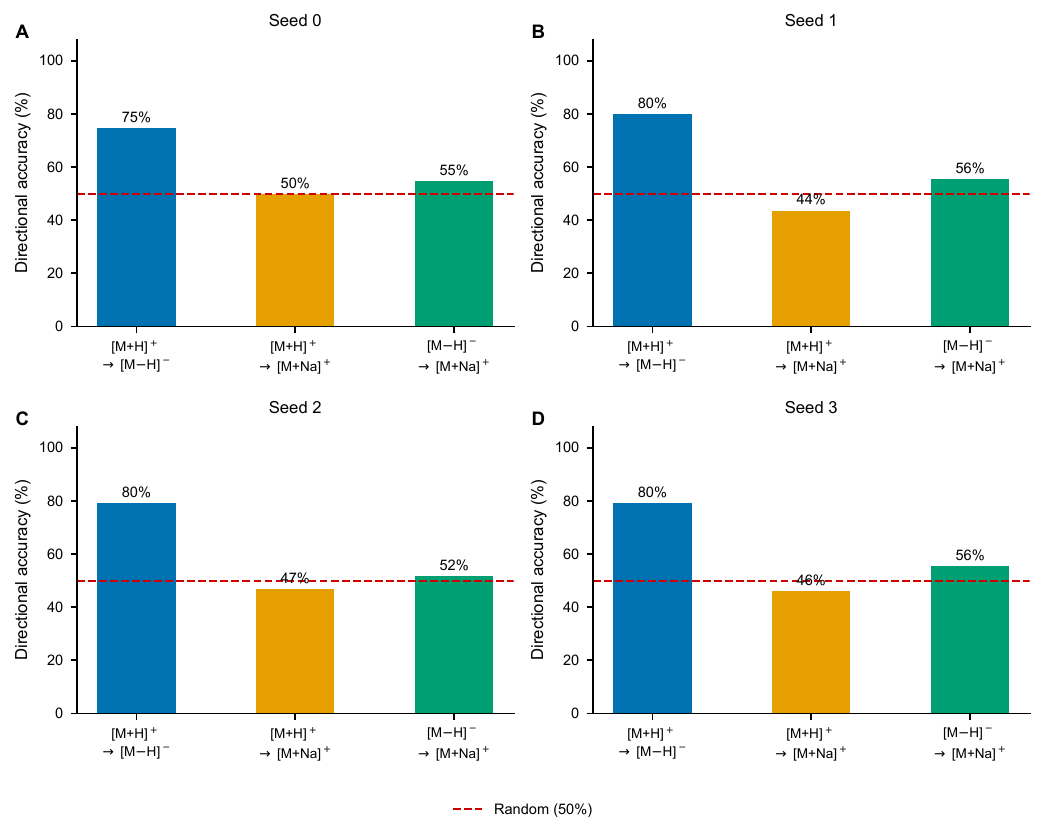}
  \caption{Directional accuracy of counterfactual adduct swaps across four random seeds~(A--D).
    Bar height gives the percentage of molecules for which the model correctly predicted the
    sign of $\Delta$CCS when only the adduct label was swapped while holding the 3-D conformer
    fixed; bar colours match Fig.~\ref{fig:counterfactual-scatter} and the red dashed line
    marks the 50\% chance baseline.}
  \label{fig:counterfactual-accuracy}
\end{figure}

To probe whether the adduct conditioning in \ModelName{} encodes molecule-specific responses rather than a fixed per-adduct offset, we conducted a counterfactual swapping
analysis on the adduct-sensitive test set.
For every molecule appearing under all three adduct forms, the 3-D conformer was held fixed
while only the adduct conditioning token was changed, and the resulting change in the head's
residual output ($\Delta_\mathrm{pred}$) was compared with the corresponding measured CCS
difference ($\Delta_\mathrm{true}$).
The Ridge baseline captures a fixed mean offset per adduct type and is excluded from
$\Delta_\mathrm{pred}$ by construction; the counterfactual therefore tests whether the
head has learned molecule-specific adduct sensitivity beyond that offset.

For the charge-state swap \mbox{[M+H]$^+$}$\leftrightarrow$\mbox{[M$-$H]$^-$}, the model
achieves 75--80\% directional accuracy (correct sign of $\Delta$CCS) across all four seeds
and a Spearman rank correlation of $\rho = 0.65$
(Figures~\ref{fig:counterfactual-scatter}, \ref{fig:counterfactual-accuracy}).
The model correctly ranks which molecules show larger CCS shifts when the charge sign changes.

For sodium-containing swaps (\mbox{[M+H]$^+$}$\to$\mbox{[M+Na]$^+$} and
\mbox{[M$-$H]$^-$}$\to$\mbox{[M+Na]$^+$}), directional accuracy falls to 44--56\%, near
the 50\% chance level, while rank correlations remain positive ($\rho = 0.63$ and $0.58$,
respectively).
This apparent contradiction is a systematic consequence of residual learning: the Ridge
baseline absorbs the mean \mbox{[M+Na]$^+$} offset as a fixed linear term, leaving the
head's residual predictions centered near zero for sodium swaps and rendering the sign of
$\Delta_\mathrm{pred}$ uninformative even when the model correctly encodes relative magnitude.

The counterfactual results show that \ModelName{} has learned adduct-sensitive molecular
representations beyond fixed per-adduct offsets.
The signal is clearest for the charge-state swap, where the Ridge baseline does not absorb
the directional information.

\subsection{Limitations and Future Directions}

The most significant limitation is that conformers are generated and weighted on the
neutral molecule.
Generating conformers after adduct formation using an ionic force field or semiempirical
optimization, and weighting by the corresponding ionic energies, would eliminate the
mismatch between the Boltzmann ensemble and the actual ion population and is likely to
recover the benefit of multi-conformer pooling on the adduct-sensitive split.

A second limitation is dataset size: with $\sim$9{,}200 \Dataset{} measurements across three adducts,
learned attention pooling overfits and the adduct embedding has only three distinct inputs
to learn from.
Both problems would diminish with larger datasets covering more adduct types and instrument
conditions.
Predictions for very out-of-distribution molecule-adduct pairs may degrade, and
well-converged physics-based approaches remain a more reliable option in such cases.

Finally, the current evaluation covers only
[\textit{M}+H]\textsuperscript{+},
[\textit{M}$-$H]\textsuperscript{$-$}, and
[\textit{M}+Na]\textsuperscript{+}.
Extension to multiply-charged ions, metal adducts, and lipid-relevant adducts such as
[\textit{M}+NH\textsubscript{4}]\textsuperscript{+} and
[\textit{M}+HCOO]\textsuperscript{$-$} would require additional curation but would
substantially broaden applicability in untargeted metabolomics workflows.

\section{Conclusion}
\label{sec:conclusion}

We presented \ModelName, a 3D CCS predictor with two chemically motivated design choices:
adduct conditioning within the encoder attention layers via parameter-efficient LoRA adapters,
and residual learning against a physical descriptor baseline.

\ModelName{} achieves the lowest MPD on all three \Dataset{} evaluation splits and
competitive RMSE relative to GraphCCS and SigmaCCS.
On the random split it achieves MPD $1.67$\% versus $1.77$\% for the best competing model;
on the scaffold and adduct-sensitive splits MPD is $2.11$\% and $2.36$\% against $2.28$\%
and $2.51$\%.
The generalization gap is consistently smaller than GraphCCS across all three splits,
with the largest reduction on the random split.
Residual learning and encoder-level adduct conditioning act together; the residual target
removes the dominant size-driven component of CCS, while early fusion lets the pretrained
geometry encoder learn adduct-dependent residual structure. By initializing the model
to predict the mass- and surface-area-driven component of CCS exactly, the encoder focuses
its gradient budget on the adduct- and geometry-dependent residual from the first epoch.

The conformer pooling ablation reveals a physically interpretable failure mode.
Boltzmann weights derived from neutral-molecule MMFF energies degrade performance on the
adduct-sensitive split because the neutral thermal ensemble does not reflect the
conformational population of the charged ion.
A single low-energy conformer is a better proxy, and ion-specific conformer weighting is
a natural direction for future improvement.

The counterfactual adduct-swap analysis shows that early fusion encodes adduct identity
as a geometric signal.
The model achieves 75\% directional accuracy on charge-sign swaps
([\textit{M}+H]\textsuperscript{+}$\leftrightarrow$[\textit{M}$-$H]\textsuperscript{$-$}),
well above the 50\% random baseline.
Sodium-containing swaps fall near chance because the Ridge baseline absorbs the mean
[\textit{M}+Na]\textsuperscript{+} offset as a fixed linear term, leaving no net
directional signal in the head's residual output for those transitions; rank correlations
remain positive, confirming the model encodes relative magnitude even when direction is
indeterminate.

On this held-out experimental set, \ModelName{} outperforms the previously reported
physics-based workflows under the published settings, suggesting that learned empirical
models can be highly competitive for fast CCS estimation when trained on relevant data.

We expect the residual learning formulation to transfer to other molecular properties where
a physics-based prior explains a large fraction of the variance, and the adduct-sensitive
split to serve as a standard challenge set for future 3D CCS predictors.

The two design choices in \ModelName{} rest on a shared principle: encoding existing domain
knowledge directly into the model architecture reduces the effective complexity of what the
network must learn.
Residual learning confines the prediction target to variance that physics-based descriptors
cannot explain; adduct conditioning ensures that ionization state propagates through the
geometry-processing computation rather than serving as a post-hoc scalar correction at the
prediction head.
Neither modification requires additional training data or increased model capacity.
For molecular property prediction tasks where a partial physical model is available,
incorporating it as an inductive bias in the architecture or training objective tends to
improve generalization over unconstrained regression.
How to embed such priors systematically, rather than relying on scale to discover them
implicitly, remains an underexplored design question as pretrained 3D molecular encoders
see broader adoption across property prediction tasks.

\section*{Author Contributions} \label{sec:author_contrib}
\ddag\, P.S. and S.D. contributed equally to this work. P.S. led the development and execution of the machine-learning methodology,  model implementation, training and visualization.  S.D. and K.M. led project conceptualization. P.S., S.D. and J.M. designed the experiments.  P.S., S.D., S.S., and J.M. performed dataset curation, validation and drafted the manuscript.  P.S. and S.S. performed formal analysis.  J.M. provided project administration.  K.M. provided supervision, strategic guidance, and project oversight. All authors reviewed, edited, and approved the final manuscript.
\section*{Data and Software Availability} \label{sec:datasoftwareavail}
\ifpreprint
The source code described in this work is open-source and publicly  available at \href{https://github.com/IBM/GRACE}
{https://github.com/IBM/GRACE}.
The model weights are available at \href{https://github.com/IBM/GRACE/releases/tag/v0.1.0}{https://github.com/IBM/GRACE/releases/tag/v0.1.0}.
The model was trained and validated on data curated from publicly available sources that are referenced above.
\else
The source code described in this work is open-source and publicly  available at \newline \href{https://github.com/IBM/GRACE}
{https://github.com/IBM/GRACE}.
The model weights are available at \newline \href{https://github.com/IBM/GRACE/releases/tag/v0.1.0}{https://github.com/IBM/GRACE/releases/tag/v0.1.0}.
The model was trained and validated on data curated from publicly available sources that are referenced above.
\fi
\section*{Acknowledgments} \label{sec:acknowledgments}
The authors gratefully acknowledge financial support from the National Science Foundation (NSF) through CSSI Frameworks Grant OAC-2209717 and from the National Institutes of Health (Grant Number GM130641). The authors are grateful to the High-Performance Computing Center (iCER HPCC) at Michigan State University and the High-Performance Computing resources at Cleveland Clinic Foundation.
\bibliographystyle{unsrt}
\bibliography{references}

\clearpage
\ifpreprint
\appendix
\fi

\section{Supplementary Material}
\label{sec:supplementary}
\subsection{Dataset Curation: Excluded Molecules}
\label{sec:supp-excluded-mols}

\begin{figure}[H]
  \centering
  \includegraphics[width=\textwidth]{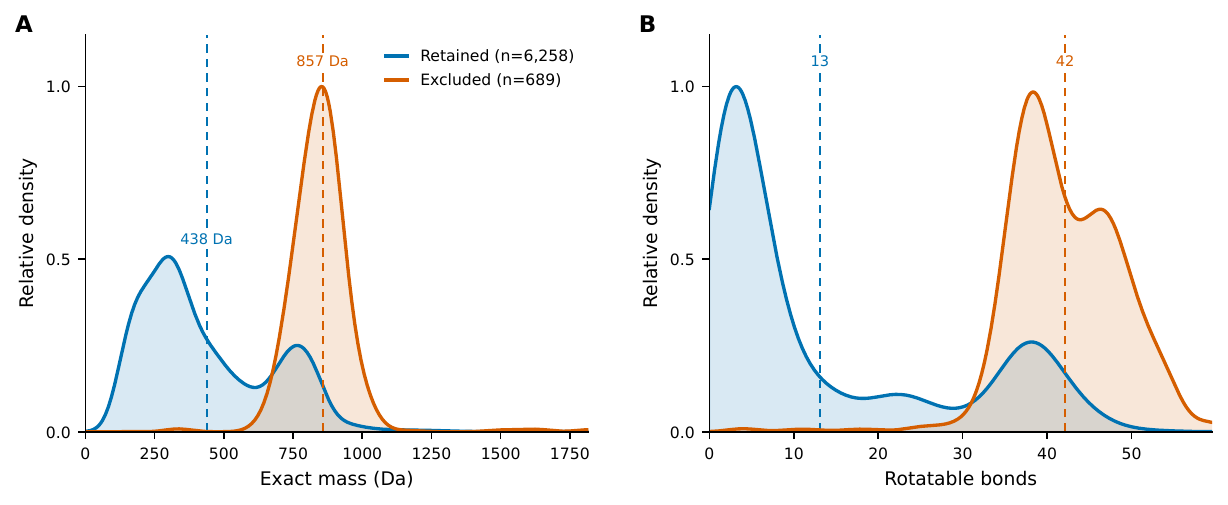}
  \caption{\textbf{Excluded molecules are large, flexible, and predominantly acyclic.}
    Distributions for the 6{,}258 unique molecules retained in the dataset (blue) and the
    689 unique molecules (855 SMILES-adduct pairs) excluded because RDKit ETKDG+MMFF94
    failed to generate valid 3D conformers (orange).
    Density curves are scaled relative to the taller of the two peaks; peak height and
    width reflect group homogeneity.
    \textbf{(A)}~Exact mass: excluded molecules average nearly twice the mass of retained
    ones (857~Da vs.\ 438~Da).
    \textbf{(B)}~Rotatable bond count: the gap is sharper.
    Excluded molecules cluster around 41--42 rotatable bonds (median~41, mean~42.1) versus
    a median of 6 (mean~13.1) for retained molecules; 94\% of excluded molecules meet a
    lipid-like criterion ($\geq$10 rotatable bonds, $\leq$1 ring), compared with 32\% of
    retained molecules.
    ETKDG conformer search cost scales combinatorially with torsional degrees of freedom,
    and MMFF94 struggles with the long, largely acyclic alkyl and ester chains typical of
    glycerophospholipids.
    Exclusion was driven by conformational flexibility rather than molecular mass.}
  \label{fig:supp-excluded-mols}
\end{figure}

\subsection{Details of External Test Sets}
\label{sec:supp-ext-datasets}
Table~\ref{tab:supp-ext-datasets} reports the sources of the external test sets, their original size, and the size of the de-duplicated sets used in this study.
For TS4 (GPCL), results are reported on all 20 molecules and the model is retrained with duplicates removed from the training set.

\begin{table}[H]
  \centering
  \caption{External test sets used in this study. Columns indicate their source, initial size,
    and the size of the de-duplicated set used after removing entries that overlap with \Dataset{}.
    For TS4 (GPCL), results are reported on all 20 molecules and the model is retrained with
    duplicates removed from the training set.}
  \label{tab:supp-ext-datasets}
  \begin{tabular}{llcc}
    \toprule
    Dataset & Source & Initial size & Production size \\
    \midrule
    TS1 & GraphCCS External 1~\cite{xie2024large}   & 1737 & 1163 \\
    TS2 & GraphCCS External 2~\cite{xie2024large}   &   68 &   49 \\
    TS3 & Experimental values from ISiCLE~\cite{colby2019isicle} & 1396 & 298 \\
    TS4 & GPCL~\cite{das2024molecular}              &   20 &   20 \\
    \bottomrule
  \end{tabular}
\end{table}

\subsection{Full \ModelName{} Ablation Across All Pooling and Token Configurations}
\label{sec:supp_full_ablation}

Table~\ref{tab:supp_ablation} reports all 12 \ModelName{} configurations (three conformer
pooling modes $\times$ two token representations $\times$ two fusion strategies) across all
three evaluation splits.

\begin{table}[H]
  \centering
  % Old: \ModelName{} early fusion performance across all conformer pooling modes. Residual target.
  \caption{Effect of conformer pooling strategy on \ModelName{} prediction accuracy on the \Dataset{} test set. Single-conformer, uniform ensemble, Boltzmann energy-weighted, and learned attention-weighted pooling are evaluated under each training split strategy using a residual learning objective. Mean $\pm$ std across 5 random seeds.}
  \label{tab:supp_ablation}
  \begin{adjustbox}{max width=\linewidth, max totalheight=0.85\textheight, keepaspectratio}
\begin{tabular}{llccc}
    \toprule
    Split & Pooling & RMSE & MPD & Pearson $R$ \\
    \midrule
  \multirow{4}{*}{Random} & Single & 4.637 $\pm$ 0.047 & 1.670 $\pm$ 0.023 & 0.9963 $\pm$ 0.0001 \\
   & Uniform & 4.750 $\pm$ 0.057 & 1.700 $\pm$ 0.009 & 0.9962 $\pm$ 0.0001 \\
   & Boltzmann & 4.711 $\pm$ 0.031 & 1.715 $\pm$ 0.009 & 0.9962 $\pm$ 0.0000 \\
   & Learned & 4.705 $\pm$ 0.077 & 1.694 $\pm$ 0.017 & 0.9962 $\pm$ 0.0001 \\
    \midrule
  \multirow{4}{*}{Scaffold} & Single & 6.539 $\pm$ 0.080 & 2.116 $\pm$ 0.022 & 0.9920 $\pm$ 0.0002 \\
   & Uniform & 6.540 $\pm$ 0.095 & 2.113 $\pm$ 0.042 & 0.9920 $\pm$ 0.0002 \\
   & Boltzmann & 6.534 $\pm$ 0.184 & 2.144 $\pm$ 0.083 & 0.9920 $\pm$ 0.0004 \\
   & Learned & 6.552 $\pm$ 0.121 & 2.134 $\pm$ 0.076 & 0.9922 $\pm$ 0.0002 \\
    \midrule
  \multirow{4}{*}{Adduct-sens.} & Single & 6.341 $\pm$ 0.076 & 2.363 $\pm$ 0.032 & 0.9919 $\pm$ 0.0002 \\
   & Uniform & 6.476 $\pm$ 0.127 & 2.425 $\pm$ 0.057 & 0.9914 $\pm$ 0.0002 \\
   & Boltzmann & 6.583 $\pm$ 0.034 & 2.458 $\pm$ 0.015 & 0.9912 $\pm$ 0.0001 \\
   & Learned & 6.446 $\pm$ 0.088 & 2.429 $\pm$ 0.062 & 0.9914 $\pm$ 0.0003 \\
    \bottomrule
  \end{tabular}
\end{adjustbox}
\end{table}

\subsection{Full \Dataset{} Results Across All Configurations and Splits}
\label{sec:supp_internal}

Table~\ref{tab:result_internal_full} extends Table~\ref{tab:result_internal_best} to all
four conformer pooling modes across all three training splits, including train RMSE and the
generalization gap.

\begin{table}[H]
  \centering
  % Old: Internal test set: all \ModelName{} pooling modes and competitors. Bold: lowest MPD per split.
  \caption{Ablation of \ModelName{} conformer pooling strategies against GraphCCS and SigmaCCS on the \Dataset{} test set. All four pooling modes (Single, Uniform, Boltzmann, and Learned) are evaluated under each training split; train and test MPD, RMSE, and all rank-correlation metrics are reported alongside the generalization gap. Competitor pooling is not applicable (—). Mean $\pm$ std across 5 random seeds.}
  \label{tab:result_internal_full}
  \begin{adjustbox}{max width=\linewidth, max totalheight=0.85\textheight, keepaspectratio}
\begin{tabular}{lll r rr rr rrrr}
    \toprule
    & & & & \multicolumn{2}{c}{Train} & \multicolumn{5}{c}{Test} & \\
    \cmidrule(lr){5-6}\cmidrule(lr){7-11}
    Split & Model & Pooling & $n$ & RMSE & MPD & RMSE & MPD & Pearson $R$ & Spearman $\rho$ & Kendall $\tau$ & Gap \\
    \midrule
  \multirow{6}{*}{Random} & GraphCCS & — & \multirow{6}{*}{922} & 3.287 $\pm$ 0.460 & 1.150 $\pm$ 0.171 & 4.858 $\pm$ 0.112 & 1.766 $\pm$ 0.050 & 0.9960 $\pm$ 0.0002 & 0.9938 $\pm$ 0.0003 & 0.9399 $\pm$ 0.0016 & 1.571 $\pm$ 0.494 \\
   & SigmaCCS & — &  & 4.664 $\pm$ 0.390 & 1.728 $\pm$ 0.102 & 5.137 $\pm$ 0.166 & 1.950 $\pm$ 0.051 & 0.9955 $\pm$ 0.0003 & 0.9932 $\pm$ 0.0004 & 0.9341 $\pm$ 0.0019 & 0.473 $\pm$ 0.251 \\
   & \ModelName{} & Single &  & 4.556 $\pm$ 0.098 & 0.996 $\pm$ 0.034 & 4.637 $\pm$ 0.047 & \textbf{1.670} $\pm$ 0.023 & 0.9963 $\pm$ 0.0001 & 0.9946 $\pm$ 0.0002 & 0.9431 $\pm$ 0.0009 & 0.081 $\pm$ 0.085 \\
   & \ModelName{} & Uniform &  & 4.611 $\pm$ 0.242 & 1.073 $\pm$ 0.074 & 4.750 $\pm$ 0.057 & 1.700 $\pm$ 0.009 & 0.9962 $\pm$ 0.0001 & 0.9943 $\pm$ 0.0001 & 0.9421 $\pm$ 0.0006 & 0.139 $\pm$ 0.263 \\
   & \ModelName{} & Boltzmann &  & 4.430 $\pm$ 0.131 & 1.007 $\pm$ 0.050 & 4.711 $\pm$ 0.031 & 1.715 $\pm$ 0.009 & 0.9962 $\pm$ 0.0000 & 0.9945 $\pm$ 0.0001 & 0.9421 $\pm$ 0.0002 & 0.281 $\pm$ 0.155 \\
   & \ModelName{} & Learned &  & 4.597 $\pm$ 0.216 & 1.079 $\pm$ 0.084 & 4.705 $\pm$ 0.077 & 1.694 $\pm$ 0.017 & 0.9962 $\pm$ 0.0001 & 0.9944 $\pm$ 0.0002 & 0.9423 $\pm$ 0.0007 & 0.108 $\pm$ 0.207 \\
    \midrule
  \multirow{6}{*}{Scaffold} & GraphCCS & — & \multirow{6}{*}{920} & 3.844 $\pm$ 0.921 & 1.357 $\pm$ 0.299 & 6.396 $\pm$ 0.282 & 2.278 $\pm$ 0.182 & 0.9925 $\pm$ 0.0004 & 0.9907 $\pm$ 0.0003 & 0.9235 $\pm$ 0.0023 & 2.552 $\pm$ 0.718 \\
   & SigmaCCS & — &  & 6.771 $\pm$ 1.183 & 2.534 $\pm$ 0.450 & 6.867 $\pm$ 0.693 & 2.795 $\pm$ 0.416 & 0.9915 $\pm$ 0.0015 & 0.9886 $\pm$ 0.0017 & 0.9126 $\pm$ 0.0072 & 0.096 $\pm$ 0.696 \\
   & \ModelName{} & Single &  & 5.190 $\pm$ 0.312 & 1.254 $\pm$ 0.144 & 6.539 $\pm$ 0.080 & 2.116 $\pm$ 0.022 & 0.9920 $\pm$ 0.0002 & 0.9912 $\pm$ 0.0002 & 0.9258 $\pm$ 0.0011 & 1.349 $\pm$ 0.266 \\
   & \ModelName{} & Uniform &  & 5.270 $\pm$ 0.567 & 1.317 $\pm$ 0.244 & 6.540 $\pm$ 0.095 & \textbf{2.113} $\pm$ 0.042 & 0.9920 $\pm$ 0.0002 & 0.9915 $\pm$ 0.0003 & 0.9271 $\pm$ 0.0014 & 1.270 $\pm$ 0.528 \\
   & \ModelName{} & Boltzmann &  & 5.329 $\pm$ 0.723 & 1.348 $\pm$ 0.329 & 6.534 $\pm$ 0.184 & 2.144 $\pm$ 0.083 & 0.9920 $\pm$ 0.0004 & 0.9914 $\pm$ 0.0006 & 0.9263 $\pm$ 0.0028 & 1.205 $\pm$ 0.577 \\
   & \ModelName{} & Learned &  & 5.269 $\pm$ 0.435 & 1.329 $\pm$ 0.179 & 6.552 $\pm$ 0.121 & 2.134 $\pm$ 0.076 & 0.9922 $\pm$ 0.0002 & 0.9916 $\pm$ 0.0002 & 0.9279 $\pm$ 0.0010 & 1.283 $\pm$ 0.381 \\
    \midrule
  \multirow{6}{*}{Adduct-sens.} & GraphCCS & — & \multirow{6}{*}{1382} & 4.117 $\pm$ 0.400 & 1.373 $\pm$ 0.086 & 6.671 $\pm$ 0.076 & 2.512 $\pm$ 0.038 & 0.9911 $\pm$ 0.0004 & 0.9875 $\pm$ 0.0005 & 0.9105 $\pm$ 0.0021 & 2.554 $\pm$ 0.385 \\
   & SigmaCCS & — &  & 5.263 $\pm$ 0.916 & 2.099 $\pm$ 0.379 & 6.918 $\pm$ 0.280 & 2.756 $\pm$ 0.193 & 0.9899 $\pm$ 0.0006 & 0.9858 $\pm$ 0.0012 & 0.9032 $\pm$ 0.0049 & 1.655 $\pm$ 0.783 \\
   & \ModelName{} & Single &  & 4.780 $\pm$ 0.161 & 1.023 $\pm$ 0.054 & 6.341 $\pm$ 0.076 & \textbf{2.363} $\pm$ 0.032 & 0.9919 $\pm$ 0.0002 & 0.9887 $\pm$ 0.0003 & 0.9149 $\pm$ 0.0012 & 1.562 $\pm$ 0.113 \\
   & \ModelName{} & Uniform &  & 4.989 $\pm$ 0.234 & 1.169 $\pm$ 0.074 & 6.476 $\pm$ 0.127 & 2.425 $\pm$ 0.057 & 0.9914 $\pm$ 0.0002 & 0.9878 $\pm$ 0.0003 & 0.9117 $\pm$ 0.0015 & 1.487 $\pm$ 0.144 \\
   & \ModelName{} & Boltzmann &  & 4.790 $\pm$ 0.210 & 1.056 $\pm$ 0.067 & 6.583 $\pm$ 0.034 & 2.458 $\pm$ 0.015 & 0.9912 $\pm$ 0.0001 & 0.9878 $\pm$ 0.0003 & 0.9114 $\pm$ 0.0008 & 1.793 $\pm$ 0.187 \\
   & \ModelName{} & Learned &  & 5.286 $\pm$ 0.395 & 1.321 $\pm$ 0.198 & 6.446 $\pm$ 0.088 & 2.429 $\pm$ 0.062 & 0.9914 $\pm$ 0.0003 & 0.9876 $\pm$ 0.0004 & 0.9113 $\pm$ 0.0020 & 1.160 $\pm$ 0.337 \\
    \bottomrule
  \end{tabular}
\end{adjustbox}
\end{table}

\noindent Per-adduct breakdown for the best-performing configuration per split is given in
Table~\ref{tab:result_internal_best_breakdown}, and for all pooling modes in
Table~\ref{tab:result_internal_breakdown_full}.

\begin{table}[H]
  \centering
  % Old: Internal test set per-adduct breakdown: best model configuration.
  \caption{Per-adduct CCS prediction accuracy on the \Dataset{} test set for the best-performing model configuration per split (\ModelName{} pooling mode as in Table~\ref{tab:result_internal_best}). \ModelName{} achieves the lowest MPD for \mbox{[M+H]$^+$} and \mbox{[M+Na]$^+$} across all splits; GraphCCS is competitive on \mbox{[M$-$H]$^-$}. GraphCCS per-adduct training metrics are unavailable (—). Mean $\pm$ std across 5 random seeds.}
  \label{tab:result_internal_best_breakdown}
  \begin{adjustbox}{max width=\linewidth, max totalheight=0.85\textheight, keepaspectratio}
\begin{tabular}{lll r rr rr r}
    \toprule
    & & & & \multicolumn{2}{c}{Train} & \multicolumn{2}{c}{Test} & \\
    \cmidrule(lr){5-6}\cmidrule(lr){7-8}
    Split & Adduct & Model & $n$ & RMSE & MPD & RMSE & MPD & Gap \\
    \midrule
  \multirow{9}{*}{Random} & \multirow{3}{*}{[M+H]+} & GraphCCS & 432 & — & — & 4.593 $\pm$ 0.118 & 1.719 $\pm$ 0.059 & — \\
   &  & SigmaCCS & 432 & 4.462 $\pm$ 0.481 & 1.617 $\pm$ 0.093 & 4.815 $\pm$ 0.176 & 1.891 $\pm$ 0.068 & 0.353 $\pm$ 0.392 \\
   &  & \ModelName{} (Single) & 432 & 5.744 $\pm$ 0.098 & 0.945 $\pm$ 0.033 & 4.382 $\pm$ 0.041 & \textbf{1.660} $\pm$ 0.013 & -1.361 $\pm$ 0.098 \\
    \cmidrule(lr){2-9}
   & \multirow{3}{*}{[M-H]-} & GraphCCS & 261 & — & — & 4.868 $\pm$ 0.102 & 1.637 $\pm$ 0.045 & — \\
   &  & SigmaCCS & 261 & 4.773 $\pm$ 0.260 & 1.812 $\pm$ 0.093 & 5.222 $\pm$ 0.086 & 1.881 $\pm$ 0.037 & 0.449 $\pm$ 0.198 \\
   &  & \ModelName{} (Single) & 261 & 3.205 $\pm$ 0.134 & 1.044 $\pm$ 0.045 & 4.717 $\pm$ 0.117 & \textbf{1.591} $\pm$ 0.051 & 1.512 $\pm$ 0.104 \\
    \cmidrule(lr){2-9}
   & \multirow{3}{*}{[M+Na]+} & GraphCCS & 229 & — & — & 5.057 $\pm$ 0.095 & 1.855 $\pm$ 0.033 & — \\
   &  & SigmaCCS & 229 & 4.898 $\pm$ 0.418 & 1.840 $\pm$ 0.139 & 5.603 $\pm$ 0.274 & 2.138 $\pm$ 0.089 & 0.705 $\pm$ 0.153 \\
   &  & \ModelName{} (Single) & 229 & 3.138 $\pm$ 0.125 & 1.039 $\pm$ 0.035 & 4.997 $\pm$ 0.098 & \textbf{1.778} $\pm$ 0.035 & 1.860 $\pm$ 0.129 \\
    \midrule
  \multirow{9}{*}{Scaffold} & \multirow{3}{*}{[M+H]+} & GraphCCS & 474 & — & — & 6.535 $\pm$ 0.344 & 2.198 $\pm$ 0.113 & — \\
   &  & SigmaCCS & 473 & 7.414 $\pm$ 1.655 & 2.637 $\pm$ 0.597 & 6.605 $\pm$ 0.834 & 2.732 $\pm$ 0.531 & -0.809 $\pm$ 1.174 \\
   &  & \ModelName{} (Uniform) & 474 & 6.573 $\pm$ 0.560 & 1.320 $\pm$ 0.240 & 6.069 $\pm$ 0.038 & \textbf{2.040} $\pm$ 0.013 & -0.504 $\pm$ 0.556 \\
    \cmidrule(lr){2-9}
   & \multirow{3}{*}{[M-H]-} & GraphCCS & 259 & — & — & 5.418 $\pm$ 0.368 & 2.083 $\pm$ 0.146 & — \\
   &  & SigmaCCS & 258 & 5.906 $\pm$ 0.606 & 2.384 $\pm$ 0.311 & 6.891 $\pm$ 0.573 & 2.828 $\pm$ 0.377 & 0.985 $\pm$ 0.336 \\
   &  & \ModelName{} (Uniform) & 259 & 3.922 $\pm$ 0.534 & 1.338 $\pm$ 0.223 & 6.356 $\pm$ 0.113 & \textbf{2.002} $\pm$ 0.068 & 2.433 $\pm$ 0.489 \\
    \cmidrule(lr){2-9}
   & \multirow{3}{*}{[M+Na]+} & GraphCCS & 187 & — & — & 7.238 $\pm$ 0.260 & 2.498 $\pm$ 0.167 & — \\
   &  & SigmaCCS & 186 & 6.461 $\pm$ 0.922 & 2.526 $\pm$ 0.379 & 7.447 $\pm$ 0.630 & 2.912 $\pm$ 0.237 & 0.986 $\pm$ 0.314 \\
   &  & \ModelName{} (Uniform) & 187 & 3.767 $\pm$ 0.750 & 1.286 $\pm$ 0.279 & 7.817 $\pm$ 0.232 & \textbf{2.452} $\pm$ 0.124 & 4.050 $\pm$ 0.657 \\
    \midrule
  \multirow{9}{*}{Adduct-sens.} & \multirow{3}{*}{[M+H]+} & GraphCCS & 469 & — & — & 5.857 $\pm$ 0.093 & 2.166 $\pm$ 0.051 & — \\
   &  & SigmaCCS & 469 & 5.291 $\pm$ 1.247 & 2.094 $\pm$ 0.494 & 6.317 $\pm$ 0.643 & 2.533 $\pm$ 0.404 & 1.026 $\pm$ 0.816 \\
   &  & \ModelName{} (Single) & 469 & 5.917 $\pm$ 0.192 & 1.008 $\pm$ 0.058 & 5.606 $\pm$ 0.053 & \textbf{2.022} $\pm$ 0.024 & -0.311 $\pm$ 0.222 \\
    \cmidrule(lr){2-9}
   & \multirow{3}{*}{[M-H]-} & GraphCCS & 409 & — & — & 6.424 $\pm$ 0.215 & 2.436 $\pm$ 0.098 & — \\
   &  & SigmaCCS & 409 & 5.028 $\pm$ 0.460 & 2.060 $\pm$ 0.239 & 6.980 $\pm$ 0.100 & 2.836 $\pm$ 0.130 & 1.951 $\pm$ 0.489 \\
   &  & \ModelName{} (Single) & 409 & 3.140 $\pm$ 0.130 & 1.065 $\pm$ 0.047 & 6.269 $\pm$ 0.111 & \textbf{2.203} $\pm$ 0.057 & 3.129 $\pm$ 0.106 \\
    \cmidrule(lr){2-9}
   & \multirow{3}{*}{[M+Na]+} & GraphCCS & 504 & — & — & 7.170 $\pm$ 0.202 & 2.826 $\pm$ 0.070 & — \\
   &  & SigmaCCS & 504 & 5.464 $\pm$ 0.707 & 2.173 $\pm$ 0.303 & 7.368 $\pm$ 0.295 & 2.898 $\pm$ 0.139 & 1.904 $\pm$ 0.798 \\
   &  & \ModelName{} (Single) & 504 & 2.984 $\pm$ 0.162 & 1.004 $\pm$ 0.055 & 7.008 $\pm$ 0.119 & \textbf{2.811} $\pm$ 0.042 & 4.024 $\pm$ 0.087 \\
    \bottomrule
  \end{tabular}
\end{adjustbox}
\end{table}

\begin{table}[H]
  \centering
  % Old: Internal test set per-adduct breakdown: all \ModelName{} pooling modes and competitors.
  \caption{Per-adduct CCS prediction accuracy on the \Dataset{} test set for all \ModelName{} conformer pooling modes and competitors. GraphCCS per-adduct training metrics are unavailable (—); test Spearman~$\rho$ and Kendall~$\tau$ are available from the test predictions. Mean $\pm$ std across 5 random seeds.}
  \label{tab:result_internal_breakdown_full}
  \begin{adjustbox}{max width=\linewidth, max totalheight=0.85\textheight, keepaspectratio}
\begin{tabular}{llll r rr rr rrrr}
    \toprule
    & & & & & \multicolumn{2}{c}{Train} & \multicolumn{5}{c}{Test} & \\
    \cmidrule(lr){6-7}\cmidrule(lr){8-12}
    Split & Adduct & Model & Pooling & $n$ & RMSE & MPD & RMSE & MPD & Pearson $R$ & Spearman $\rho$ & Kendall $\tau$ & Gap \\
    \midrule
  \multirow{18}{*}{Random} & \multirow{6}{*}{[M+H]+} & GraphCCS & — & 432 & — & — & 4.593 $\pm$ 0.118 & 1.719 $\pm$ 0.059 & 0.9965 $\pm$ 0.0001 & 0.9934 $\pm$ 0.0004 & 0.9381 $\pm$ 0.0022 & — \\
   &  & SigmaCCS & — & 432 & 4.462 $\pm$ 0.481 & 1.617 $\pm$ 0.093 & 4.815 $\pm$ 0.176 & 1.891 $\pm$ 0.068 & 0.9962 $\pm$ 0.0003 & 0.9929 $\pm$ 0.0006 & 0.9334 $\pm$ 0.0022 & 0.353 $\pm$ 0.392 \\
   &  & \ModelName{} & Single & 432 & 5.744 $\pm$ 0.098 & 0.945 $\pm$ 0.033 & 4.382 $\pm$ 0.041 & \textbf{1.660} $\pm$ 0.013 & 0.9968 $\pm$ 0.0001 & 0.9947 $\pm$ 0.0001 & 0.9433 $\pm$ 0.0009 & -1.361 $\pm$ 0.098 \\
   &  & \ModelName{} & Uniform & 432 & 5.810 $\pm$ 0.260 & 1.070 $\pm$ 0.071 & 4.435 $\pm$ 0.092 & 1.681 $\pm$ 0.017 & 0.9968 $\pm$ 0.0001 & 0.9945 $\pm$ 0.0002 & 0.9433 $\pm$ 0.0010 & -1.375 $\pm$ 0.322 \\
   &  & \ModelName{} & Boltzmann & 432 & 5.501 $\pm$ 0.155 & 0.950 $\pm$ 0.059 & 4.334 $\pm$ 0.067 & 1.661 $\pm$ 0.019 & 0.9969 $\pm$ 0.0001 & 0.9949 $\pm$ 0.0002 & 0.9440 $\pm$ 0.0006 & -1.167 $\pm$ 0.204 \\
   &  & \ModelName{} & Learned & 432 & 5.792 $\pm$ 0.208 & 1.074 $\pm$ 0.077 & 4.403 $\pm$ 0.040 & 1.680 $\pm$ 0.003 & 0.9968 $\pm$ 0.0001 & 0.9948 $\pm$ 0.0002 & 0.9440 $\pm$ 0.0010 & -1.389 $\pm$ 0.222 \\
    \cmidrule(lr){2-13}
   & \multirow{6}{*}{[M-H]-} & GraphCCS & — & 261 & — & — & 4.868 $\pm$ 0.102 & 1.637 $\pm$ 0.045 & 0.9963 $\pm$ 0.0001 & 0.9929 $\pm$ 0.0003 & 0.9395 $\pm$ 0.0011 & — \\
   &  & SigmaCCS & — & 261 & 4.773 $\pm$ 0.260 & 1.812 $\pm$ 0.093 & 5.222 $\pm$ 0.086 & 1.881 $\pm$ 0.037 & 0.9958 $\pm$ 0.0001 & 0.9923 $\pm$ 0.0003 & 0.9341 $\pm$ 0.0014 & 0.449 $\pm$ 0.198 \\
   &  & \ModelName{} & Single & 261 & 3.205 $\pm$ 0.134 & 1.044 $\pm$ 0.045 & 4.717 $\pm$ 0.117 & 1.591 $\pm$ 0.051 & 0.9965 $\pm$ 0.0001 & 0.9929 $\pm$ 0.0004 & 0.9399 $\pm$ 0.0020 & 1.512 $\pm$ 0.104 \\
   &  & \ModelName{} & Uniform & 261 & 3.356 $\pm$ 0.220 & 1.121 $\pm$ 0.066 & 4.793 $\pm$ 0.111 & 1.608 $\pm$ 0.057 & 0.9965 $\pm$ 0.0001 & 0.9930 $\pm$ 0.0004 & 0.9402 $\pm$ 0.0021 & 1.436 $\pm$ 0.272 \\
   &  & \ModelName{} & Boltzmann & 261 & 3.320 $\pm$ 0.120 & 1.090 $\pm$ 0.041 & 4.960 $\pm$ 0.084 & 1.720 $\pm$ 0.030 & 0.9962 $\pm$ 0.0001 & 0.9922 $\pm$ 0.0003 & 0.9356 $\pm$ 0.0013 & 1.639 $\pm$ 0.174 \\
   &  & \ModelName{} & Learned & 261 & 3.322 $\pm$ 0.204 & 1.121 $\pm$ 0.069 & 4.743 $\pm$ 0.094 & \textbf{1.588} $\pm$ 0.029 & 0.9965 $\pm$ 0.0001 & 0.9931 $\pm$ 0.0003 & 0.9408 $\pm$ 0.0008 & 1.422 $\pm$ 0.188 \\
    \cmidrule(lr){2-13}
   & \multirow{6}{*}{[M+Na]+} & GraphCCS & — & 229 & — & — & 5.057 $\pm$ 0.095 & 1.855 $\pm$ 0.033 & 0.9949 $\pm$ 0.0002 & 0.9924 $\pm$ 0.0007 & 0.9331 $\pm$ 0.0026 & — \\
   &  & SigmaCCS & — & 229 & 4.898 $\pm$ 0.418 & 1.840 $\pm$ 0.139 & 5.603 $\pm$ 0.274 & 2.138 $\pm$ 0.089 & 0.9938 $\pm$ 0.0005 & 0.9906 $\pm$ 0.0007 & 0.9222 $\pm$ 0.0029 & 0.705 $\pm$ 0.153 \\
   &  & \ModelName{} & Single & 229 & 3.138 $\pm$ 0.125 & 1.039 $\pm$ 0.035 & 4.997 $\pm$ 0.098 & \textbf{1.778} $\pm$ 0.035 & 0.9951 $\pm$ 0.0002 & 0.9928 $\pm$ 0.0003 & 0.9339 $\pm$ 0.0017 & 1.860 $\pm$ 0.129 \\
   &  & \ModelName{} & Uniform & 229 & 3.039 $\pm$ 0.287 & 1.021 $\pm$ 0.096 & 5.247 $\pm$ 0.132 & 1.843 $\pm$ 0.044 & 0.9946 $\pm$ 0.0002 & 0.9918 $\pm$ 0.0005 & 0.9293 $\pm$ 0.0013 & 2.208 $\pm$ 0.206 \\
   &  & \ModelName{} & Boltzmann & 229 & 3.081 $\pm$ 0.155 & 1.016 $\pm$ 0.049 & 5.087 $\pm$ 0.133 & 1.810 $\pm$ 0.033 & 0.9949 $\pm$ 0.0003 & 0.9926 $\pm$ 0.0003 & 0.9331 $\pm$ 0.0014 & 2.006 $\pm$ 0.224 \\
   &  & \ModelName{} & Learned & 229 & 3.059 $\pm$ 0.331 & 1.039 $\pm$ 0.123 & 5.185 $\pm$ 0.173 & 1.840 $\pm$ 0.043 & 0.9947 $\pm$ 0.0003 & 0.9920 $\pm$ 0.0006 & 0.9309 $\pm$ 0.0026 & 2.126 $\pm$ 0.284 \\
    \midrule
  \multirow{18}{*}{Scaffold} & \multirow{6}{*}{[M+H]+} & GraphCCS & — & 474 & — & — & 6.535 $\pm$ 0.344 & 2.198 $\pm$ 0.113 & 0.9921 $\pm$ 0.0008 & 0.9896 $\pm$ 0.0007 & 0.9226 $\pm$ 0.0030 & — \\
   &  & SigmaCCS & — & 473 & 7.414 $\pm$ 1.655 & 2.637 $\pm$ 0.597 & 6.605 $\pm$ 0.834 & 2.732 $\pm$ 0.531 & 0.9931 $\pm$ 0.0010 & 0.9889 $\pm$ 0.0015 & 0.9181 $\pm$ 0.0063 & -0.809 $\pm$ 1.174 \\
   &  & \ModelName{} & Single & 474 & 6.536 $\pm$ 0.272 & 1.255 $\pm$ 0.131 & 5.959 $\pm$ 0.100 & \textbf{2.004} $\pm$ 0.030 & 0.9936 $\pm$ 0.0002 & 0.9908 $\pm$ 0.0003 & 0.9262 $\pm$ 0.0013 & -0.576 $\pm$ 0.266 \\
   &  & \ModelName{} & Uniform & 474 & 6.573 $\pm$ 0.560 & 1.320 $\pm$ 0.240 & 6.069 $\pm$ 0.038 & 2.040 $\pm$ 0.013 & 0.9934 $\pm$ 0.0001 & 0.9907 $\pm$ 0.0005 & 0.9274 $\pm$ 0.0016 & -0.504 $\pm$ 0.556 \\
   &  & \ModelName{} & Boltzmann & 474 & 6.595 $\pm$ 0.717 & 1.332 $\pm$ 0.343 & 5.925 $\pm$ 0.213 & 2.035 $\pm$ 0.085 & 0.9937 $\pm$ 0.0003 & 0.9909 $\pm$ 0.0004 & 0.9268 $\pm$ 0.0016 & -0.670 $\pm$ 0.523 \\
   &  & \ModelName{} & Learned & 474 & 6.657 $\pm$ 0.465 & 1.387 $\pm$ 0.187 & 6.193 $\pm$ 0.195 & 2.116 $\pm$ 0.111 & 0.9935 $\pm$ 0.0002 & 0.9909 $\pm$ 0.0006 & 0.9273 $\pm$ 0.0023 & -0.464 $\pm$ 0.447 \\
    \cmidrule(lr){2-13}
   & \multirow{6}{*}{[M-H]-} & GraphCCS & — & 259 & — & — & 5.418 $\pm$ 0.368 & 2.083 $\pm$ 0.146 & 0.9949 $\pm$ 0.0007 & 0.9936 $\pm$ 0.0008 & 0.9345 $\pm$ 0.0038 & — \\
   &  & SigmaCCS & — & 258 & 5.906 $\pm$ 0.606 & 2.384 $\pm$ 0.311 & 6.891 $\pm$ 0.573 & 2.828 $\pm$ 0.377 & 0.9916 $\pm$ 0.0013 & 0.9900 $\pm$ 0.0019 & 0.9166 $\pm$ 0.0083 & 0.985 $\pm$ 0.336 \\
   &  & \ModelName{} & Single & 259 & 3.758 $\pm$ 0.371 & 1.259 $\pm$ 0.147 & 6.358 $\pm$ 0.102 & 2.011 $\pm$ 0.011 & 0.9929 $\pm$ 0.0002 & 0.9937 $\pm$ 0.0002 & 0.9367 $\pm$ 0.0011 & 2.600 $\pm$ 0.367 \\
   &  & \ModelName{} & Uniform & 259 & 3.922 $\pm$ 0.534 & 1.338 $\pm$ 0.223 & 6.356 $\pm$ 0.113 & 2.002 $\pm$ 0.068 & 0.9929 $\pm$ 0.0003 & 0.9938 $\pm$ 0.0004 & 0.9365 $\pm$ 0.0020 & 2.433 $\pm$ 0.489 \\
   &  & \ModelName{} & Boltzmann & 259 & 4.009 $\pm$ 0.741 & 1.378 $\pm$ 0.312 & 6.641 $\pm$ 0.198 & 2.171 $\pm$ 0.110 & 0.9922 $\pm$ 0.0006 & 0.9930 $\pm$ 0.0008 & 0.9331 $\pm$ 0.0037 & 2.632 $\pm$ 0.635 \\
   &  & \ModelName{} & Learned & 259 & 3.805 $\pm$ 0.409 & 1.309 $\pm$ 0.170 & 6.262 $\pm$ 0.222 & \textbf{1.973} $\pm$ 0.098 & 0.9931 $\pm$ 0.0005 & 0.9942 $\pm$ 0.0003 & 0.9388 $\pm$ 0.0017 & 2.457 $\pm$ 0.296 \\
    \cmidrule(lr){2-13}
   & \multirow{6}{*}{[M+Na]+} & GraphCCS & — & 187 & — & — & 7.238 $\pm$ 0.260 & 2.498 $\pm$ 0.167 & 0.9896 $\pm$ 0.0007 & 0.9828 $\pm$ 0.0016 & 0.9007 $\pm$ 0.0055 & — \\
   &  & SigmaCCS & — & 186 & 6.461 $\pm$ 0.922 & 2.526 $\pm$ 0.379 & 7.447 $\pm$ 0.630 & 2.912 $\pm$ 0.237 & 0.9885 $\pm$ 0.0019 & 0.9761 $\pm$ 0.0041 & 0.8818 $\pm$ 0.0106 & 0.986 $\pm$ 0.314 \\
   &  & \ModelName{} & Single & 187 & 3.647 $\pm$ 0.452 & 1.246 $\pm$ 0.167 & 8.019 $\pm$ 0.163 & 2.545 $\pm$ 0.043 & 0.9873 $\pm$ 0.0004 & 0.9815 $\pm$ 0.0015 & 0.8974 $\pm$ 0.0035 & 4.372 $\pm$ 0.295 \\
   &  & \ModelName{} & Uniform & 187 & 3.767 $\pm$ 0.750 & 1.286 $\pm$ 0.279 & 7.817 $\pm$ 0.232 & 2.452 $\pm$ 0.124 & 0.9880 $\pm$ 0.0005 & 0.9840 $\pm$ 0.0013 & 0.9048 $\pm$ 0.0045 & 4.050 $\pm$ 0.657 \\
   &  & \ModelName{} & Boltzmann & 187 & 3.912 $\pm$ 0.870 & 1.341 $\pm$ 0.327 & 7.742 $\pm$ 0.225 & \textbf{2.382} $\pm$ 0.138 & 0.9881 $\pm$ 0.0007 & 0.9841 $\pm$ 0.0023 & 0.9059 $\pm$ 0.0077 & 3.829 $\pm$ 0.753 \\
   &  & \ModelName{} & Learned & 187 & 3.650 $\pm$ 0.479 & 1.249 $\pm$ 0.184 & 7.716 $\pm$ 0.204 & 2.401 $\pm$ 0.053 & 0.9882 $\pm$ 0.0006 & 0.9844 $\pm$ 0.0008 & 0.9061 $\pm$ 0.0022 & 4.066 $\pm$ 0.504 \\
    \midrule
  \multirow{18}{*}{Adduct-sens.} & \multirow{6}{*}{[M+H]+} & GraphCCS & — & 469 & — & — & 5.857 $\pm$ 0.093 & 2.166 $\pm$ 0.051 & 0.9929 $\pm$ 0.0002 & 0.9869 $\pm$ 0.0004 & 0.9133 $\pm$ 0.0014 & — \\
   &  & SigmaCCS & — & 469 & 5.291 $\pm$ 1.247 & 2.094 $\pm$ 0.494 & 6.317 $\pm$ 0.643 & 2.533 $\pm$ 0.404 & 0.9920 $\pm$ 0.0013 & 0.9856 $\pm$ 0.0025 & 0.9075 $\pm$ 0.0098 & 1.026 $\pm$ 0.816 \\
   &  & \ModelName{} & Single & 469 & 5.917 $\pm$ 0.192 & 1.008 $\pm$ 0.058 & 5.606 $\pm$ 0.053 & \textbf{2.022} $\pm$ 0.024 & 0.9935 $\pm$ 0.0001 & 0.9888 $\pm$ 0.0003 & 0.9214 $\pm$ 0.0005 & -0.311 $\pm$ 0.222 \\
   &  & \ModelName{} & Uniform & 469 & 6.084 $\pm$ 0.247 & 1.180 $\pm$ 0.068 & 5.691 $\pm$ 0.041 & 2.030 $\pm$ 0.034 & 0.9933 $\pm$ 0.0001 & 0.9881 $\pm$ 0.0003 & 0.9195 $\pm$ 0.0012 & -0.393 $\pm$ 0.248 \\
   &  & \ModelName{} & Boltzmann & 469 & 5.890 $\pm$ 0.226 & 1.039 $\pm$ 0.058 & 5.726 $\pm$ 0.071 & 2.081 $\pm$ 0.039 & 0.9932 $\pm$ 0.0002 & 0.9881 $\pm$ 0.0002 & 0.9187 $\pm$ 0.0012 & -0.164 $\pm$ 0.255 \\
   &  & \ModelName{} & Learned & 469 & 6.364 $\pm$ 0.379 & 1.319 $\pm$ 0.194 & 5.704 $\pm$ 0.097 & 2.071 $\pm$ 0.044 & 0.9933 $\pm$ 0.0003 & 0.9881 $\pm$ 0.0007 & 0.9202 $\pm$ 0.0023 & -0.660 $\pm$ 0.462 \\
    \cmidrule(lr){2-13}
   & \multirow{6}{*}{[M-H]-} & GraphCCS & — & 409 & — & — & 6.424 $\pm$ 0.215 & 2.436 $\pm$ 0.098 & 0.9929 $\pm$ 0.0005 & 0.9898 $\pm$ 0.0006 & 0.9258 $\pm$ 0.0025 & — \\
   &  & SigmaCCS & — & 409 & 5.028 $\pm$ 0.460 & 2.060 $\pm$ 0.239 & 6.980 $\pm$ 0.100 & 2.836 $\pm$ 0.130 & 0.9918 $\pm$ 0.0003 & 0.9874 $\pm$ 0.0004 & 0.9169 $\pm$ 0.0024 & 1.951 $\pm$ 0.489 \\
   &  & \ModelName{} & Single & 409 & 3.140 $\pm$ 0.130 & 1.065 $\pm$ 0.047 & 6.269 $\pm$ 0.111 & \textbf{2.203} $\pm$ 0.057 & 0.9930 $\pm$ 0.0003 & 0.9888 $\pm$ 0.0003 & 0.9243 $\pm$ 0.0011 & 3.129 $\pm$ 0.106 \\
   &  & \ModelName{} & Uniform & 409 & 3.493 $\pm$ 0.247 & 1.204 $\pm$ 0.087 & 6.349 $\pm$ 0.043 & 2.308 $\pm$ 0.035 & 0.9929 $\pm$ 0.0001 & 0.9891 $\pm$ 0.0001 & 0.9249 $\pm$ 0.0006 & 2.855 $\pm$ 0.260 \\
   &  & \ModelName{} & Boltzmann & 409 & 3.286 $\pm$ 0.231 & 1.115 $\pm$ 0.089 & 6.467 $\pm$ 0.114 & 2.288 $\pm$ 0.062 & 0.9925 $\pm$ 0.0002 & 0.9889 $\pm$ 0.0003 & 0.9240 $\pm$ 0.0011 & 3.181 $\pm$ 0.203 \\
   &  & \ModelName{} & Learned & 409 & 3.794 $\pm$ 0.399 & 1.342 $\pm$ 0.186 & 6.375 $\pm$ 0.207 & 2.363 $\pm$ 0.137 & 0.9929 $\pm$ 0.0002 & 0.9892 $\pm$ 0.0003 & 0.9241 $\pm$ 0.0018 & 2.581 $\pm$ 0.293 \\
    \cmidrule(lr){2-13}
   & \multirow{6}{*}{[M+Na]+} & GraphCCS & — & 504 & — & — & 7.170 $\pm$ 0.202 & 2.826 $\pm$ 0.070 & 0.9903 $\pm$ 0.0004 & 0.9856 $\pm$ 0.0006 & 0.9004 $\pm$ 0.0024 & — \\
   &  & SigmaCCS & — & 504 & 5.464 $\pm$ 0.707 & 2.173 $\pm$ 0.303 & 7.368 $\pm$ 0.295 & 2.898 $\pm$ 0.139 & 0.9889 $\pm$ 0.0009 & 0.9841 $\pm$ 0.0011 & 0.8958 $\pm$ 0.0040 & 1.904 $\pm$ 0.798 \\
   &  & \ModelName{} & Single & 504 & 2.984 $\pm$ 0.162 & 1.004 $\pm$ 0.055 & 7.008 $\pm$ 0.119 & \textbf{2.811} $\pm$ 0.042 & 0.9913 $\pm$ 0.0002 & 0.9866 $\pm$ 0.0004 & 0.9040 $\pm$ 0.0013 & 4.024 $\pm$ 0.087 \\
   &  & \ModelName{} & Uniform & 504 & 3.258 $\pm$ 0.274 & 1.084 $\pm$ 0.101 & 7.218 $\pm$ 0.321 & 2.886 $\pm$ 0.142 & 0.9912 $\pm$ 0.0003 & 0.9867 $\pm$ 0.0005 & 0.9049 $\pm$ 0.0017 & 3.960 $\pm$ 0.222 \\
   &  & \ModelName{} & Boltzmann & 504 & 2.981 $\pm$ 0.232 & 1.012 $\pm$ 0.086 & 7.374 $\pm$ 0.034 & 2.947 $\pm$ 0.023 & 0.9909 $\pm$ 0.0003 & 0.9865 $\pm$ 0.0006 & 0.9048 $\pm$ 0.0023 & 4.393 $\pm$ 0.216 \\
   &  & \ModelName{} & Learned & 504 & 3.693 $\pm$ 0.590 & 1.293 $\pm$ 0.237 & 7.112 $\pm$ 0.248 & 2.816 $\pm$ 0.102 & 0.9909 $\pm$ 0.0003 & 0.9863 $\pm$ 0.0006 & 0.9033 $\pm$ 0.0019 & 3.419 $\pm$ 0.502 \\
    \bottomrule
  \end{tabular}
\end{adjustbox}
\end{table}

\subsection{External Benchmark Results Across All Configurations and Splits}
\label{sec:supp_external}

Table~\ref{tab:result_external_full} and Table~\ref{tab:main_external}   report CCS
prediction on all four external test sets for all \ModelName{} conformer pooling modes
and competitors across all training splits.

\begin{table}[H]
  \centering
  % Old: External test set: all \ModelName{} pooling modes and competitors. Bold: lowest MPD per (testset, split).
  \caption{CCS prediction on four independent external test sets across all \ModelName{} conformer pooling modes and competitors, stratified by training split strategy. Because no validation signal is available to select the optimal split for external evaluation, all training split/pooling combinations are reported for transparency. Mean $\pm$ std across 5 random seeds. $^*$Testset~4: the molecule \texttt{[H]/N=C(/N)NCC(=O)O} contains an explicit hydrogen absent from the SigmaCCS atom vocabulary; SigmaCCS is therefore evaluated on 19 of the 20 testset molecules ($n = 19$).}
  \label{tab:result_external_full}
  \begin{adjustbox}{max width=\linewidth, max totalheight=0.85\textheight, keepaspectratio}
\begin{tabular}{lll l r rrr}
    \toprule
    & & & & & \multicolumn{3}{c}{Test} \\
    \cmidrule(lr){6-8}
    Testset & Split & Model & Pooling & $n$ & RMSE & MPD & Pearson $R$ \\
    \midrule
  \multirow{18}{*}{Testset 1} & \multirow{6}{*}{Random} & GraphCCS & — & 1163 & 5.620 $\pm$ 0.157 & 2.350 $\pm$ 0.058 & 0.9840 $\pm$ 0.0010 \\
   &  & SigmaCCS & — & 1163 & 5.630 $\pm$ 0.209 & 2.342 $\pm$ 0.110 & 0.9841 $\pm$ 0.0008 \\
   &  & \ModelName{} & Single & 1163 & 5.467 $\pm$ 0.108 & \textbf{2.138} $\pm$ 0.035 & 0.9848 $\pm$ 0.0006 \\
   &  & \ModelName{} & Uniform & 1163 & 5.838 $\pm$ 0.217 & 2.327 $\pm$ 0.045 & 0.9827 $\pm$ 0.0011 \\
   &  & \ModelName{} & Boltzmann & 1163 & 5.504 $\pm$ 0.080 & 2.201 $\pm$ 0.036 & 0.9847 $\pm$ 0.0004 \\
   &  & \ModelName{} & Learned & 1163 & 5.854 $\pm$ 0.179 & 2.321 $\pm$ 0.063 & 0.9827 $\pm$ 0.0010 \\
   & \multirow{6}{*}{Scaffold} & GraphCCS & — & 1163 & 5.605 $\pm$ 0.236 & 2.354 $\pm$ 0.097 & 0.9841 $\pm$ 0.0013 \\
   &  & SigmaCCS & — & 1163 & 6.451 $\pm$ 0.519 & 2.954 $\pm$ 0.402 & 0.9806 $\pm$ 0.0019 \\
   &  & \ModelName{} & Single & 1163 & 5.377 $\pm$ 0.102 & \textbf{2.147} $\pm$ 0.015 & 0.9852 $\pm$ 0.0005 \\
   &  & \ModelName{} & Uniform & 1163 & 5.541 $\pm$ 0.159 & 2.216 $\pm$ 0.051 & 0.9843 $\pm$ 0.0009 \\
   &  & \ModelName{} & Boltzmann & 1163 & 5.423 $\pm$ 0.119 & 2.180 $\pm$ 0.038 & 0.9849 $\pm$ 0.0006 \\
   &  & \ModelName{} & Learned & 1163 & 5.590 $\pm$ 0.137 & 2.245 $\pm$ 0.041 & 0.9840 $\pm$ 0.0008 \\
   & \multirow{6}{*}{Adduct-sens.} & GraphCCS & — & 1163 & 5.504 $\pm$ 0.079 & 2.280 $\pm$ 0.046 & 0.9845 $\pm$ 0.0004 \\
   &  & SigmaCCS & — & 1163 & 6.177 $\pm$ 0.265 & 2.688 $\pm$ 0.221 & 0.9816 $\pm$ 0.0014 \\
   &  & \ModelName{} & Single & 1163 & 5.745 $\pm$ 0.058 & 2.217 $\pm$ 0.017 & 0.9831 $\pm$ 0.0003 \\
   &  & \ModelName{} & Uniform & 1163 & 5.800 $\pm$ 0.083 & 2.287 $\pm$ 0.034 & 0.9827 $\pm$ 0.0005 \\
   &  & \ModelName{} & Boltzmann & 1163 & 5.500 $\pm$ 0.091 & \textbf{2.214} $\pm$ 0.036 & 0.9844 $\pm$ 0.0005 \\
   &  & \ModelName{} & Learned & 1163 & 5.800 $\pm$ 0.088 & 2.331 $\pm$ 0.057 & 0.9830 $\pm$ 0.0005 \\
    \midrule
  \multirow{18}{*}{Testset 2} & \multirow{6}{*}{Random} & GraphCCS & — & 49 & 11.439 $\pm$ 1.142 & 3.390 $\pm$ 0.278 & 0.9793 $\pm$ 0.0048 \\
   &  & SigmaCCS & — & 49 & 11.420 $\pm$ 0.725 & \textbf{3.386} $\pm$ 0.304 & 0.9821 $\pm$ 0.0030 \\
   &  & \ModelName{} & Single & 49 & 12.555 $\pm$ 0.679 & 3.510 $\pm$ 0.164 & 0.9779 $\pm$ 0.0031 \\
   &  & \ModelName{} & Uniform & 49 & 12.747 $\pm$ 0.799 & 3.503 $\pm$ 0.250 & 0.9803 $\pm$ 0.0016 \\
   &  & \ModelName{} & Boltzmann & 49 & 12.434 $\pm$ 0.820 & 3.457 $\pm$ 0.240 & 0.9790 $\pm$ 0.0027 \\
   &  & \ModelName{} & Learned & 49 & 13.262 $\pm$ 0.496 & 3.574 $\pm$ 0.172 & 0.9785 $\pm$ 0.0024 \\
   & \multirow{6}{*}{Scaffold} & GraphCCS & — & 49 & 14.034 $\pm$ 2.854 & 4.339 $\pm$ 0.934 & 0.9742 $\pm$ 0.0077 \\
   &  & SigmaCCS & — & 49 & 12.940 $\pm$ 1.112 & 4.024 $\pm$ 0.475 & 0.9797 $\pm$ 0.0023 \\
   &  & \ModelName{} & Single & 49 & 12.251 $\pm$ 0.852 & \textbf{3.239} $\pm$ 0.174 & 0.9805 $\pm$ 0.0015 \\
   &  & \ModelName{} & Uniform & 49 & 12.684 $\pm$ 0.410 & 3.336 $\pm$ 0.122 & 0.9793 $\pm$ 0.0012 \\
   &  & \ModelName{} & Boltzmann & 49 & 12.363 $\pm$ 0.743 & 3.467 $\pm$ 0.166 & 0.9787 $\pm$ 0.0018 \\
   &  & \ModelName{} & Learned & 49 & 13.045 $\pm$ 0.618 & 3.417 $\pm$ 0.192 & 0.9784 $\pm$ 0.0010 \\
   & \multirow{6}{*}{Adduct-sens.} & GraphCCS & — & 49 & 12.473 $\pm$ 1.594 & 3.899 $\pm$ 0.433 & 0.9818 $\pm$ 0.0013 \\
   &  & SigmaCCS & — & 49 & 11.811 $\pm$ 1.037 & 3.642 $\pm$ 0.424 & 0.9803 $\pm$ 0.0016 \\
   &  & \ModelName{} & Single & 49 & 10.641 $\pm$ 0.785 & \textbf{3.021} $\pm$ 0.141 & 0.9844 $\pm$ 0.0016 \\
   &  & \ModelName{} & Uniform & 49 & 12.401 $\pm$ 0.848 & 3.537 $\pm$ 0.205 & 0.9778 $\pm$ 0.0022 \\
   &  & \ModelName{} & Boltzmann & 49 & 11.839 $\pm$ 0.396 & 3.476 $\pm$ 0.119 & 0.9797 $\pm$ 0.0012 \\
   &  & \ModelName{} & Learned & 49 & 12.207 $\pm$ 0.239 & 3.401 $\pm$ 0.068 & 0.9784 $\pm$ 0.0009 \\
    \midrule
  \multirow{18}{*}{Testset 3} & \multirow{6}{*}{Random} & GraphCCS & — & 298 & 6.587 $\pm$ 0.410 & 2.249 $\pm$ 0.045 & 0.9945 $\pm$ 0.0007 \\
   &  & SigmaCCS & — & 298 & 9.042 $\pm$ 1.443 & 2.819 $\pm$ 0.197 & 0.9892 $\pm$ 0.0032 \\
   &  & \ModelName{} & Single & 298 & 6.530 $\pm$ 0.086 & 2.351 $\pm$ 0.040 & 0.9952 $\pm$ 0.0001 \\
   &  & \ModelName{} & Uniform & 298 & 6.155 $\pm$ 0.319 & \textbf{2.229} $\pm$ 0.102 & 0.9955 $\pm$ 0.0002 \\
   &  & \ModelName{} & Boltzmann & 298 & 6.273 $\pm$ 0.205 & 2.263 $\pm$ 0.064 & 0.9956 $\pm$ 0.0002 \\
   &  & \ModelName{} & Learned & 298 & 6.125 $\pm$ 0.173 & 2.240 $\pm$ 0.062 & 0.9954 $\pm$ 0.0002 \\
   & \multirow{6}{*}{Scaffold} & GraphCCS & — & 298 & 6.945 $\pm$ 0.616 & 2.317 $\pm$ 0.112 & 0.9934 $\pm$ 0.0011 \\
   &  & SigmaCCS & — & 298 & 9.039 $\pm$ 1.066 & 3.198 $\pm$ 0.318 & 0.9896 $\pm$ 0.0020 \\
   &  & \ModelName{} & Single & 298 & 6.037 $\pm$ 0.083 & 2.191 $\pm$ 0.024 & 0.9954 $\pm$ 0.0001 \\
   &  & \ModelName{} & Uniform & 298 & 6.237 $\pm$ 0.304 & 2.198 $\pm$ 0.053 & 0.9951 $\pm$ 0.0004 \\
   &  & \ModelName{} & Boltzmann & 298 & 6.334 $\pm$ 0.199 & 2.252 $\pm$ 0.057 & 0.9951 $\pm$ 0.0004 \\
   &  & \ModelName{} & Learned & 298 & 6.127 $\pm$ 0.283 & \textbf{2.150} $\pm$ 0.082 & 0.9951 $\pm$ 0.0004 \\
   & \multirow{6}{*}{Adduct-sens.} & GraphCCS & — & 298 & 7.458 $\pm$ 0.316 & 2.552 $\pm$ 0.070 & 0.9930 $\pm$ 0.0006 \\
   &  & SigmaCCS & — & 298 & 11.976 $\pm$ 1.515 & 3.496 $\pm$ 0.263 & 0.9810 $\pm$ 0.0054 \\
   &  & \ModelName{} & Single & 298 & 7.376 $\pm$ 0.128 & 2.661 $\pm$ 0.040 & 0.9940 $\pm$ 0.0001 \\
   &  & \ModelName{} & Uniform & 298 & 6.992 $\pm$ 0.224 & 2.497 $\pm$ 0.075 & 0.9943 $\pm$ 0.0002 \\
   &  & \ModelName{} & Boltzmann & 298 & 7.288 $\pm$ 0.157 & 2.558 $\pm$ 0.076 & 0.9942 $\pm$ 0.0003 \\
   &  & \ModelName{} & Learned & 298 & 6.840 $\pm$ 0.184 & \textbf{2.464} $\pm$ 0.076 & 0.9943 $\pm$ 0.0002 \\
    \midrule
  \multirow{18}{*}{Testset 4\textsuperscript{*}} & \multirow{6}{*}{Random} & GraphCCS & — & 20 & 3.193 $\pm$ 0.268 & 1.798 $\pm$ 0.184 & 0.9613 $\pm$ 0.0070 \\
   &  & SigmaCCS & — & 19 & 3.623 $\pm$ 0.279 & 1.836 $\pm$ 0.145 & 0.9466 $\pm$ 0.0079 \\
   &  & \ModelName{} & Single & 20 & 2.268 $\pm$ 0.095 & \textbf{1.340} $\pm$ 0.066 & 0.9850 $\pm$ 0.0006 \\
   &  & \ModelName{} & Uniform & 20 & 2.983 $\pm$ 0.298 & 1.701 $\pm$ 0.146 & 0.9763 $\pm$ 0.0039 \\
   &  & \ModelName{} & Boltzmann & 20 & 2.914 $\pm$ 0.224 & 1.741 $\pm$ 0.120 & 0.9754 $\pm$ 0.0046 \\
   &  & \ModelName{} & Learned & 20 & 2.851 $\pm$ 0.247 & 1.582 $\pm$ 0.177 & 0.9786 $\pm$ 0.0028 \\
   & \multirow{6}{*}{Scaffold} & GraphCCS & — & 20 & 3.640 $\pm$ 0.180 & 2.122 $\pm$ 0.126 & 0.9488 $\pm$ 0.0096 \\
   &  & SigmaCCS & — & 19 & 4.898 $\pm$ 1.012 & 2.979 $\pm$ 0.725 & 0.9381 $\pm$ 0.0227 \\
   &  & \ModelName{} & Single & 20 & 3.281 $\pm$ 0.193 & \textbf{1.962} $\pm$ 0.139 & 0.9604 $\pm$ 0.0041 \\
   &  & \ModelName{} & Uniform & 20 & 3.500 $\pm$ 0.275 & 2.106 $\pm$ 0.141 & 0.9548 $\pm$ 0.0052 \\
   &  & \ModelName{} & Boltzmann & 20 & 3.525 $\pm$ 0.186 & 2.101 $\pm$ 0.164 & 0.9519 $\pm$ 0.0052 \\
   &  & \ModelName{} & Learned & 20 & 3.539 $\pm$ 0.374 & 2.171 $\pm$ 0.214 & 0.9573 $\pm$ 0.0065 \\
   & \multirow{6}{*}{Adduct-sens.} & GraphCCS & — & 20 & 3.277 $\pm$ 0.259 & 2.026 $\pm$ 0.166 & 0.9613 $\pm$ 0.0086 \\
   &  & SigmaCCS & — & 19 & 4.285 $\pm$ 0.954 & 2.613 $\pm$ 0.681 & 0.9424 $\pm$ 0.0183 \\
   &  & \ModelName{} & Single & 20 & 3.193 $\pm$ 0.160 & \textbf{1.710} $\pm$ 0.132 & 0.9673 $\pm$ 0.0026 \\
   &  & \ModelName{} & Uniform & 20 & 3.833 $\pm$ 0.243 & 2.291 $\pm$ 0.186 & 0.9641 $\pm$ 0.0054 \\
   &  & \ModelName{} & Boltzmann & 20 & 3.307 $\pm$ 0.261 & 1.786 $\pm$ 0.128 & 0.9648 $\pm$ 0.0043 \\
   &  & \ModelName{} & Learned & 20 & 3.540 $\pm$ 0.492 & 2.152 $\pm$ 0.289 & 0.9682 $\pm$ 0.0031 \\
    \bottomrule
  \end{tabular}
\end{adjustbox}
\end{table}

\begin{table}[H]
  \centering
  \caption{\ModelName{} generalises to four independent external test sets, achieving lower RMSE than GraphCCS and SigmaCCS in the majority of configurations. \ModelName{} reports results for the conformer pooling mode minimising RMSE per (split, testset) combination (superscript). Mean $\pm$ std across 5 random seeds. $^*$Testset~4: the molecule \texttt{[H]/N=C(/N)NCC(=O)O} contains an explicit hydrogen absent from the SigmaCCS atom vocabulary; SigmaCCS is therefore evaluated on 19 of the 20 testset molecules ($n = 19$).}
  \label{tab:main_external}
  \begin{adjustbox}{max width=\linewidth, max totalheight=0.85\textheight, keepaspectratio}
\begin{tabular}{llr r@{$\,\pm\,$}l r@{$\,\pm\,$}l r@{$\,\pm\,$}l}
    \toprule
    Split & Set & $n$
      & \multicolumn{2}{c}{GraphCCS}
      & \multicolumn{2}{c}{SigmaCCS}
      & \multicolumn{2}{c}{\ModelName{} (ours)} \\
    \midrule
  \multirow{4}{*}{Random}
   & TS1                    & 1163 &  5.620 & 0.157 &  5.630 & 0.209 & \textbf{5.467}  & 0.108\textsuperscript{(Single)} \\
   & TS2                    &   49 & 11.439 & 1.142 & \textbf{11.420} & 0.725 & 12.434 & 0.820\textsuperscript{(Boltzmann)} \\
   & TS3                    &  298 &  6.587 & 0.410 &  9.042 & 1.443 & \textbf{6.155}  & 0.319\textsuperscript{(Uniform)} \\
   & TS4\textsuperscript{*} &   20 &  3.193 & 0.268 &  3.623 & 0.279 & \textbf{2.268}  & 0.095\textsuperscript{(Single)} \\
    \midrule
  \multirow{4}{*}{Scaffold}
   & TS1                    & 1163 &  5.605 & 0.236 &  6.451 & 0.519 & \textbf{5.377}  & 0.102\textsuperscript{(Single)} \\
   & TS2                    &   49 & 14.034 & 2.854 & 12.940 & 1.112 & \textbf{12.251} & 0.852\textsuperscript{(Single)} \\
   & TS3                    &  298 &  6.945 & 0.616 &  9.039 & 1.066 & \textbf{6.127}  & 0.283\textsuperscript{(Learned)} \\
   & TS4\textsuperscript{*} &   20 &  3.640 & 0.180 &  4.898 & 1.012 & \textbf{3.281}  & 0.193\textsuperscript{(Single)} \\
    \midrule
  \multirow{4}{*}{Adduct-sens.}
   & TS1                    & 1163 &  5.504 & 0.079 &  6.177 & 0.265 & \textbf{5.500}  & 0.091\textsuperscript{(Boltzmann)} \\
   & TS2                    &   49 & 12.473 & 1.594 & 11.811 & 1.037 & \textbf{10.641} & 0.785\textsuperscript{(Single)} \\
   & TS3                    &  298 &  7.458 & 0.316 & 11.976 & 1.515 & \textbf{6.840}  & 0.184\textsuperscript{(Learned)} \\
   & TS4\textsuperscript{*} &   20 &  3.277 & 0.259 &  4.285 & 0.954 & \textbf{3.193}  & 0.160\textsuperscript{(Single)} \\
    \bottomrule
  \end{tabular}
\end{adjustbox}
\end{table}

\noindent Per-adduct breakdown for the best-performing external configuration is given in
Table~\ref{tab:result_external_best_breakdown}.

\begin{table}[H]
  \centering
  % Old: External test set per-adduct breakdown: best model configuration per testset.
  \caption{Per-adduct CCS prediction accuracy on four independent external test sets for the best-performing model configuration per testset (training split and pooling as in Table~\ref{tab:result_external_best}). \ModelName{} shows consistent gains over competitors on \mbox{[M+H]$^+$} and \mbox{[M+Na]$^+$} adducts across all testsets. Mean $\pm$ std across 5 random seeds. $^*$Testset~4: the molecule \texttt{[H]/N=C(/N)NCC(=O)O} contains an explicit hydrogen absent from the SigmaCCS atom vocabulary; SigmaCCS is therefore evaluated on 19 of the 20 testset molecules ($n = 19$).}
  \label{tab:result_external_best_breakdown}
  \begin{adjustbox}{max width=\linewidth, max totalheight=0.85\textheight, keepaspectratio}
\begin{tabular}{ll ll r rr}
    \toprule
    & & & & & \multicolumn{2}{c}{Test} \\
    \cmidrule(lr){6-7}
    Testset & Adduct & Model & Best Config & $n$ & RMSE & MPD \\
    \midrule
  \multirow{9}{*}{Testset 1} & \multirow{3}{*}{[M+H]+} & GraphCCS & Adduct-sens. & 412 & 4.890 $\pm$ 0.159 & 2.092 $\pm$ 0.086 \\
   &  & SigmaCCS & Random & 412 & 4.943 $\pm$ 0.169 & 2.118 $\pm$ 0.103 \\
   &  & \ModelName{} (Single) & Single, Random & 412 & 4.362 $\pm$ 0.070 & \textbf{1.867} $\pm$ 0.033 \\
    \cmidrule(lr){2-7}
   & \multirow{3}{*}{[M-H]-} & GraphCCS & Adduct-sens. & 452 & 4.850 $\pm$ 0.111 & \textbf{2.123} $\pm$ 0.071 \\
   &  & SigmaCCS & Random & 452 & 5.308 $\pm$ 0.213 & 2.290 $\pm$ 0.104 \\
   &  & \ModelName{} (Single) & Single, Random & 452 & 5.317 $\pm$ 0.210 & 2.134 $\pm$ 0.054 \\
    \cmidrule(lr){2-7}
   & \multirow{3}{*}{[M+Na]+} & GraphCCS & Adduct-sens. & 299 & 7.018 $\pm$ 0.197 & 2.775 $\pm$ 0.112 \\
   &  & SigmaCCS & Random & 299 & 6.853 $\pm$ 0.354 & 2.731 $\pm$ 0.193 \\
   &  & \ModelName{} (Single) & Single, Random & 299 & 6.874 $\pm$ 0.149 & \textbf{2.519} $\pm$ 0.081 \\
    \midrule
  \multirow{9}{*}{Testset 2} & \multirow{3}{*}{[M+H]+} & GraphCCS & Random & 15 & 10.112 $\pm$ 2.717 & 3.078 $\pm$ 0.905 \\
   &  & SigmaCCS & Random & 15 & 6.465 $\pm$ 1.660 & 2.021 $\pm$ 0.534 \\
   &  & \ModelName{} (Single) & Single, Adduct-sens. & 15 & 4.308 $\pm$ 0.761 & \textbf{1.350} $\pm$ 0.234 \\
    \cmidrule(lr){2-7}
   & \multirow{3}{*}{[M-H]-} & GraphCCS & Random & 19 & 13.627 $\pm$ 1.457 & \textbf{4.462} $\pm$ 0.351 \\
   &  & SigmaCCS & Random & 19 & 15.463 $\pm$ 1.812 & 5.324 $\pm$ 0.605 \\
   &  & \ModelName{} (Single) & Single, Adduct-sens. & 19 & 15.212 $\pm$ 1.429 & 5.115 $\pm$ 0.451 \\
    \cmidrule(lr){2-7}
   & \multirow{3}{*}{[M+Na]+} & GraphCCS & Random & 15 & 8.979 $\pm$ 1.878 & 2.346 $\pm$ 0.462 \\
   &  & SigmaCCS & Random & 15 & 8.486 $\pm$ 2.047 & 2.298 $\pm$ 0.699 \\
   &  & \ModelName{} (Single) & Single, Adduct-sens. & 15 & 7.509 $\pm$ 0.813 & \textbf{2.039} $\pm$ 0.225 \\
    \midrule
  \multirow{9}{*}{Testset 3} & \multirow{3}{*}{[M+H]+} & GraphCCS & Random & 148 & 6.595 $\pm$ 1.006 & 1.972 $\pm$ 0.070 \\
   &  & SigmaCCS & Random & 148 & 10.280 $\pm$ 2.491 & 2.717 $\pm$ 0.302 \\
   &  & \ModelName{} (Learned) & Learned, Scaffold & 148 & 5.528 $\pm$ 0.415 & \textbf{1.811} $\pm$ 0.142 \\
    \cmidrule(lr){2-7}
   & \multirow{3}{*}{[M-H]-} & GraphCCS & Random & 75 & 5.204 $\pm$ 0.376 & \textbf{2.555} $\pm$ 0.107 \\
   &  & SigmaCCS & Random & 75 & 6.957 $\pm$ 0.474 & 3.173 $\pm$ 0.222 \\
   &  & \ModelName{} (Learned) & Learned, Scaffold & 75 & 5.519 $\pm$ 0.272 & 2.575 $\pm$ 0.131 \\
    \cmidrule(lr){2-7}
   & \multirow{3}{*}{[M+Na]+} & GraphCCS & Random & 75 & 7.605 $\pm$ 0.430 & 2.490 $\pm$ 0.209 \\
   &  & SigmaCCS & Random & 75 & 7.976 $\pm$ 0.259 & 2.668 $\pm$ 0.164 \\
   &  & \ModelName{} (Learned) & Learned, Scaffold & 75 & 7.626 $\pm$ 0.367 & \textbf{2.395} $\pm$ 0.066 \\
    \midrule
  \multirow{6}{*}{Testset 4\textsuperscript{*}} & \multirow{3}{*}{[M+H]+} & GraphCCS & Random & 17 & 3.356 $\pm$ 0.313 & 1.900 $\pm$ 0.241 \\
   &  & SigmaCCS & Random & 16 & 3.783 $\pm$ 0.298 & 1.852 $\pm$ 0.127 \\
   &  & \ModelName{} (Single) & Single, Random & 17 & 2.390 $\pm$ 0.138 & \textbf{1.425} $\pm$ 0.082 \\
    \cmidrule(lr){2-7}
   & \multirow{3}{*}{[M-H]-} & GraphCCS & Random & 3 & 1.984 $\pm$ 0.368 & 1.220 $\pm$ 0.255 \\
   &  & SigmaCCS & Random & 3 & 2.566 $\pm$ 0.478 & 1.755 $\pm$ 0.347 \\
   &  & \ModelName{} (Single) & Single, Random & 3 & 1.288 $\pm$ 0.480 & \textbf{0.859} $\pm$ 0.301 \\
    \cmidrule(lr){2-7}
    \bottomrule
  \end{tabular}
\end{adjustbox}
\end{table}

\subsection{Late Fusion Results Across All Pooling Modes and Token Representations}
\label{sec:supp_late_fusion}

Table~\ref{tab:late_fusion_ablation} reports the late-fusion counterpart to
Table~\ref{tab:ccs3d_graphccs_main}, enabling direct comparison of early versus late adduct
conditioning under matched pooling configurations.

\begin{table}[H]
  \centering
  \caption{\ModelName{} late-fusion ablation on the \Dataset{} test set, with GraphCCS and SigmaCCS reference rows at the top of each split. Late fusion appends adduct identity after the geometry encoder backbone, in contrast to early fusion where the adduct token is injected directly into the encoder. Three conformer pooling strategies (Single, Uniform, Boltzmann) and two token representations (CLS, Gasteiger) are evaluated under each training split. Pooling and Repr are not applicable ({---}) for GraphCCS and SigmaCCS.}
  \label{tab:late_fusion_ablation}
\begin{adjustbox}{max width=\linewidth, max totalheight=0.85\textheight, keepaspectratio}
\begin{tabular}{lll r rr rr rrrr}
    \toprule
    Split & Pooling & Repr & $n$ & Train RMSE & Train MPD & Test RMSE & Test MPD & Test Pearson $R$ & Test Spearman $\rho$ & Test Kendall $\tau$ & Gap \\
    \midrule
  \multirow{8}{*}{Random} & GraphCCS & — & \multirow{8}{*}{922} & 3.287 $\pm$ 0.460 & 1.150 $\pm$ 0.171 & 4.858 $\pm$ 0.112 & \textbf{1.766} $\pm$ 0.050 & 0.9960 $\pm$ 0.0002 & 0.9938 $\pm$ 0.0003 & 0.9399 $\pm$ 0.0016 & 1.571 $\pm$ 0.494 \\
   & SigmaCCS & — &  & 4.664 $\pm$ 0.390 & 1.728 $\pm$ 0.102 & 5.137 $\pm$ 0.166 & 1.950 $\pm$ 0.051 & 0.9955 $\pm$ 0.0003 & 0.9932 $\pm$ 0.0004 & 0.9341 $\pm$ 0.0019 & 0.473 $\pm$ 0.251 \\
   & Single & CLS &  & 5.092 $\pm$ 0.848 & 1.853 $\pm$ 0.232 & 5.847 $\pm$ 0.273 & 2.263 $\pm$ 0.102 & 0.9945 $\pm$ 0.0005 & 0.9923 $\pm$ 0.0005 & 0.9276 $\pm$ 0.0024 & 0.755 $\pm$ 0.608 \\
   & Single & Gasteiger &  & 4.872 $\pm$ 0.446 & 1.853 $\pm$ 0.173 & 5.885 $\pm$ 0.164 & 2.299 $\pm$ 0.095 & 0.9944 $\pm$ 0.0002 & 0.9922 $\pm$ 0.0003 & 0.9271 $\pm$ 0.0016 & 1.014 $\pm$ 0.332 \\
   & Uniform & CLS &  & 4.383 $\pm$ 1.033 & 1.607 $\pm$ 0.294 & 5.385 $\pm$ 0.386 & 2.099 $\pm$ 0.153 & 0.9952 $\pm$ 0.0007 & 0.9930 $\pm$ 0.0008 & 0.9317 $\pm$ 0.0044 & 1.002 $\pm$ 0.710 \\
   & Uniform & Gasteiger &  & 4.067 $\pm$ 0.567 & 1.543 $\pm$ 0.213 & 5.296 $\pm$ 0.288 & 2.062 $\pm$ 0.108 & 0.9953 $\pm$ 0.0005 & 0.9931 $\pm$ 0.0007 & 0.9321 $\pm$ 0.0037 & 1.229 $\pm$ 0.329 \\
   & Boltzmann & CLS &  & 4.160 $\pm$ 0.153 & 1.577 $\pm$ 0.045 & 5.505 $\pm$ 0.093 & 2.155 $\pm$ 0.026 & 0.9950 $\pm$ 0.0001 & 0.9928 $\pm$ 0.0001 & 0.9297 $\pm$ 0.0004 & 1.344 $\pm$ 0.225 \\
   & Boltzmann & Gasteiger &  & 4.038 $\pm$ 0.320 & 1.527 $\pm$ 0.130 & 5.538 $\pm$ 0.137 & 2.156 $\pm$ 0.056 & 0.9949 $\pm$ 0.0003 & 0.9926 $\pm$ 0.0004 & 0.9287 $\pm$ 0.0021 & 1.500 $\pm$ 0.198 \\
    \midrule
  \multirow{8}{*}{Scaffold} & GraphCCS & — & \multirow{8}{*}{920} & 3.844 $\pm$ 0.921 & 1.357 $\pm$ 0.299 & 6.396 $\pm$ 0.282 & 2.278 $\pm$ 0.182 & 0.9925 $\pm$ 0.0004 & 0.9907 $\pm$ 0.0003 & 0.9235 $\pm$ 0.0023 & 2.552 $\pm$ 0.718 \\
   & SigmaCCS & — &  & 6.771 $\pm$ 1.183 & 2.534 $\pm$ 0.450 & 6.867 $\pm$ 0.693 & 2.795 $\pm$ 0.416 & 0.9915 $\pm$ 0.0015 & 0.9886 $\pm$ 0.0017 & 0.9126 $\pm$ 0.0072 & 0.096 $\pm$ 0.696 \\
   & Single & CLS &  & 4.330 $\pm$ 0.454 & 1.617 $\pm$ 0.191 & 6.122 $\pm$ 0.164 & 2.309 $\pm$ 0.108 & 0.9934 $\pm$ 0.0003 & 0.9911 $\pm$ 0.0006 & 0.9229 $\pm$ 0.0029 & 1.792 $\pm$ 0.376 \\
   & Single & Gasteiger &  & 4.222 $\pm$ 0.531 & 1.607 $\pm$ 0.190 & 6.158 $\pm$ 0.285 & 2.356 $\pm$ 0.133 & 0.9933 $\pm$ 0.0005 & 0.9907 $\pm$ 0.0005 & 0.9212 $\pm$ 0.0027 & 1.936 $\pm$ 0.297 \\
   & Uniform & CLS &  & 4.217 $\pm$ 0.427 & 1.605 $\pm$ 0.171 & 6.246 $\pm$ 0.140 & 2.276 $\pm$ 0.084 & 0.9929 $\pm$ 0.0002 & 0.9909 $\pm$ 0.0006 & 0.9229 $\pm$ 0.0031 & 2.029 $\pm$ 0.340 \\
   & Uniform & Gasteiger &  & 4.363 $\pm$ 0.486 & 1.669 $\pm$ 0.204 & 6.847 $\pm$ 0.912 & 2.377 $\pm$ 0.142 & 0.9912 $\pm$ 0.0027 & 0.9895 $\pm$ 0.0011 & 0.9181 $\pm$ 0.0042 & 2.484 $\pm$ 0.759 \\
   & Boltzmann & CLS &  & 3.826 $\pm$ 0.219 & 1.438 $\pm$ 0.098 & 6.068 $\pm$ 0.125 & \textbf{2.220} $\pm$ 0.034 & 0.9936 $\pm$ 0.0003 & 0.9914 $\pm$ 0.0001 & 0.9255 $\pm$ 0.0003 & 2.243 $\pm$ 0.160 \\
   & Boltzmann & Gasteiger &  & 4.275 $\pm$ 0.637 & 1.626 $\pm$ 0.236 & 6.138 $\pm$ 0.248 & 2.352 $\pm$ 0.128 & 0.9932 $\pm$ 0.0005 & 0.9905 $\pm$ 0.0004 & 0.9212 $\pm$ 0.0026 & 1.863 $\pm$ 0.470 \\
    \midrule
  \multirow{8}{*}{Adduct-sens.} & GraphCCS & — & \multirow{8}{*}{1382} & 4.117 $\pm$ 0.400 & 1.373 $\pm$ 0.086 & 6.671 $\pm$ 0.076 & \textbf{2.512} $\pm$ 0.038 & 0.9911 $\pm$ 0.0004 & 0.9875 $\pm$ 0.0005 & 0.9105 $\pm$ 0.0021 & 2.554 $\pm$ 0.385 \\
   & SigmaCCS & — &  & 5.263 $\pm$ 0.916 & 2.099 $\pm$ 0.379 & 6.918 $\pm$ 0.280 & 2.756 $\pm$ 0.193 & 0.9899 $\pm$ 0.0006 & 0.9858 $\pm$ 0.0012 & 0.9032 $\pm$ 0.0049 & 1.655 $\pm$ 0.783 \\
   & Single & CLS &  & 5.130 $\pm$ 0.815 & 1.883 $\pm$ 0.297 & 8.219 $\pm$ 0.182 & 3.288 $\pm$ 0.094 & 0.9863 $\pm$ 0.0007 & 0.9808 $\pm$ 0.0012 & 0.8832 $\pm$ 0.0041 & 3.089 $\pm$ 0.760 \\
   & Single & Gasteiger &  & 4.675 $\pm$ 0.735 & 1.805 $\pm$ 0.222 & 7.957 $\pm$ 0.036 & 3.167 $\pm$ 0.026 & 0.9868 $\pm$ 0.0002 & 0.9817 $\pm$ 0.0001 & 0.8866 $\pm$ 0.0008 & 3.281 $\pm$ 0.750 \\
   & Uniform & CLS &  & 4.381 $\pm$ 0.889 & 1.547 $\pm$ 0.191 & 8.813 $\pm$ 0.234 & 3.478 $\pm$ 0.071 & 0.9854 $\pm$ 0.0003 & 0.9793 $\pm$ 0.0006 & 0.8790 $\pm$ 0.0018 & 4.432 $\pm$ 0.937 \\
   & Uniform & Gasteiger &  & 4.516 $\pm$ 0.808 & 1.608 $\pm$ 0.190 & 8.571 $\pm$ 0.376 & 3.335 $\pm$ 0.133 & 0.9860 $\pm$ 0.0006 & 0.9806 $\pm$ 0.0005 & 0.8833 $\pm$ 0.0014 & 4.055 $\pm$ 0.520 \\
   & Boltzmann & CLS &  & 4.932 $\pm$ 1.021 & 1.728 $\pm$ 0.273 & 8.625 $\pm$ 0.111 & 3.435 $\pm$ 0.083 & 0.9853 $\pm$ 0.0007 & 0.9794 $\pm$ 0.0011 & 0.8786 $\pm$ 0.0037 & 3.693 $\pm$ 0.971 \\
   & Boltzmann & Gasteiger &  & 4.131 $\pm$ 0.522 & 1.585 $\pm$ 0.210 & 8.285 $\pm$ 0.096 & 3.280 $\pm$ 0.051 & 0.9858 $\pm$ 0.0003 & 0.9805 $\pm$ 0.0003 & 0.8823 $\pm$ 0.0012 & 4.154 $\pm$ 0.519 \\
    \bottomrule
  \end{tabular}
\end{adjustbox}
\end{table}

\subsection{Early vs.\ Late Fusion: Effect of Residual Learning}
\label{sec:supp_lf_ef}

Table~\ref{tab:lf-ef-ablation} tests how much of \ModelName{}'s generalisation comes from
residual learning versus encoder-level adduct conditioning.
All three variants share the same backbone; what changes is where the adduct enters and what
the model predicts.
Late fusion (LF) concatenates the adduct one-hot only at the MLP head and trains directly on
CCS, the adduct-na\"{i}ve encoder baseline.
LF+Residual uses the same architecture with the residual target.
That substitution is counterproductive: residual learning strips away the mass-CCS
correlation that otherwise anchors the encoder, but without an adduct token inside the
transformer there is nothing to replace it, so the backbone overfits to training-set geometry
(gap $+3.0$~\AA$^2$ on the random split).
Early fusion (EF, \ModelName{}) injects the adduct as a learned delta on the CLS token
before the first transformer layer;
LoRA-adapted attention then routes adduct information through the full encoder stack, and the
residual target is retained.
The gap falls to $+0.08$~\AA$^2$, a 37-fold reduction, suggesting that adduct-in-encoder
conditioning is necessary for residual learning to generalise.

\begin{table}[H]
  \centering
  \caption{Train and test RMSE (mean $\pm$ s.d.) and generalisation gap for three architecture variants on the Random and Adduct-sensitive splits. EF: 5~seeds; LF variants: 3~seeds. Bold entries are best per split.}
  \label{tab:lf-ef-ablation}
  \begin{tabular}{ll r r r r}
    \toprule
    Split & Model & $n$ & Train RMSE & Test RMSE & Gap \\
    \midrule
    \multirow{3}{*}{Random} & LF & 3 & 5.43 $\pm$ 0.69 & 5.89 $\pm$ 0.24 & +0.46 \\
     & LF+Residual & 3 & 1.85 $\pm$ 0.13 & 4.85 $\pm$ 0.06 & +3.00 \\
     & EF & 5 & \textbf{4.56 $\pm$ 0.10} & \textbf{4.64 $\pm$ 0.05} & \textbf{+0.08} \\
    \midrule
    \multirow{3}{*}{Adduct-sensitive} & LF & 3 & 6.95 $\pm$ 0.55 & 8.02 $\pm$ 0.20 & +1.07 \\
     & LF+Residual & 3 & 6.26 $\pm$ 0.35 & 7.16 $\pm$ 0.03 & +0.90 \\
     & EF & 5 & \textbf{4.78 $\pm$ 0.16} & \textbf{6.34 $\pm$ 0.08} & \textbf{+1.56} \\
    \bottomrule
  \end{tabular}
\end{table}

\subsection{From Late Fusion to \ModelName{}}
\label{sec:supp_design_progression}

\begin{figure}[htbp]
    \centering
    \includegraphics[width=\textwidth]{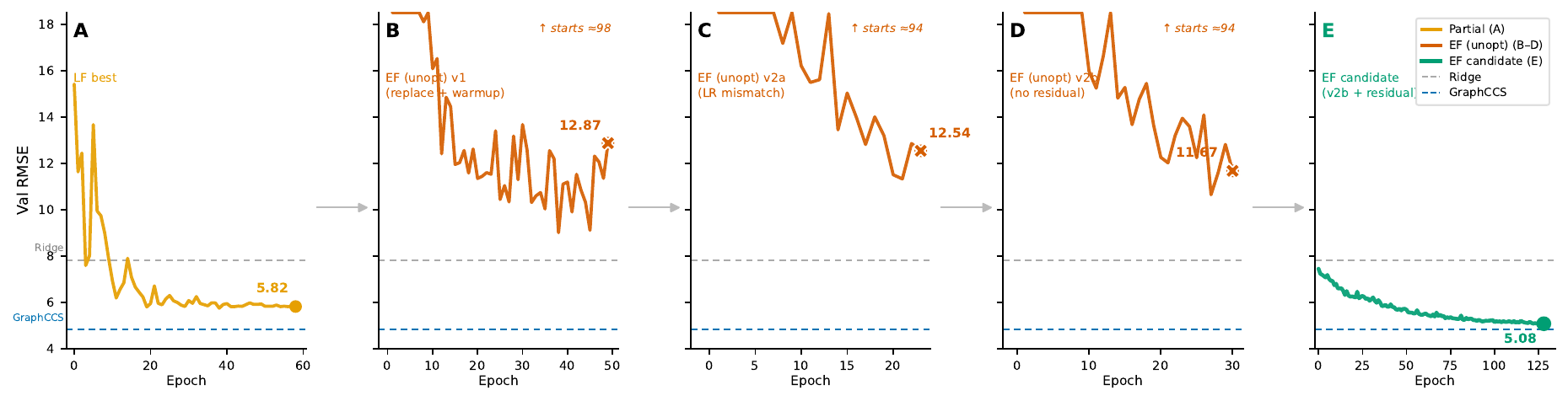}
    \caption{\textbf{From late fusion to \ModelName{}.}
      Validation RMSE on the random split (seed~0) across training epochs
      for five successive architecture candidates.
      Dashed grey line: Ridge descriptor baseline (7.82~\AA$^2$).
      Dashed blue line: GraphCCS reference (4.82~\AA$^2$).
      \texttimes{} marks runs stopped early before convergence.
      \textbf{(A)}~Late fusion best (uniform pooling, CLS token; labelled LF best):
        converges to 5.82~\AA$^2$ but leaves a 1.4~\AA$^2$ gap relative to
        GraphCCS on the adduct-sensitive split, motivating encoder-level
        adduct conditioning.
      \textbf{(B)}~Early fusion v1 with CLS token replacement, warmup, and
        gradient clipping: initialises at $\approx$98~\AA$^2$ and plateaus
        near 12.87~\AA$^2$; the pretrained CLS signal is discarded on every
        forward pass, destabilising all 15 transformer layers simultaneously.
      \textbf{(C)}~Early fusion v2a with additive delta and mismatched
        learning rates ($\delta$ at $10^{-4}$, LoRA at $10^{-5}$):
        initialises at $\approx$94~\AA$^2$, stopped at epoch~23 due to
        periodic validation spikes caused by representation drift.
      \textbf{(D)}~Early fusion v2b with additive delta co-trained with
        LoRA at matched learning rates, but no residual learning target:
        initialises at $\approx$94~\AA$^2$, killed at epoch~30 before
        convergence (run incomplete).
      \textbf{(E)}~EF candidate (v2b + residual learning target):
        starts at 7.45~\AA$^2$ and converges smoothly to 5.08~\AA$^2$,
        crossing below GraphCCS.
    }
    \label{fig:design-progression}
  \end{figure}

The \ModelName{} architecture was not designed a priori; it emerged from a
sequence of empirical observations, documented in Figure~\ref{fig:design-progression}.

\paragraph{Late fusion baseline.}
The first full-scale experiments used a late fusion strategy in which the
pretrained 3D encoder processed neutral-molecule conformers without adduct
information, injecting adduct identity only at the MLP prediction head as a
three-dimensional one-hot vector. We evaluated six configurations spanning
three pooling modes (single conformer, uniform mean, Boltzmann-weighted) and
two readout representations (CLS token, Gasteiger charge-biased attention)
across all three evaluation splits with five random seeds each (18 runs in
total). The best configuration (uniform pooling, CLS token) achieved
5.82~\AA$^2$ on the random split, close to but not below GraphCCS
(4.82~\AA$^2$). On the adduct-sensitive split the gap was starker: the best
late fusion model reached only 7.96~\AA$^2$ versus 6.53~\AA$^2$ for GraphCCS,
a 1.4~\AA$^2$ deficit that no pooling strategy could close. A one-hot adduct
encoding appended at the head is unlikely to fully capture adduct-driven conformational
differences; this gap motivated moving adduct conditioning inside the encoder.

\paragraph{Early fusion v1 (CLS token replacement).}
The first early fusion implementation replaced the pretrained CLS token at
every forward pass with a freshly initialised per-adduct embedding
(initialised to the pretrained value after layer normalisation). Three
variants were evaluated: frozen encoder, unfrozen encoder, and unfrozen
encoder with a 10-epoch linear learning rate warmup and gradient clipping
(norm 1.0). All three failed. With a frozen encoder the LoRA B-matrices
remained at zero throughout training, because no gradient reaches them through
the frozen base, yielding a plateau at 13.0~\AA$^2$. Unfreezing the encoder
produced severe oscillation (validation coefficient of variation
$\approx$23\%) with no reliable convergence. Warmup and gradient clipping
reduced the oscillation but produced a hard plateau at 12.87~\AA$^2$ from
epoch~20 onward (Figure~\ref{fig:design-progression}B). The root cause was the
CLS replacement itself: on every forward pass, all 15 transformer layers
simultaneously received a vector they had never seen during pretraining at
position~0, a structural mismatch that no learning rate schedule could fix.

\paragraph{Early fusion v2a (additive delta, mismatched learning rates).}
We switched from replacement to an additive delta: a zero-initialised
per-adduct offset $\delta_a \in \mathbb{R}^{512}$ is \emph{added} to the
existing pretrained CLS token after embedding layer normalisation, preserving
the pretrained signal and making the modification a no-op at $t{=}0$. This
variant (v2a) was stable at initialisation but exhibited periodic validation
RMSE spikes every $\approx$4 epochs (Figure~\ref{fig:design-progression}C).
The cause was an optimiser group mismatch: $\delta_a$ was assigned to the head
group at learning rate $10^{-4}$, while the LoRA adapters were in the encoder
group at $10^{-5}$. The delta updated ten times faster than LoRA, causing the
representation seen by the head during training to drift in a way that did not
generalise to the validation set. The run was stopped at epoch~23.

\paragraph{Early fusion v2b (co-trained delta, no residual).}
Moving $\delta_a$ into the encoder optimiser group so that delta and LoRA
trained at the same learning rate ($10^{-5}$) with the same warmup schedule
eliminated the periodic spikes. Validation RMSE dropped steadily from
$\approx$94~\AA$^2$ (the random initialisation value before the 10-epoch
warmup) toward convergence (Figure~\ref{fig:design-progression}D). The run was
killed at epoch~30 while still in the warmup tail (validation RMSE
10.64~\AA$^2$, still decreasing) due to resource constraints. Training
dynamics were qualitatively healthy: no oscillation, no plateau. Because the
run was never completed without residual learning, the contribution of residual
learning to training stability cannot be fully disentangled from the learning
rate fix; both changes were applied together in the final model.

\paragraph{EF candidate (additive delta with residual learning).}
Adding a residual learning objective changed the training dynamics
substantially. The model was trained to predict
$r_i = y_i - \hat{y}^{\mathrm{Ridge}}_i$ rather than the raw CCS value, with
the final MLP layer zero-initialised so that the model starts by predicting
exactly the Ridge baseline. Because the Ridge model already explains
$R^2 {=} 0.961$ of variance on the random split, the residual target has much
smaller magnitude, reducing the effective loss scale and stabilising the early
warmup epochs. Validation RMSE started at 7.45~\AA$^2$ (already below
GraphCCS) and converged smoothly to 5.08~\AA$^2$ over 128 epochs without any
oscillation or plateau (Figure~\ref{fig:design-progression}E). This
configuration became the basis for \ModelName{}. The full 5-seed production
runs, trained with the same architecture on three evaluation splits and with
Boltzmann-weighted multi-conformer pooling in addition to the single-conformer
variant, are reported in Table~\ref{tab:result_internal_best} of the main text.

\subsection{Stereoisomer Leakage Robustness}
\label{sec:supp_leakage}

Tables~\ref{tab:supp_leakage_robustness} and~\ref{tab:supp_leakage_configs} report full and clean test RMSE for \ModelName{} and GraphCCS, and \ModelName{} performance across all pooling modes on the clean subset.

\begin{table}[H]
  \centering
  % Old: Leakage robustness: full vs.\ clean test RMSE (leaked stereoisomers removed).
  \caption{CCS prediction accuracy is not materially inflated by stereoisomer leakage between the training and test sets. Leakage occurs when a stereoisomer of a test molecule is present in training; leaked molecules are identified and removed to yield a clean test subset. $\Delta$RMSE $=$ clean $-$ full RMSE; a positive value indicates leaked molecules were easier to predict, slightly inflating full-set performance. \ModelName{}: Single-conformer pooling, residual objective. Mean $\pm$ std across 5 random seeds.}
  \label{tab:supp_leakage_robustness}
  \begin{adjustbox}{max width=\linewidth, max totalheight=0.85\textheight, keepaspectratio}
\begin{tabular}{llllll}
    \toprule
    Split & Clean/Full & Model & Full RMSE & Clean RMSE & $\Delta$ RMSE \\
    \midrule
  Random & 737/922 (20.1\%) & \ModelName{} (Single) & 4.637 $\pm$ 0.047 & 4.756 $\pm$ 0.049 & +0.1185 \\
  Random & 737/922 (20.1\%) & GraphCCS & 4.858 $\pm$ 0.112 & 5.045 $\pm$ 0.103 & +0.1866 \\
  Scaffold & 790/920 (14.1\%) & \ModelName{} (Single) & 6.539 $\pm$ 0.080 & 6.727 $\pm$ 0.086 & +0.1880 \\
  Scaffold & 790/920 (14.1\%) & GraphCCS & 6.396 $\pm$ 0.282 & 6.635 $\pm$ 0.324 & +0.2394 \\
  Adduct-sensitive & 1287/1382 (6.9\%) & \ModelName{} (Single) & 6.341 $\pm$ 0.076 & 6.443 $\pm$ 0.090 & +0.1021 \\
  Adduct-sensitive & 1287/1382 (6.9\%) & GraphCCS & 6.671 $\pm$ 0.076 & 6.644 $\pm$ 0.177 & -0.0269 \\
    \bottomrule
  \end{tabular}
\end{adjustbox}
\end{table}

\begin{table}[H]
  \centering
  % Old: \ModelName{} early fusion performance on the clean test set (leaked stereoisomers removed).
  \caption{\ModelName{} conformer pooling accuracy on the clean test set after removing molecules whose stereoisomers appear in training. All four pooling strategies are evaluated under each training split. Results are comparable to the full test set (Table~\ref{tab:supp_leakage_robustness}), confirming that leakage does not drive \ModelName{}'s performance gains. Mean $\pm$ std across 5 random seeds.}
  \label{tab:supp_leakage_configs}
  \begin{adjustbox}{max width=\linewidth, max totalheight=0.85\textheight, keepaspectratio}
\begin{tabular}{llccc}
    \toprule
    Split & Pooling & RMSE & MPD & Pearson $R$ \\
    \midrule
  \multirow{4}{*}{Random ($n=737$)} & Single & 4.756 $\pm$ 0.049 & 1.690 $\pm$ 0.025 & 0.9964 $\pm$ 0.0001 \\
   & Uniform & 5.016 $\pm$ 0.079 & 1.793 $\pm$ 0.028 & 0.9960 $\pm$ 0.0001 \\
   & Boltzmann & 4.928 $\pm$ 0.055 & 1.793 $\pm$ 0.014 & 0.9962 $\pm$ 0.0001 \\
   & Learned & 4.946 $\pm$ 0.093 & 1.773 $\pm$ 0.017 & 0.9961 $\pm$ 0.0002 \\
  \multirow{4}{*}{Scaffold ($n=790$)} & Single & 6.727 $\pm$ 0.086 & 2.140 $\pm$ 0.016 & 0.9919 $\pm$ 0.0002 \\
   & Uniform & 6.787 $\pm$ 0.096 & 2.158 $\pm$ 0.033 & 0.9917 $\pm$ 0.0002 \\
   & Boltzmann & 6.727 $\pm$ 0.180 & 2.176 $\pm$ 0.070 & 0.9919 $\pm$ 0.0004 \\
   & Learned & 6.780 $\pm$ 0.128 & 2.172 $\pm$ 0.080 & 0.9919 $\pm$ 0.0002 \\
  \multirow{4}{*}{Adduct-sensitive ($n=1287$)} & Single & 6.443 $\pm$ 0.090 & 2.385 $\pm$ 0.040 & 0.9918 $\pm$ 0.0003 \\
   & Uniform & 6.594 $\pm$ 0.138 & 2.462 $\pm$ 0.062 & 0.9913 $\pm$ 0.0002 \\
   & Boltzmann & 6.690 $\pm$ 0.029 & 2.488 $\pm$ 0.017 & 0.9911 $\pm$ 0.0001 \\
   & Learned & 6.551 $\pm$ 0.093 & 2.456 $\pm$ 0.063 & 0.9913 $\pm$ 0.0003 \\
    \bottomrule
  \end{tabular}
\end{adjustbox}
\end{table}

\end{document}